\documentclass{article}

\usepackage{arxiv}
\usepackage{fancyhdr}
\usepackage[utf8]{inputenc} 
\usepackage[T1]{fontenc}    
\usepackage{hyperref}       
\usepackage{url}            
\usepackage{booktabs}       
\usepackage{amsfonts}       
\usepackage{nicefrac}       
\usepackage{microtype}      
\usepackage{lipsum}		
\usepackage{graphicx}
\usepackage[numbers]{natbib}
\usepackage{doi}
\usepackage{amsmath}

\usepackage{color}

\title{Discrete Solution Operator Learning for Geometry-Dependent PDEs}

\author{ \href{https://orcid.org/0000-0002-0753-6428}{\includegraphics[scale=0.06]{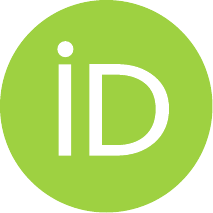}\hspace{1mm}Jinshuai Bai} \thanks{These authors contributed equally to this work and should be considered co-first authors.} { \textsuperscript{,}} \thanks{Corresponding authors.}\\
	Institute of Biomechanics and Medical Engineering\\
	Applied Mechanics Laboratory (AML)\\
	Tsinghua University
    Beijing 100084, China \\
	\texttt{bjs@mail.tsinghua.edu.cn} \\
	\And
	\href{https://orcid.org/0000-0002-4001-5253}{\includegraphics[scale=0.06]{orcid.pdf}\hspace{1mm}Haolin Li} \textsuperscript{ *} \\
	Department of Aeronautics \\
	Imperial College London\\
	London, SW7 2AZ, United Kingdom \\
	\texttt{haolin.li20@imperial.ac.uk} \\
    \And
    \href{https://orcid.org/0000-0001-5106-2197}{\includegraphics[scale=0.06]{orcid.pdf}\hspace{1mm}Zahra Sharif Khodaei} \\
	Department of Aeronautics\\
	Imperial College London\\
	London, SW7 2AZ, United Kingdom \\
	\texttt{z.sharif-khodaei@imperial.ac.uk} \\
    \And
    \href{https://orcid.org/0000-0002-2883-2461}{\includegraphics[scale=0.06]
    {orcid.pdf}\hspace{1mm}M. H. Aliabadi} \\
	Department of Aeronautics\\
	Imperial College London\\
	London, SW7 2AZ, United Kingdom \\
	\texttt{m.h.aliabadi@imperial.ac.uk} \\
    \And
	\href{https://orcid.org/0000-0002-2770-5014}{\includegraphics[scale=0.06]{orcid.pdf}\hspace{1mm}YuanTong Gu}\textsuperscript{ $\dagger$} \\
	School of Mechanical, Medical, and Process Engineering\\
	Queensland University of Technology\\
	Brisbane, QLD, 4000, Australia \\
	\texttt{yuantong.gu@qut.edu.au} \\
    \And
	\href{https://orcid.org/0000-0001-6894-7979}{\includegraphics[scale=0.06]{orcid.pdf}\hspace{1mm}Xi-Qiao Feng} \textsuperscript{ $\dagger$}\\
	Institute of Biomechanics and Medical Engineering\\
	Applied Mechanics Laboratory (AML)\\
	Tsinghua University
    Beijing 100084, China \\
	\texttt{fengxq@tsinghua.edu.cn} \\
}

\renewcommand{\shorttitle}{\textit{arXiv} Template}

\hypersetup{
pdftitle={Discrete Solution Operator Learning (DiSOL): Learning Discrete Solution Procedures for Geometry-Dependent PDEs},
pdfsubject={q-bio.NC, q-bio.QM},
pdfauthor={Jinshuai Bai, Haolin Li, },
pdfkeywords={First keyword, Second keyword, More},
}

\begin{document}
\maketitle

\begin{abstract}
Neural operator learning accelerates PDE solution by approximating operators as mappings between continuous function spaces. Yet in many engineering settings, varying geometry induces discrete structural changes, including topological changes, abrupt changes in boundary conditions or boundary types, and changes in the computational domain, which break the smooth-variation premise. Here we introduce Discrete Solution Operator Learning (DiSOL), a complementary paradigm that learns discrete solution procedures rather than continuous function-space operators. DiSOL factorizes the solver into learnable stages that mirror classical discretizations: local contribution encoding, multiscale assembly, and implicit solution reconstruction on an embedded grid, thereby preserving procedure-level consistency while adapting to geometry-dependent discrete structures. Across geometry-dependent Poisson, advection-diffusion, linear elasticity, as well as spatiotemporal heat conduction problems, DiSOL produces stable and accurate predictions under both in-distribution and strongly out-of-distribution geometries, including discontinuous boundaries and topological changes. These results highlight the need for procedural operator representations in geometry-dominated problems and position discrete solution operator learning as a distinct, complementary direction in scientific machine learning.
\end{abstract}

\keywords{Neural Operator \and Discrete numerical methods \and Deep Learning \and Partial Differential Equations}

\section{Introduction}

\begin{figure}
    \centering
    \includegraphics[width=1.\linewidth]{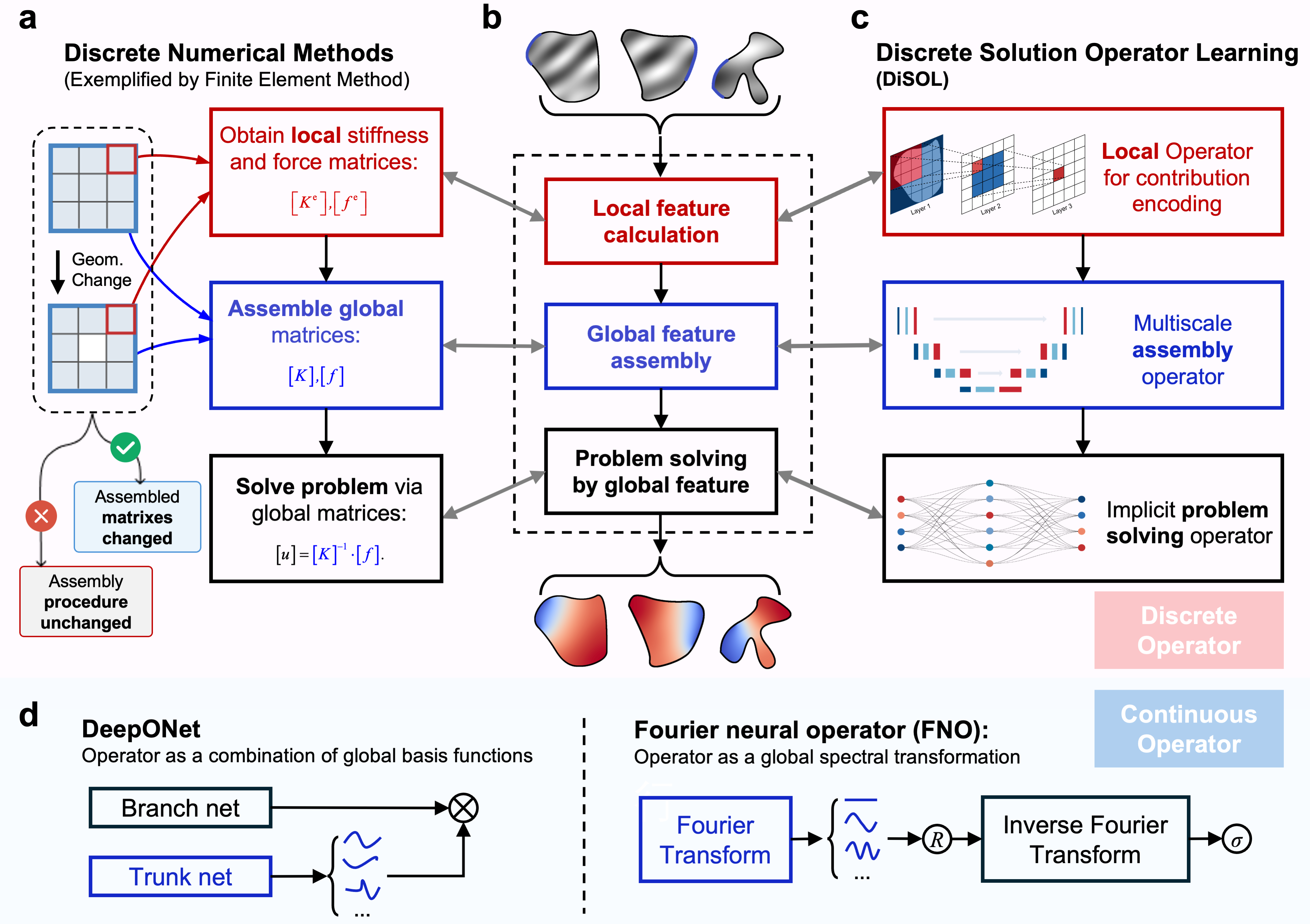}
    \caption{\textbf{Discrete Solution Operator Learning (DiSOL) in contrast to continuous neural operator paradigms.} \textbf{a}, Classical discrete numerical methods for partial differential equations (PDEs), exemplified by the finite element method. A PDE operator can be realized through a sequence of fixed discrete procedures, including local operator evaluation (e.g., element-level stiffness and force computation), global assembly, and solution via assembled global systems. Under geometry variation, the assembled results change, whereas the underlying assembly procedures remain unchanged. \textbf{b}, An abstracted view of discrete problem solving, highlighting the canonical workflow shared by many numerical methods: local contributions evaluation, global information assembly, and solution through globally aggregated representations. \textbf{c}, Discrete Solution Operator Learning (DiSOL). Instead of explicitly constructing numerical matrices, DiSOL represents the discrete solution process through learnable neural operators that conceptually correspond to the local operator, the multiscale assembly operator, and the implicit problem-solving operator. These components collectively form a discrete operator that preserves procedure-level invariance under geometry variation, in the sense that the underlying solution procedure remains unchanged, while adapting its outputs to geometry-dependent discrete structures. \textbf{d}, Representative continuous neural operator paradigms, illustrated by DeepONet \cite{DeepONetLu2021} and FNO \cite{FNOLi2020}. These approaches model PDE operators as global mappings in continuous function spaces, using global basis combinations or spectral transformations. Such representations are primarily designed for smooth function-space mappings, and may become misaligned when geometry changes and discontinuities dominate the problem setting. In contrast, DiSOL does not aim to approximate continuous operators directly, but learns discrete solution procedures, leading to a fundamentally different inductive bias in geometry-dependent problem settings.}
    \label{fig:fig1}
\end{figure}

The numerical solution of partial differential equations (PDEs) lies at the core of scientific computing and engineering analysis \cite{ames2014numerical,reddy1993introduction,aliabadi2002boundary,liu2005introduction}. For decades, mature discrete numerical algorithms, such as finite element \cite{reddy1993introduction}, finite volume \cite{leveque2002finite}, and finite difference \cite{thomas2013numerical} methods, have provided reliable and accurate solutions across a wide range of applications. A defining characteristic of these methods is that they operate on discrete computational representations: local contributions are evaluated on elements, assembled into global representations, and subsequently solved through well-defined procedures. While geometry variation alters the instantiated discrete system, the underlying solution procedures by which local contributions are evaluated and assembled remain unchanged (\textbf{Fig.}~\ref{fig:fig1}a). That is, the same local evaluation and assembly rules are reused wherever they are activated, even though the activated locations change with geometry. In this sense, classical discrete numerical algorithms implement a solution operator: they define a mapping from problem specification (geometry, coefficients, sources, and boundary conditions) to the discrete solution field, executed through an explicit local-to-global procedure.

Recently, data-driven surrogate and operator learning methods have emerged as a promising alternative for accelerating PDE solving across varying inputs and physical parameters \cite{Bai2025AMS,NMIEditorial,Azizzadenesheli2024,Brandstetter2025}. Two complementary directions have been widely explored. One treats PDE learning as field-to-field regression on discrete grids (or strong discretization-based surrogates) using convolutional architectures, often in encoder-decoder or UNet-based forms \cite{pmlrlong18PDENET, LONG2019108925,zhou2025unisolver}, achieving strong empirical performance when inputs and outputs are naturally represented on fixed-resolution discretizations. The other focuses on continuous neural operators—such as DeepONet \cite{DeepONetLu2021}, the Fourier Neural Operator (FNO) \cite{FNOLi2020}, and their variants \cite{Cao2024,eshaghi2025variational,he2024sequential,li2025architectural,yin2024scalable,li2023fourier, chen2024learning,lu2022comprehensive,wang2021learning,tran2021factorized,bonev2023spherical}—that approximate smooth mappings between input and output functions in continuous function spaces (\textbf{Fig.}~\ref{fig:fig1}d). Many of these methods have demonstrated impressive performance on benchmark problems with regular domains and smoothly varying geometries \cite{winovich2019convpde,DUPREZ2026109131}. These advances also highlight the role of inductive bias, i.e., structural assumptions encoded in model architectures or learning objectives that shape what solution mechanisms can be learned efficiently from finite data. 

However, in many practical engineering problems, geometry variation does not occur smoothly. Instead, it often manifests through sharp corners, internal holes, topological changes, or discontinuous boundary specifications \cite{yau1982survey,sapiro2006geometric}. When geometry-induced non-smooth variations dominate the problem complexity, the mapping from inputs to solutions may not, in general, be reliably captured through approximation within a smooth function-space operator representation.  

This observation highlights a distinction that is often overlooked: although classical discrete numerical solvers can be regarded as operators, learning a continuous operator is not equivalent to learning a discrete numerical solution procedure. Classical solvers do not approximate a single global mapping; instead, they execute a sequence of fixed procedures (local contribution evaluation, global assembly, and solution) which is invariant under geometry variation, even though the assembled global information changes (\textbf{Figs.}~\ref{fig:fig1}a,b) \cite{liu2003smoothed, reddy2005introduction, belytschko2014nonlinear, BAI2024117159}. Approximating a smooth functional mapping may therefore be misaligned with the procedural nature of geometry-dependent PDE solution processes. 

Here, we introduce Discrete Solution Operator Learning (DiSOL), a paradigm that explicitly targets operator learning towards discrete solution procedures. DiSOL shifts the learning target from approximating smooth functional mappings to learning how local contributions, boundary constraints, and multi-scale information are instantiated and assembled on discretized computational representations. Conceptually, DiSOL mirrors the canonical stages of numerical solvers, namely local evaluation, global assembly, and solution, while remaining fully data-driven and end-to-end differentiable (\textbf{Fig.}~\ref{fig:fig1}c). DiSOL should therefore be viewed not as a variant of continuous neural operators, but as a complementary paradigm with a different modeling objective and inductive bias: it preserves procedure-level invariance under geometry variation while allowing the instantiated discrete structure to adapt to geometry-dependent configurations. 

We validate DiSOL on four classes of geometry-dependent PDEs with increasing complexity: a Poisson equation, an advection-diffusion equation spanning diffusion- and transport-dominated cases, a vector-valued linear elasticity problem, and a spatiotemporal thermal conduction problem. Across these cases, DiSOL delivers stable and accurate predictions under both in-distribution (ID) and out-of-distribution (OOD) computational domains, whereas continuous neural operators exhibit structural degradation when geometry-induced discrete variation dominates the problem setting. Moreover, our results indicate that these failures arise primarily from a mismatch between the assumptions of continuous operators and geometry-induced discrete solution structures, rather than from optimization difficulty or insufficient model capacity. 

Taken together, our findings advocate a shift in how operator learning is conceptualized for geometry-dependent PDEs and other settings where discrete structural variation dominates the solution mechanism: learning discrete solution procedures provides a more faithful pathway toward reliable data-driven solvers that generalize across complex geometries. 

\section{Results}
\label{sec:results}
DiSOL targets the learning of a discrete solution operator defined on a complete discrete problem specification:
\begin{equation}
    \mathcal{G}_h:(\Omega_h,\Gamma_h,f_h,\eta_h) \mapsto u_h,
    \label{eq:eq1}
\end{equation}
where $\Omega_h$ denotes the discrete computational domain (the geometry mask and the fixed grid-neighborhood connectivity), $\Gamma_h$ encodes boundary-type on that domain, $f_h$ denotes source fields, $\eta_h$ collects problem-dependent physical parameters (e.g., diffusion coefficient, advection velocity, elastic moduli) that are fixed or prescribed for a given task, and $u_h$ denotes the normalized solution \emph{pattern} field on the grid. In the main text we focus on learning the normalized solution pattern $u_h := U_h/u_{\lim}$ (see Methods, Section~\ref{sect:4.1} and Supplementary Information~D); all models predict $\hat{u}_h$, and all reported metrics are computed on $u_h$ restricted to $\Omega_h$. All experiments below follow Eq.~\ref{eq:eq1}.

Here, the subscript $h$ denotes a fixed discrete representation (grid resolution in this work) as in standard numerical discretization. All primary experiments train and evaluate at a fixed resolution, consistent with practical surrogate usage. We additionally report a preliminary zero-shot cross-resolution test in the Supplementary Information E.5. In all experiments, zero Dirichlet conditions are imposed on the selected boundary segments provided as input, while homogeneous Neumann conditions are implicitly assumed on all other boundaries, consistent with the finite-element data generation. Details of the data generation procedure and parameter settings are provided in Supplementary Information~B, and detailed architecture and training settings of all models are provided in Supplementary Information~C.

\subsection{Geometry-dependent Poisson problem}

\begin{figure}
    \centering
    \includegraphics[width=1.0\linewidth]{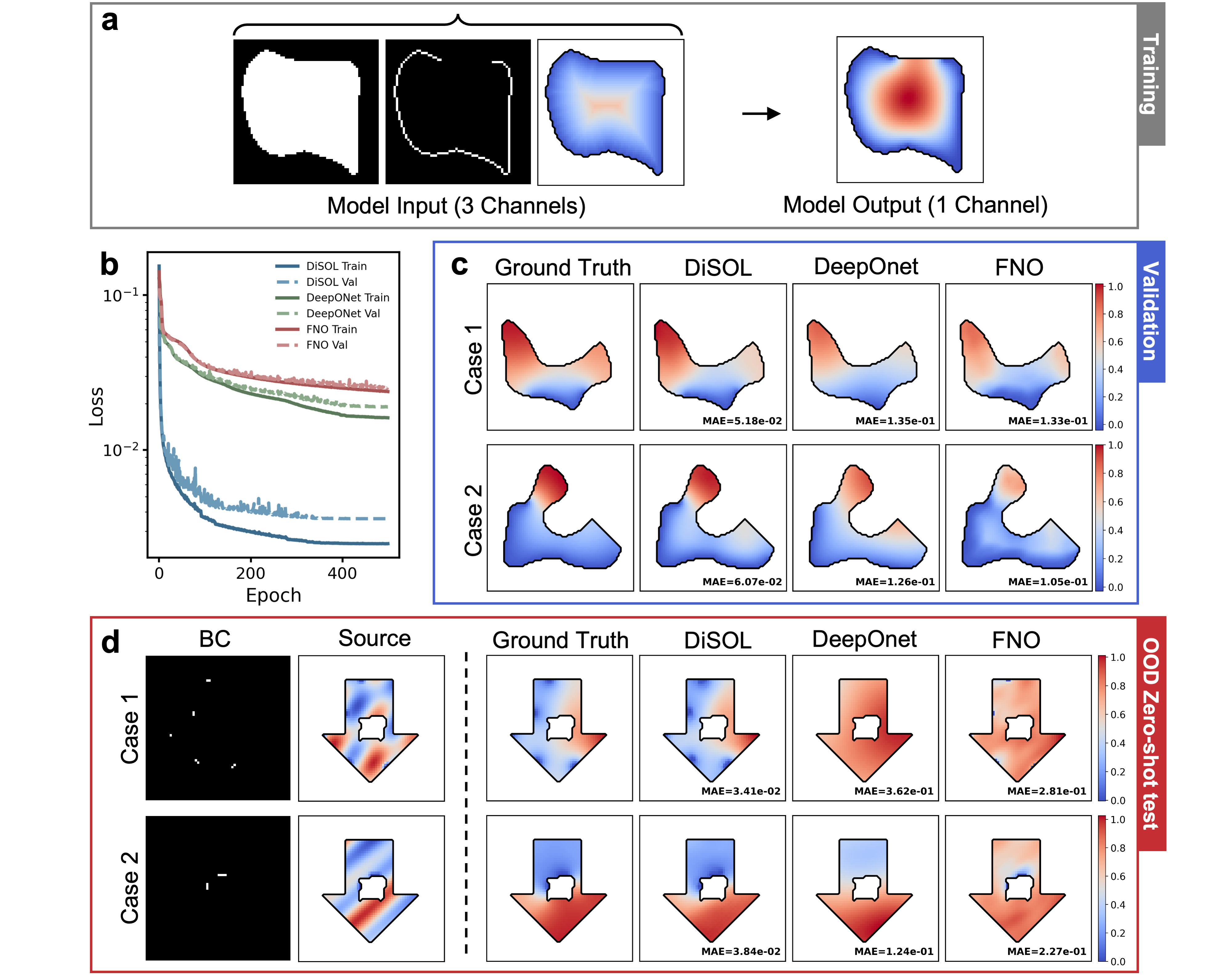}
    \caption{\textbf{Discrete solution operator learning (DiSOL) for a geometry-dependent 2D Poisson problem.} \textbf{a}, Problem formulation and data representation. The input to the model consists of three discrete fields defined on a fixed Cartesian grid: a geometry mask, a boundary-condition selection map, and a source term. The model outputs the corresponding discrete solution pattern. \textbf{b}, Training and validation loss histories for DiSOL, DeepONet and the FNO under comparable model capacity and training settings. DiSOL converges faster and achieves substantially lower validation loss (All models are trained under the same optimizer/schedule and have comparable parameter counts, $\approx0.13$M each; details in Supplementary Information C.4 and C.5). \textbf{c}, Representative ID validation cases. While the reference solutions are globally smooth and continuous operator baselines capture the coarse trend, DeepONet and FNO show mild yet systematic deviations in the normalized pattern, particularly near irregular geometry and boundary-adjacent regions where geometry-conditioned boundary activation influences local responses. DiSOL provides closer agreement with the ground truth and achieves lower errors under the same ID protocol. \textbf{d}, OOD generalization tests involving unseen geometries with sharp corners and internal holes, discontinuous and localized boundary conditions, and high-frequency source terms. DiSOL remains stable and consistent with the reference solutions, whereas DeepONet and FNO show substantial degradation in solution-pattern fidelity. See Supplementary Information A.1, B and E.1 for full setup and extended statistics.}
    \label{fig:fig2}
\end{figure}

We first consider a geometry-dependent elliptic problem as a minimal setting to isolate the effect of domain variation. The governing equation is fixed, while the computational domain and boundary configuration vary across samples, including discrete changes in topology and boundary activation. We study a scalar Poisson-type problem; detailed formulations are provided in Supplementary Information A.1.

Unlike standard operator-learning benchmarks defined on regular domains, geometry, boundary specifications, and source distributions are provided as discrete input channels (\textbf{Fig.}~\ref{fig:fig2}a), and the solution field is predicted on the same embedded grid. Although the representation is grid-based, geometric variation changes the effective computational region and boundary segments, so generalization requires adapting to discrete structural changes rather than smooth interpolation on a fixed domain.

The training dataset contains simply connected geometries with smooth outer boundaries, continuous boundary conditions along the exterior boundary, and low-frequency source terms, yielding globally smooth reference solutions. Even in this mild regime, geometry-induced changes in active regions and boundary segments expose a fundamental distinction between continuous and discrete operator learning.

Using identical training strategies and comparable model capacities, we train DiSOL, DeepONet, and FNO. As shown in \textbf{Fig.}~\ref{fig:fig2}b, DiSOL converges rapidly and achieves substantially lower training/validation errors, whereas FNO saturates early; DeepONet provides a stronger continuous-operator baseline in this setting and is used as the primary continuous baseline (FNO is reported for completeness).

\textbf{Figure}~\ref{fig:fig2}c shows representative ID test cases (additional examples in Supplementary Information E.1). While DeepONet/FNO capture the coarse trend, they exhibit systematic residual distortions near irregular geometric features and boundary-adjacent regions, leading to higher errors than DiSOL. DiSOL more faithfully reproduces both global structure and boundary-modulated local variations under the same ID setting. A capacity-matched U-Net control study is provided in Supplementary Information H.

We additionally benchmark three representative geometry-aware variants of continuous neural operators, including Diffeomorphic Mapping Operator Learning (DIMON) \cite{yin2024scalable}, Geo-FNO \cite{li2023fourier}, and GNO \cite{li2020neural}. They implement different geometric inductive bias via explicit geometric representations or graph-based aggregation. These variants yield at most modest improvements over vanilla DeepONet/FNO and remain clearly below DiSOL under both ID and geometry OOD shifts; detailed comparisons are provided in Supplementary Information E.1.

We further assess generalization under strongly OOD conditions (\textbf{Figs.}~\ref{fig:fig2}d and \textbf{S6}) that combine sharp corners and internal holes (topology changes), highly localized/discontinuous boundary conditions, and higher-frequency source components. DiSOL remains stable and closely matches the reference solutions, whereas DeepONet exhibits pronounced failures, including deviations in global solution patterns rather than only localized errors.

Overall, even for the classical two-dimensional Poisson equation with globally smooth solutions, continuous neural operator methods show intrinsic limitations when geometry induces discrete structural variation in the instantiated computational domain. By learning and assembling local contributions directly on discretized domains, DiSOL aligns with numerical solution procedures and achieves robust predictions under both ID and OOD geometric conditions.

\subsection{Geometry-dependent advection-diffusion problem}

\begin{figure}
    \centering
    \includegraphics[width=1.\linewidth]{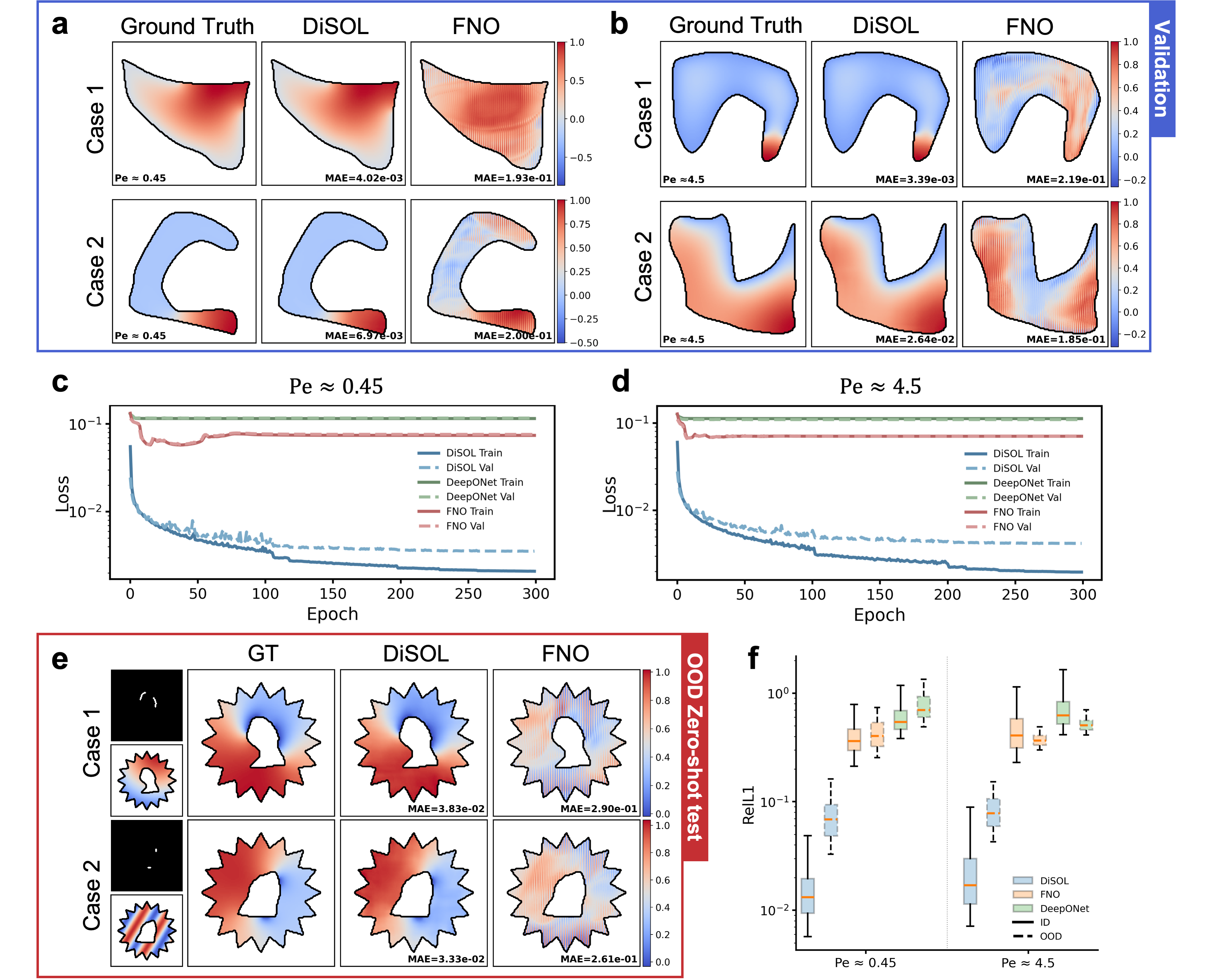}
    \caption{\textbf{Discrete solution operator learning (DiSOL) results for the advection-diffusion equation.} \textbf{a,b}, Representative ID validation cases under two transport settings: a diffusion-dominated case ($\text{Pe} \approx 0.45$) and an advection-dominated case ($\text{Pe} \approx 4.5$). Ground-truth solutions are compared with predictions from DiSOL and FNO. As transport effects become stronger and solution structures align more closely with flow direction and complex geometric boundaries, DiSOL consistently preserves the global transport patterns and boundary-induced features, whereas FNO exhibits increasingly pronounced structural deviations, particularly near irregular geometric features. \textbf{c,d}, Training and validation loss histories for DiSOL, FNO, and DeepONet under the same two Péclet number settings ($\text{Pe} \approx 0.45$ and $\text{Pe} \approx 4.5$). In both cases, DiSOL converges rapidly and achieves substantially lower training and validation errors, while the continuous neural operator baselines saturate at significantly higher error levels. \textbf{e}, OOD generalization results for the advection-diffusion equation at $\text{Pe} \approx 4.5$, involving unseen geometries with sharp corners and topological changes, together with localized and discontinuous boundary conditions. DiSOL preserves the overall transport structure and solution morphology. \textbf{f}, Statistical comparison of relative L1 errors across different models under ID and OOD settings for both $\text{Pe} \approx 0.45$ and $\text{Pe} \approx 4.5$. Box plots (logarithmic scale) show that DiSOL consistently outperforms continuous neural operator baselines in both cases, and that performance degradation is primarily driven by geometric distribution shifts rather than by increased transport dominance. See Supplementary Information A.2, B and E.2 for full setup and extended statistics.}
    \label{fig:fig3}
\end{figure}

We next consider a steady advection-diffusion problem to examine geometry generalization under transport-dominated dynamics. While the governing equation and physical parameters are fixed within each Péclet case, variations in domain geometry alter effective transport pathways and boundary-layer structures, providing a stringent test of generalization under geometry-induced structural change. The formulation and parameter settings are provided in Supplementary Information A.2.

We use the same embedded-grid representation as in the Poisson case (mask, boundary-condition map, and source as inputs; scalar field as output) and train separate operators for two regimes, $\mathrm{Pe}\approx0.45$ and $\mathrm{Pe}\approx4.5$. Representative ID validation examples are shown in \textbf{Figs.}~\ref{fig:fig3}a--b (additional results in Supplementary Information E.2). Across both regimes, DiSOL captures the main transport structures and their coupling with complex geometries, whereas continuous-operator baselines exhibit noticeable structural distortions, particularly near geometric boundaries and irregular features.

\textbf{Figures}~\ref{fig:fig3}c--d report training/validation histories for the two Péclet cases. DiSOL converges stably and reaches substantially lower errors, while continuous baselines saturate early in both regimes, suggesting that increased transport dominance alone is not the primary source of difficulty in operator learning under geometry variation (DeepONet is reported for completeness).

We then evaluate generalization under strongly OOD geometries at high Péclet numbers (\textbf{Figs.}~\ref{fig:fig3}e--f and \textbf{S10}), combining sharp corners/topology changes with highly localized and spatially discontinuous boundary conditions. DiSOL maintains stable performance with superior error distributions, and the relative $L_1$ boxplots indicate that degradation is driven primarily by geometry shifts rather than the increase in $\mathrm{Pe}$.

As an additional test, we benchmark geometry-aware variants of continuous neural operators (GNO and Geo-FNO, and DIMON where applicable). These approaches provide at most modest gains over vanilla DeepONet/FNO and remain clearly below DiSOL under both ID and geometry OOD shifts; full results are reported in Supplementary Information E.2.

Taken together, these results show that the limitations of continuous neural operator baselines are not explained by transport strength alone, but by geometry-induced discrete structural changes that strongly modulate transport-dominated solutions. By learning and assembling local contributions directly on discretized domains, DiSOL captures the coupling between geometry and transport effects and achieves robust generalization in both ID and OOD settings.

\subsection{Geometry-dependent linear elasticity problem}

\begin{figure}
    \centering
    \includegraphics[width=1.\linewidth]{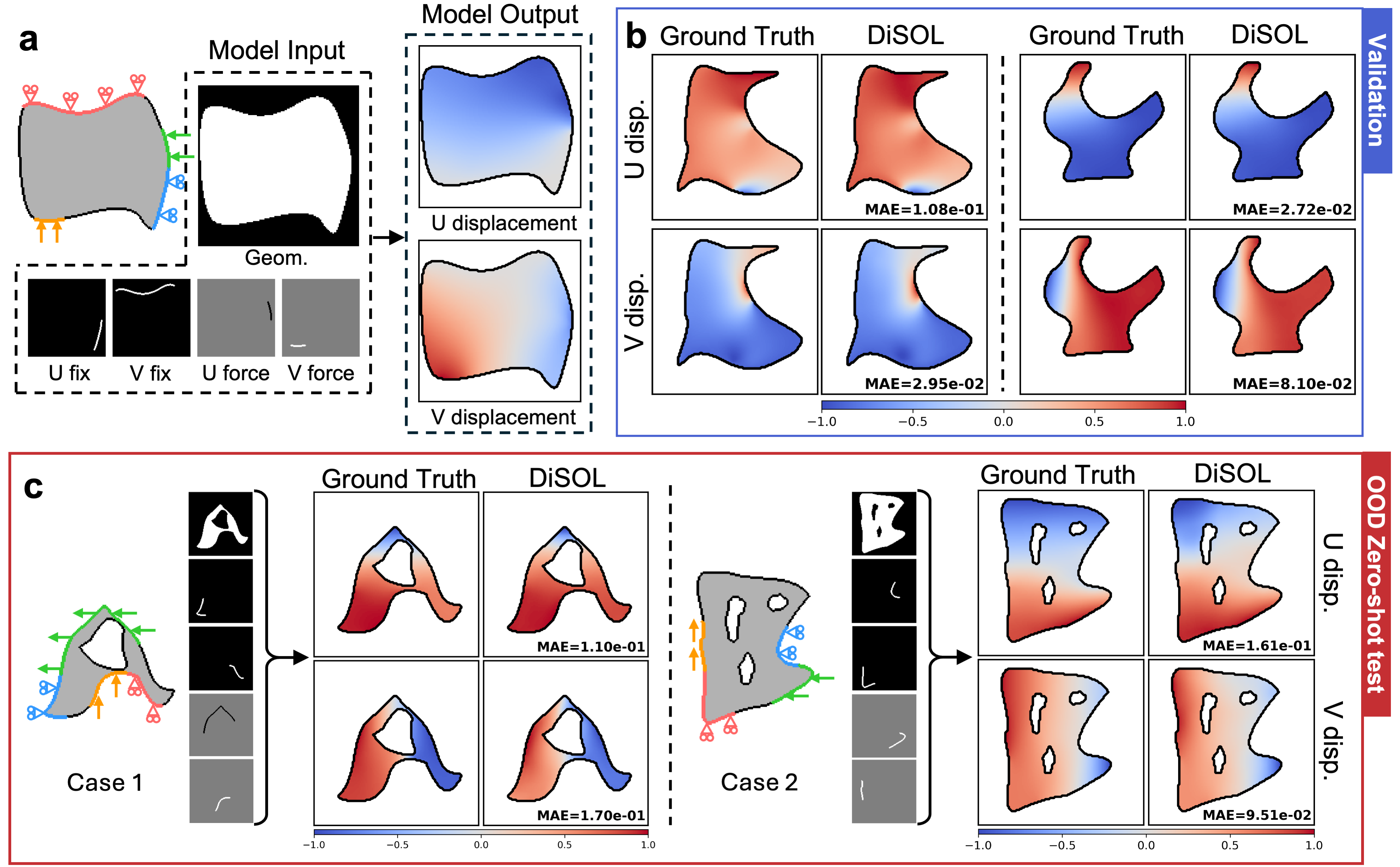}
    \caption{\textbf{Discrete solution operator learning (DiSOL) results for geometry-dependent 2D linear elasticity.} \textbf{a}, Problem formulation and data representation for the linear elastic solid mechanics problem. The input to the model consists of five discrete channels defined on a fixed Cartesian grid, encoding the geometry mask, displacement boundary conditions in the horizontal and vertical directions, and externally applied force components. The model outputs the corresponding two-component displacement field $(\hat{u}_x,\hat{u}_y)$ defined on the same grid. \textbf{b}, Representative ID validation results. Ground-truth displacement fields are compared with predictions from the proposed discrete solution operator (DiSOL) for two representative geometries and loading configurations. DiSOL accurately reconstructs both horizontal and vertical displacement components, capturing the global deformation patterns and boundary-induced responses without introducing spurious oscillations or rigid-body artifacts. \textbf{c}, OOD generalization results involving unseen geometries with increased complexity, including internal holes, topological changes, and altered boundary condition configurations. Despite the substantial changes in geometric structure and boundary constraints, DiSOL preserves the overall deformation modes and displacement distributions, maintaining consistency with the reference solutions for both displacement components. See Supplementary Information A.3, B and E.3 for full setup and extended statistics.}
    \label{fig:fig4}
\end{figure}

Building upon the scalar-field problems discussed above, we further consider a two-dimensional linear elasticity problem to evaluate geometry generalization for vector-valued and coupled solution patterns. Under the same discrete operator formulation $\mathcal{G}_h$, the learning task targets a vector-valued displacement field $(u_x, u_y)$. The discrete problem specification is encoded through five input channels (\textbf{Fig.}~\ref{fig:fig4}a), which jointly represent the geometry mask $\Omega_h$ and the boundary and loading conditions required to determine the elasticity solution. Linear elasticity is a fundamental model in computational mechanics, whose solution behavior is jointly governed by geometric shape, displacement constraints, and external loads. This makes it a stringent test case for assessing whether operator learning methods can robustly handle multi-component, coupled solution patterns under geometry-dependent discrete settings. The precise formulation and parameter settings are provided in Supplementary Information A.3.

As in the previous cases, the computational domain geometry is specified by a discrete mask on a fixed Cartesian grid. The five input channels encode the geometry mask, prescribed displacement constraints in the horizontal and vertical directions, and the corresponding external force distributions in each direction (\textbf{Fig.}~\ref{fig:fig4}a). The model output is a two-component displacement field defined on the same grid, representing the horizontal ($u_x$) and vertical ($u_y$) displacement components. Compared to the scalar-field problems considered earlier, this setting introduces additional challenges at the operator level, including multi-channel inputs, vector-valued outputs, and strong coupling between boundary conditions and body forces. These factors collectively increase the structural complexity of the underlying solution operator.

\textbf{Figure}~\ref{fig:fig4}b presents representative ID validation results (more results are presented in Supplementary Information E.3). DiSOL accurately reconstructs the spatial distributions of both displacement components, with predicted patterns closely matching the reference results in terms of global deformation patterns, boundary-induced displacement responses, and local gradient variations. Notably, no common non-physical artifacts, such as unphysical oscillations, rigid-body drift, or inconsistencies between displacement components, are observed. This indicates that, within the training distribution, DiSOL successfully captures the discrete solution structure jointly determined by geometry, boundary conditions, and external loading.

We further evaluate generalization under OOD geometric conditions (\textbf{Fig.}~\ref{fig:fig4}c) by introducing complex geometries containing internal holes, which induce topological changes and substantially modify the discrete computational structure. Although such configurations are absent during training, DiSOL preserves the overall deformation patterns of the displacement fields. The predicted patterns remain consistent with the numerical references in their spatial organization, and no pronounced structural instability or global distortion is observed.

Taken together, these results demonstrate that the proposed discrete solution operator learning framework extends naturally beyond scalar elliptic and transport-type equations to geometry-dependent, vector-valued mechanics problems. By explicitly modeling local contributions on discrete computational domains and assembling them across multiple scales, DiSOL captures discrete structural variations induced jointly by geometry, boundary conditions, and external loads. As a result, the framework delivers stable and reliable predictions under both ID and OOD scenarios.

\subsection{Geometry-dependent thermal conduction problem}

\begin{figure}
    \centering
    \includegraphics[width=1.0\linewidth]{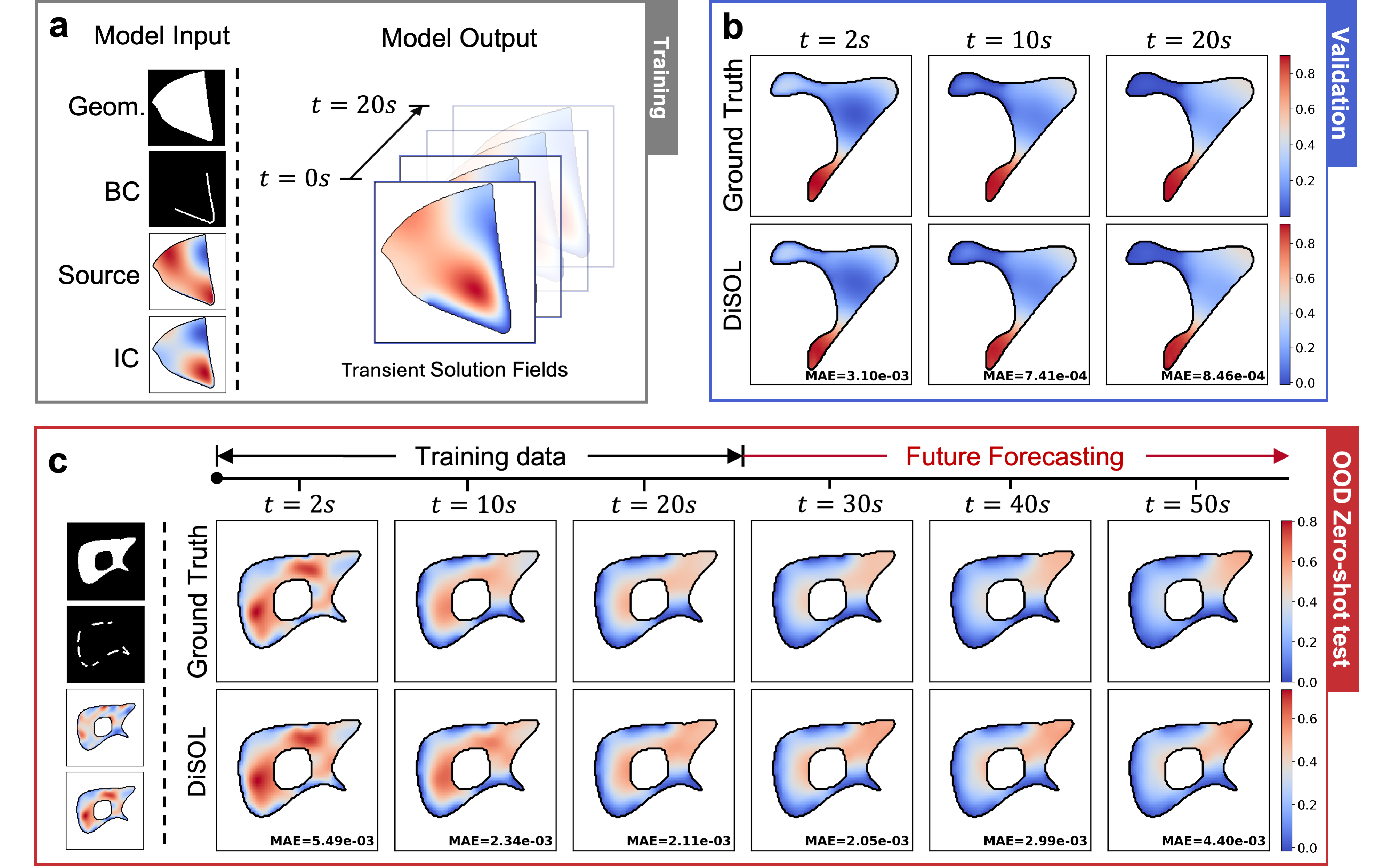}
    \caption{\textbf{Discrete solution operator learning (DiSOL) for the heat conduction problem.} \textbf{a}, Problem formulation and data representation. The model input consists of discrete channels defined on a fixed Cartesian grid, encoding the geometry mask, boundary condition locations, source term distribution, and IC. The time coordinate is provided as an additional input channel. The model predicts the temperature field at queried times $t$; during training, $t$ is sampled from $[0,20]\,\mathrm{s}$. \textbf{b}, Representative ID validation results at $t=2\,\mathrm{s}$, $10\,\mathrm{s}$, and $20\,\mathrm{s}$, showing close agreement between ground-truth solutions and DiSOL predictions. \textbf{c}, Zero-shot OOD evaluation combining geometric extrapolation and temporal forecasting. The model is tested on an unseen geometry with internal voids and altered boundary configurations. Predictions at $t=30\,\mathrm{s}$, $40\,\mathrm{s}$, and $50\,\mathrm{s}$ extend beyond the training temporal interval, corresponding to genuine future forecasting scenarios. Color bars indicate normalized temperature magnitude. See Supplementary Information A.4, B and E.4 for full setup and extended statistics.}
    \label{fig:fig5}
\end{figure}

The preceding examples have demonstrated the effectiveness of DiSOL for geometry-dependent operator learning problems in static settings. We next examine whether the proposed discrete operator learning framework extends naturally to dynamic problems involving spatiotemporal evolution. To this end, we consider a two-dimensional heat equation as a representative time-dependent PDE; the precise formulation is provided in Supplementary Information A.4. 

As in the previous cases, discretized geometry, boundary conditions, source terms, and the initial condition (IC) are supplied as discrete input channels to the model (\textbf{Fig.}~\ref{fig:fig5}a). To account for temporal evolution, the time coordinate is incorporated as an additional input channel and provided in parallel with the spatial inputs. Under this formulation, the learning task corresponds to approximating a discrete spatiotemporal solution operator defined on geometry-dependent computational domains. The model is trained on trajectories spanning the temporal interval $t\in[0,20]\,\mathrm{s}$.

\textbf{Figure}~\ref{fig:fig5}b presents representative validation results at $t=2\,\mathrm{s}$, $10\,\mathrm{s}$, and $20\,\mathrm{s}$ (more results are presented in Supplementary Information E.4). DiSOL accurately reproduces the evolving temperature field across all evaluated time instances, capturing both the global diffusion patterns and geometry-induced local variations. The close agreement between predicted and reference solutions indicates stable learning of the underlying spatiotemporal dynamics within the training temporal window.

We further assess generalization under a zero-shot OOD setting (\textbf{Figs.}~\ref{fig:fig5}c and \textbf{S15}), where multiple unseen factors are introduced simultaneously. These include increased geometric complexity with internal voids, altered and discontinuous boundary configurations, and higher-frequency sources and IC fields. In addition to spatial extrapolation, this test also probes temporal extrapolation: predictions at $t=30\,\mathrm{s}$, $40\,\mathrm{s}$, and $50\,\mathrm{s}$ extend beyond the training temporal interval and therefore constitute genuine future forecasting rather than interpolation. Despite this combined OOD setting in geometry, boundary conditions, sources and IC fields, and time, DiSOL remains stable and produces solution patterns that closely match the numerical reference solutions across all evaluated time instances.

These results indicate that the discrete operator formulation underlying DiSOL extends coherently to dynamic, geometry-dependent problems. By explicitly modeling local contributions on discrete computational domains and assembling them across multiple scales, the framework preserves robustness under simultaneous spatial and temporal distribution shifts.

\section{Discussion}

In this work, we introduced the DiSOL framework for geometry-dependent partial differential equations. Rather than approximating a global mapping between continuous function spaces, DiSOL operates directly on discretized computational domains and learns to assemble local solution contributions conditioned on geometry and boundary information. Through a series of benchmark problems—including the Poisson equation, transport-dominated advection-diffusion, vector-valued linear elasticity, and time-dependent heat conduction—we demonstrated that this discrete operator formulation enables robust generalization across geometry-dependent problem settings.

A central insight emerging from our results is that the primary challenge in geometry-dependent PDE learning does not stem from model capacity or optimization, but from the nature of geometric variation itself. Importantly, the degradation of continuous-operator baselines should not be interpreted as a fundamental limitation of continuous neural operator learning. Specifically, in many practical settings, changes in geometry induce discrete and non-smooth structural variations—such as alterations in domain topology, boundary classification, or active computational regions—that fundamentally modify how local contributions are assembled. When such geometry-induced effects dominate problem complexity, the mapping from problem specification to solution field is no longer well described as interpolation within a smooth function-space operator. Continuous operator methods remain highly effective when variations in problem specification, including geometry, can be treated as smooth perturbations of an underlying function-space mapping. This interpretation is further supported by additional benchmarks against representative geometry-aware variants of continuous neural operators, including GNO, Geo-FNO, and DIMON. These methods bring geometric inductive bias through explicit geometric representations or mapping/alignment mechanisms, and can yield modest improvements over vanilla DeepONet/FNO. However, our results show that they do not eliminate the dominant failure modes under strong geometry shifts (see Supplementary Information E), indicating that geometric awareness alone is insufficient when geometry induces discrete structural changes in the instantiated solver structure.

While the governing physical laws and numerical solution procedures remain invariant, geometric changes discretely alter how these procedures are instantiated on a computational domain.  In this sense, what changes is not the underlying algorithmic logic, but the discrete structure through which local contributions are activated, combined, and propagated.  Within this perspective, the advantage of DiSOL does not arise from a particular network architecture, but from its alignment with the discrete nature of geometry-dependent numerical problems.  By operating directly on discretized representations, the learned operator mirrors the structure of classical numerical solvers, explicitly accommodating geometry-induced structural variation without forcing such effects into a smooth function-space interpolation framework.

DiSOL's advantage can be understood through a discrete-operator perspective. The local contribution operator learns reusable, stencil-like transformation rules that map neighborhood-level specifications (geometry, boundary activation, and sources) to local responses, analogous to local discretization stencils shared across the domain. The multiscale assembly pathway then aggregates these local contributions across resolutions, resembling an implicit form of domain decomposition in which coarse features convey long-range coupling while fine features refine boundary-adjacent structure. Under geometry shifts, the procedure (local rules + assembly) remains invariant, while only the set of activated locations changes; this separation helps preserve local validity and enables robust reassembly under new global connectivity and boundary configurations. A more detailed conceptual analysis is provided in Supplementary Information G.

This viewpoint provides a unified explanation for the consistent performance of DiSOL observed across all studied cases, ranging from scalar Poisson equation to transport-dominated dynamics, vector-valued elasticity, and spatiotemporal evolution. In each setting, performance degradation of continuous operator baselines is closely associated with geometry-induced discrete structural changes, rather than with increased physical complexity or stronger coupling alone. From a broader perspective, discrete and continuous operator learning paradigms should be viewed as complementary rather than competing: continuous operators offer elegant representations for problems dominated by smooth parametric variation, whereas discrete operator formulations are more naturally suited to cases in which geometry plays a central and structurally discrete role. The present study is limited to fixed-resolution discrete representations; extending the framework to variable-resolution discretizations (with additional cross-resolution results reported in Supplementary Information E.5), incorporating stricter physical constraints, or developing hybrid formulations that combine discrete and continuous operator principles are promising directions for future work. We also document challenging elasticity regimes where DiSOL exhibits localized artifacts (Supplementary Information F), indicating that stronger boundary-consistent constraints and improved representations may be required in these cases.

\section{Methods}

\subsection{Notation for normalized solutions}
\label{sect:4.1}

In this work, we let $U_h$ denote the ground-truth solution field with physical amplitude on the embedded grid (scalar for Poisson/advection-diffusion/heat conduction problems and vector-valued for elasticity). We define a per-sample amplitude $u_{\lim}$ (Supplementary Information~D, e.g., $u_{\lim}=\max_x |U_h(x)|$) and the dimensionless \emph{solution pattern} $u_h := U_h/u_{\lim}$, so that $\max_x |u_h(x)|=1$. All models (DiSOL and all baselines) are trained to predict the normalized pattern $\hat{u}_h$.
If an absolute-amplitude prediction is required, it can be reconstructed as $\hat{U}_h=\hat{u}_{\lim}\,\hat{u}_h$ using an optional amplitude regressor (Supplementary Sec.~D).

\subsection{Problem setup and discrete learning objective}
\label{sect:4.2}

We formulate geometry-dependent PDE solving as learning a mapping from a discretized problem specification to its (discretized) solution on a fixed Cartesian grid. Each problem instance is represented by a multi-channel tensor $\boldsymbol{X}\in\mathbb{R}^{C_{\mathrm{in}}\times H\times W}$ that encodes (i) a binary geometry mask $m\in\{0,1\}^{H\times W}$ indicating the computational domain $\Omega_h$, (ii) boundary-condition indicators on the embedded grid, and (iii) problem-dependent driving terms (e.g., sources, loads, or IC fields). The goal is to learn an operator $\mathcal{F}_{\theta}$ that predicts the corresponding \emph{normalized solution pattern} (or vector-valued patterns for elasticity) on the same grid
\begin{equation}
    \hat{u}_h=\mathcal{F}_{\theta}(\boldsymbol{X}) \in \mathbb{R}^{C_{\text{out}}\times H\times W}.
\end{equation}

Within each benchmark, the operator is learned and evaluated at a fixed resolution; the grid spacing $h$ denotes the discrete representation scale rather than a refinement parameter. To enforce geometric feasibility and avoid spurious outside-domain predictions, we apply a fixed domain-projection operator $\Pi_m(v)=m\odot v$ ($\odot$ denotes element-wise multiplication) and compute all training losses and evaluation metrics on the projected fields (i.e., restricted to $\Omega_h$). This protocol ensures that comparisons are not influenced by trivial padding effects in inactive regions. Complete PDE specifications and numerical solver settings are provided in Supplementary Information A, and the construction of geometry/condition channels and dataset generation procedures are detailed in Supplementary Information B and C.1.

\subsection{Discrete solution operator learning (DiSOL)}

DiSOL is designed to approximate a \emph{discrete} problem-solving operator by mirroring the compositional structure of classical numerical procedures on an embedded Cartesian grid. In particular, DiSOL factorizes the mapping from a discretized problem specification to its solution into three consecutive operators: (i) a \emph{local contribution operator} that learns geometry- and condition-dependent local interactions, (ii) a \emph{multiscale assembly operator} that aggregates these local contributions across scales to form globally consistent representations on $\Omega_h$, and (iii) a lightweight \emph{problem-solving operator} that reconstructs the target solution pattern from the assembled representation. While DiSOL is implemented using multiscale convolutional feature hierarchies, its architectural intent is to instantiate these three discrete-procedure operators in a differentiable form, rather than to perform generic image-to-image translation.

\paragraph{Local contribution operator.}
DiSOL applies compact convolutional operator blocks, i.e., the local operator $\mathcal{J}_{\theta}$, that act on local neighborhoods and learn stencil-like mappings from the discretized problem specification to latent local contributions. Concretely, given a multi-channel discretized specification $\boldsymbol{X}\in\mathbb{R}^{C_{\mathrm{in}}\times H\times W}$, the local operator outputs a dense field of local contribution features
\begin{equation}
    z_{\ell}=\mathcal{J}_{\theta}(\boldsymbol{X})\in\mathbb{R}^{C_{\ell}\times H\times W},
\end{equation}
where each spatial location encodes geometry- and condition-dependent response information for that cell/pixel. This design emphasizes locality at the level of computation: the global solution behavior is modeled as the composition of locally supported operations whose responses adapt to the encoded geometry and boundary specifications.

\paragraph{Multiscale assembly operator.}
The learned local contributions are then integrated by a multiscale assembly operator that propagates and fuses information from coarse to fine resolutions. Coarser representations capture long-range coupling induced by geometry and boundary constraints, whereas finer representations restore localized structures near boundaries, interfaces, and thin features. Formally, the assembly operator aggregates $z_{\ell}\in\mathbb{R}^{C_{\ell}\times H\times W}$ across scales to produce an assembled, globally consistent representation
\begin{equation}
    z_{a}=\mathcal{A}_{\theta}(z_{\ell})\in\mathbb{R}^{C_{a}\times H\times W},
\end{equation}
in which long-range interactions are integrated through multiresolution pathways while preserving local detail through cross-scale fusion.

\paragraph{Problem-solving operator.}
Given the assembled multiscale representation, DiSOL applies a lightweight problem-solving operator $\mathcal{S}_{\theta}$ to map latent features to the physically meaningful output channels on the embedded grid. Specifically, it converts $z_a\in\mathbb{R}^{C_{a}\times H\times W}$ into the predicted normalized solution pattern
\begin{equation}
    \hat{u}_h=\mathcal{S}_{\theta}(z_a)\in\mathbb{R}^{C_{\mathrm{out}}\times H\times W},
\end{equation}
which can be interpreted as a learned, differentiable surrogate of the final \emph{solve/reconstruction} stage in a discrete procedure.

\paragraph{Auxiliary geometry-consistent mechanisms.}
In addition to the three core operators above, DiSOL incorporates several lightweight mechanisms to improve robustness under geometry shifts. First, \emph{geometry-aware fusion} is implemented via a conditioning branch that processes geometry/boundary channels to generate feature-wise modulation parameters (FiLM) \cite{perez2018film}, which scale and shift intermediate features in the main pathway, enabling geometry-dependent adaptation of local operator responses. Second, \emph{multiscale information gating} injects the geometry mask into cross-scale connections to gate feature propagation and suppress out-of-domain activations, reducing spurious information leakage through inactive regions. Finally, since both inputs and outputs are represented on an embedded Cartesian grid, we apply a fixed domain-projection operator 
\begin{equation}
    \Pi_m(v)=v\odot m.
\end{equation}
By doing so, all losses/metrics are computed on the projected field.

Collectively, the full DiSOL can be written as
\begin{equation}
    \hat{u}_h=\mathcal{F}_{\theta}(\boldsymbol{X})=\Pi_m\!\left[\mathcal{S}_{\theta}\circ\mathcal{A}_{\theta}\circ\mathcal{J}_{\theta}(\boldsymbol{X})\right].
\end{equation}
Architectural details of the local contribution operator, multiscale assembly operator, problem-solving operator, and the auxiliary FiLM/gating mechanisms are provided in Supplementary Information C.2, with model sizes reported in Supplementary Information C.5.

Although DiSOL is implemented with convolutional modules for efficiency on embedded grids, its inductive bias is procedural: it explicitly factors local contribution computation, multiscale assembly, and implicit solving, rather than performing generic image-to-image translation. A controlled U-Net baseline is evaluated under matched capacity and identical training protocols in Supplementary Information~H.

\subsection{Training objective and evaluation metrics}

DiSOL is trained in a supervised manner using solution patterns generated by conventional numerical solvers (Supplementary Information A.5). Let $u$ denote the reference normalized solution pattern and $\hat{u}=\mathcal{F}_{\theta}(\boldsymbol{X})$ the predicted pattern on the embedded grid. To ensure that optimization and evaluation are consistent within the geometry domains, we adopt a masked training objective and compute all errors only on $\Omega_h$ as indicated by the geometry mask $m$. Specifically, we use a masked L1 loss 
\begin{equation}
    \mathcal{L}(\theta)=\|(\hat{u}-u)\odot m\|_1,
\end{equation}
where norm is computed over spatial locations (and channels when applicable). This protocol avoids trivial contributions from inactive regions and provides a consistent basis for comparing models under geometry shifts.

For reporting performance, we primarily use masked relative error metrics computed on $\Omega_h$, including relative L1 (Rel L1) and, when appropriate, relative L2 (Rel L2). In visualizations, we additionally report masked mean absolute error (MAE) for interpretability. Unless stated otherwise, all metrics are evaluated on the projected prediction $\Pi_m(\hat{u})$ and on the corresponding projected reference field $\Pi_m(u)$. Exact metric definitions, including normalization conventions, are provided in Supplementary Information~D, and extended quantitative summaries are reported in Supplementary Information~E.

When absolute-amplitude solutions are required for downstream use, they can be reconstructed via an optional amplitude recovery step (Supplementary Information~D), but this post-processing is not part of the main-text comparative evaluation.

Training is performed with a standard gradient-based optimizer under a unified protocol across DiSOL and all baselines. Optimization settings (optimizer, learning-rate schedule, batch size, and training epochs) are reported in Supplementary Information C.4.

\section{Data availability}
All datasets used in this study are publicly available on Zenodo (DOI: \href{https://doi.org/10.5281/zenodo.18639634}{https://doi.org/10.5281/zenodo.18639634}) \cite{ZenodoDataset}, including the full datasets for all four benchmark cases.

\section{Code availability}
The full source code for DiSOL is available at \href{https://github.com/JinshuaiBai/Discrete-Solution-Operator-Learning-DiSOL}{https://github.com/JinshuaiBai/Discrete-Solution-Operator-Learning-DiSOL}. An executable Code Ocean compute capsule is also provided for reproducibility and ease of use at \href{https://codeocean.com/capsule/4396603/tree/v1}{https://codeocean.com/capsule/4396603/tree/v1}.

\section{Acknowledgements}

Support from the National Natural Science Foundation of China (Grant no. T2488101) is acknowledged (X.-Q. Feng and J. Bai). Support from the National Natural Science Foundation of China (Grant no. 12502234) is acknowledged (J. Bai). 

\bibliographystyle{unsrtnat}
\bibliography{references}  

@Article{DeepONetLu2021,
author={Lu, Lu
and Jin, Pengzhan
and Pang, Guofei
and Zhang, Zhongqiang
and Karniadakis, George Em},
title={Learning nonlinear operators via DeepONet based on the universal approximation theorem of operators},
journal={Nature Machine Intelligence},
year={2021},
month={Mar},
day={01},
volume={3},
number={3},
pages={218-229},
issn={2522-5839},
doi={10.1038/s42256-021-00302-5},
url={https://doi.org/10.1038/s42256-021-00302-5}
}

@article{FNOLi2020,
  title={Fourier neural operator for parametric partial differential equations},
  author={Li, Zongyi and Kovachki, Nikola and Azizzadenesheli, Kamyar and Liu, Burigede and Bhattacharya, Kaushik and Stuart, Andrew and Anandkumar, Anima},
  journal={arXiv preprint arXiv:2010.08895},
  year={2020}
}

@book{ames2014numerical,
  title={Numerical methods for partial differential equations},
  author={Ames, William F},
  year={2014},
  publisher={Academic press}
}

@article{reddy1993introduction,
  title={An introduction to the finite element method},
  author={Reddy, Junuthula Narasimha},
  journal={New York},
  volume={27},
  number={14},
  year={1993}
}

@book{aliabadi2002boundary,
  title={The boundary element method, volume 2: applications in solids and structures},
  author={Aliabadi, Mohammad H},
  volume={2},
  year={2002},
  publisher={John Wiley \& Sons}
}

@book{liu2005introduction,
  title={An introduction to meshfree methods and their programming},
  author={Liu, Gui-Rong and Gu, Yuan-Tong},
  year={2005},
  publisher={Springer}
}

@book{leveque2002finite,
  title={Finite volume methods for hyperbolic problems},
  author={LeVeque, Randall J},
  volume={31},
  year={2002},
  publisher={Cambridge university press}
}

@book{thomas2013numerical,
  title={Numerical partial differential equations: finite difference methods},
  author={Thomas, James William},
  volume={22},
  year={2013},
  publisher={Springer Science \& Business Media}
}

@article{he2024sequential,
  title={Sequential deep operator networks (s-deeponet) for predicting full-field solutions under time-dependent loads},
  author={He, Junyan and Kushwaha, Shashank and Park, Jaewan and Koric, Seid and Abueidda, Diab and Jasiuk, Iwona},
  journal={Engineering Applications of Artificial Intelligence},
  volume={127},
  pages={107258},
  year={2024},
  publisher={Elsevier}
}

@article{li2025architectural,
  title={An architectural analysis of DeepOnet and a general extension of the physics-informed DeepOnet model on solving nonlinear parametric partial differential equations},
  author={Li, Haolin and Miao, Yuyang and Khodaei, Zahra Sharif and Aliabadi, MH},
  journal={Neurocomputing},
  volume={611},
  pages={128675},
  year={2025},
  publisher={Elsevier}
}

@article{yin2024scalable,
  title={A scalable framework for learning the geometry-dependent solution operators of partial differential equations},
  author={Yin, Minglang and Charon, Nicolas and Brody, Ryan and Lu, Lu and Trayanova, Natalia and Maggioni, Mauro},
  journal={Nature computational science},
  volume={4},
  number={12},
  pages={928--940},
  year={2024},
  publisher={Nature Publishing Group US New York}
}

@article{li2023fourier,
  title={Fourier neural operator with learned deformations for pdes on general geometries},
  author={Li, Zongyi and Huang, Daniel Zhengyu and Liu, Burigede and Anandkumar, Anima},
  journal={Journal of Machine Learning Research},
  volume={24},
  number={388},
  pages={1--26},
  year={2023}
}

@article{chen2024learning,
  title={Learning neural operators on riemannian manifolds},
  author={Chen, Gengxiang and Liu, Xu and Meng, Qinglu and Chen, Lu and Liu, Changqing and Li, Yingguang},
  journal={National Science Open},
  volume={3},
  number={6},
  pages={20240001},
  year={2024},
  publisher={China Science Publishing \& Media Ltd. and EDP Sciences}
}

@article{lu2022comprehensive,
  title={A comprehensive and fair comparison of two neural operators (with practical extensions) based on fair data},
  author={Lu, Lu and Meng, Xuhui and Cai, Shengze and Mao, Zhiping and Goswami, Somdatta and Zhang, Zhongqiang and Karniadakis, George Em},
  journal={Computer Methods in Applied Mechanics and Engineering},
  volume={393},
  pages={114778},
  year={2022},
  publisher={Elsevier}
}

@inproceedings{yau1982survey,
  title={Survey on partial differential equations in differential geometry},
  author={Yau, Shing Tung},
  booktitle={Seminar on Differential Geometry},
  volume={102},
  pages={3--71},
  year={1982},
  organization={Princeton University Press Princeton, NJ:}
}

@book{sapiro2006geometric,
  title={Geometric partial differential equations and image analysis},
  author={Sapiro, Guillermo},
  year={2006},
  publisher={Cambridge university press}
}

@article{wang2021learning,
  title={Learning the solution operator of parametric partial differential equations with physics-informed DeepONets},
  author={Wang, Sifan and Wang, Hanwen and Perdikaris, Paris},
  journal={Science advances},
  volume={7},
  number={40},
  pages={eabi8605},
  year={2021},
  publisher={American Association for the Advancement of Science}
}

@article{tran2021factorized,
  title={Factorized fourier neural operators},
  author={Tran, Alasdair and Mathews, Alexander and Xie, Lexing and Ong, Cheng Soon},
  journal={arXiv preprint arXiv:2111.13802},
  year={2021}
}

@inproceedings{bonev2023spherical,
  title={Spherical fourier neural operators: Learning stable dynamics on the sphere},
  author={Bonev, Boris and Kurth, Thorsten and Hundt, Christian and Pathak, Jaideep and Baust, Maximilian and Kashinath, Karthik and Anandkumar, Anima},
  booktitle={International conference on machine learning},
  pages={2806--2823},
  year={2023},
  organization={PMLR}
}

@article{winovich2019convpde,
  title={ConvPDE-UQ: Convolutional neural networks with quantified uncertainty for heterogeneous elliptic partial differential equations on varied domains},
  author={Winovich, Nick and Ramani, Karthik and Lin, Guang},
  journal={Journal of Computational Physics},
  volume={394},
  pages={263--279},
  year={2019},
  publisher={Elsevier}
}

@inproceedings{perez2018film,
  title={Film: Visual reasoning with a general conditioning layer},
  author={Perez, Ethan and Strub, Florian and De Vries, Harm and Dumoulin, Vincent and Courville, Aaron},
  booktitle={Proceedings of the AAAI conference on artificial intelligence},
  volume={32},
  year={2018}
}

@article{DUPREZ2026109131,
title = {$\phi$-FEM-FNO: A new approach to train a Neural Operator as a fast PDE solver for variable geometries},
journal = {Communications in Nonlinear Science and Numerical Simulation},
volume = {152},
pages = {109131},
year = {2026},
issn = {1007-5704},
doi = {https://doi.org/10.1016/j.cnsns.2025.109131},
url = {https://www.sciencedirect.com/science/article/pii/S1007570425005428},
author = {Michel Duprez and Vanessa Lleras and Alexei Lozinski and Vincent Vigon and Killian Vuillemot}
}

@InProceedings{pmlrlong18PDENET,
  title = 	 {{PDE}-Net: Learning {PDE}s from Data},
  author =       {Long, Zichao and Lu, Yiping and Ma, Xianzhong and Dong, Bin},
  booktitle = 	 {Proceedings of the 35th International Conference on Machine Learning},
  pages = 	 {3208--3216},
  year = 	 {2018},
  editor = 	 {Dy, Jennifer and Krause, Andreas},
  volume = 	 {80},
  series = 	 {Proceedings of Machine Learning Research},
  month = 	 {10--15 Jul},
  publisher =    {PMLR},
  url = 	 {https://proceedings.mlr.press/v80/long18a.html}
}

@Article{Bai2025AMS,
author={Bai, Jinshuai
and Wang, Yizheng
and Jeong, Hyogu
and Chu, Shiyuan
and Wang, Qingxia
and Alzubaidi, Laith
and Zhuang, Xiaoying
and Rabczuk, Timon
and Xie, Yi Min
and Feng, Xi-Qiao
and Gu, Yuantong},
title={Towards the future of physics- and data-guided AI frameworks in computational mechanics},
journal={Acta Mechanica Sinica},
year={2025},
month={Jul},
day={09},
volume={41},
number={7},
pages={225340},
issn={1614-3116},
doi={10.1007/s10409-025-25340-x},
url={https://doi.org/10.1007/s10409-025-25340-x}
}

@article{LONG2019108925,
title = {PDE-Net 2.0: Learning PDEs from data with a numeric-symbolic hybrid deep network},
journal = {Journal of Computational Physics},
volume = {399},
pages = {108925},
year = {2019},
issn = {0021-9991},
doi = {https://doi.org/10.1016/j.jcp.2019.108925},
url = {https://www.sciencedirect.com/science/article/pii/S0021999119306308},
author = {Zichao Long and Yiping Lu and Bin Dong}
}

@inproceedings{
zhou2025unisolver,
title={Unisolver: {PDE}-Conditional Transformers Towards Universal Neural {PDE} Solvers},
author={Hang Zhou and Yuezhou Ma and Haixu Wu and Haowen Wang and Mingsheng Long},
booktitle={Forty-second International Conference on Machine Learning},
year={2025},
url={https://openreview.net/forum?id=r1ryQoI9iZ}
}

@Article{Cao2024,
author={Cao, Qianying
and Goswami, Somdatta
and Karniadakis, George Em},
title={Laplace neural operator for solving differential equations},
journal={Nature Machine Intelligence},
year={2024},
month={Jun},
day={01},
volume={6},
number={6},
pages={631-640},
issn={2522-5839},
doi={10.1038/s42256-024-00844-4},
url={https://doi.org/10.1038/s42256-024-00844-4}
}

@Article{NMIEditorial,
title={Machine learning solutions looking for PDE problems},
journal={Nature Machine Intelligence},
author={},
year={2025},
month={Jan},
day={01},
volume={7},
number={1},
pages={1-1},
issn={2522-5839},
doi={10.1038/s42256-025-00989-w},
url={https://doi.org/10.1038/s42256-025-00989-w}
}

@Article{Azizzadenesheli2024,
author={Azizzadenesheli, Kamyar
and Kovachki, Nikola
and Li, Zongyi
and Liu-Schiaffini, Miguel
and Kossaifi, Jean
and Anandkumar, Anima},
title={Neural operators for accelerating scientific simulations and design},
journal={Nature Reviews Physics},
year={2024},
month={May},
day={01},
volume={6},
number={5},
pages={320-328},
issn={2522-5820},
doi={10.1038/s42254-024-00712-5},
url={https://doi.org/10.1038/s42254-024-00712-5}
}

@Article{Brandstetter2025,
author={Brandstetter, Johannes},
title={Envisioning better benchmarks for machine learning PDE solvers},
journal={Nature Machine Intelligence},
year={2025},
month={Jan},
day={01},
volume={7},
number={1},
pages={2-3},
issn={2522-5839},
doi={10.1038/s42256-024-00962-z},
url={https://doi.org/10.1038/s42256-024-00962-z}
}

@article{BAI2024117159,
title = {A robust radial point interpolation method empowered with neural network solvers (RPIM-NNS) for nonlinear solid mechanics},
journal = {Computer Methods in Applied Mechanics and Engineering},
volume = {429},
pages = {117159},
year = {2024},
issn = {0045-7825},
doi = {https://doi.org/10.1016/j.cma.2024.117159},
url = {https://www.sciencedirect.com/science/article/pii/S0045782524004158},
author = {Jinshuai Bai and Gui-Rong Liu and Timon Rabczuk and Yizheng Wang and Xi-Qiao Feng and YuanTong Gu}
}

@book{liu2003smoothed,
  title={Smoothed particle hydrodynamics: a meshfree particle method},
  author={Liu, Gui-Rong and Liu, Moubin B},
  year={2003},
  publisher={World scientific}
}

@book{reddy2005introduction,
  title={An introduction to the finite element method},
  author={Reddy, J. N.},
  volume={3},
  year={2005},
  publisher={McGraw-Hill New York}
}

@book{belytschko2014nonlinear,
  title={Nonlinear finite elements for continua and structures},
  author={Belytschko, Ted and Liu, Wing Kam and Moran, Brian and Elkhodary, Khalil},
  year={2014},
  publisher={John wiley \& sons}
}

@article{eshaghi2025variational,
  title={Variational physics-informed neural operator (VINO) for solving partial differential equations},
  author={Eshaghi, Mohammad Sadegh and Anitescu, Cosmin and Thombre, Manish and Wang, Yizheng and Zhuang, Xiaoying and Rabczuk, Timon},
  journal={Computer Methods in Applied Mechanics and Engineering},
  volume={437},
  pages={117785},
  year={2025},
  publisher={Elsevier}
}

@book{hollig2013approximation,
  title={Approximation and modeling with B-splines},
  author={H{\"o}llig, Klaus and H{\"o}rner, J{\"o}rg},
  year={2013},
  publisher={SIAM}
}

@article{UNetRonnebergerFB15,
  author       = {Olaf Ronneberger and
                  Philipp Fischer and
                  Thomas Brox},
  title        = {U-Net: Convolutional Networks for Biomedical Image Segmentation},
  journal      = {CoRR},
  volume       = {abs/1505.04597},
  year         = {2015},
  url          = {http://arxiv.org/abs/1505.04597},
  eprinttype    = {arXiv},
  eprint       = {1505.04597},
  bibsource    = {dblp computer science bibliography, https://dblp.org}
}

@inproceedings{hu2018squeeze,
  title={Squeeze-and-excitation networks},
  author={Hu, Jie and Shen, Li and Sun, Gang},
  booktitle={Proceedings of the IEEE conference on computer vision and pattern recognition},
  pages={7132--7141},
  year={2018}
}

@article{li2020neural,
  title={Neural operator: Graph kernel network for partial differential equations},
  author={Li, Zongyi and Kovachki, Nikola and Azizzadenesheli, Kamyar and Liu, Burigede and Bhattacharya, Kaushik and Stuart, Andrew and Anandkumar, Anima},
  journal={arXiv preprint arXiv:2003.03485},
  year={2020}
}

@dataset{ZenodoDataset,
  author = {Bai, Jinshuai
  and Li, Haolin
  and Sharif Khodaei, Zahra
  and Aliabadi, M.H.
  and Gu, YuanTong
  and Feng, Xi-Qiao},
  year = {2026},
  publisher = {Zenodo},
  doi = {10.5281/zenodo.18639634},
  url = {https://doi.org/10.5281/zenodo.18639634}
}






\end{document}


\maketitle

\newpage
\tableofcontents

\clearpage
\section{Governing equations, boundary conditions, and numerical solvers}
\label{sect:1}

In this work, all learning tasks are formulated as operator mappings between discretized physical problem instances and their corresponding numerical solutions. For completeness and reproducibility, we summarize below the governing equations, boundary conditions, and numerical solvers used to generate the ground-truth datasets for all experiments.

\paragraph{Notation (consistent with the main text).}
Throughout this Supplementary Information, $U$ denotes the ground-truth solution field with physical amplitude, and $u$ denotes the corresponding normalized solution \emph{pattern}. For each sample we define a scalar amplitude $u_{\text{lim}}$ (e.g., $u_{\text{lim}}=\max_x |U(x)|$ on $\Omega_h$) and set $u := U/u_{\text{lim}}$. Model predictions are denoted with a hat: $\hat{u}$ for the predicted pattern and, when applicable, $\hat{u}_{\lim}$ for
the predicted amplitude, so that $\hat{U}=\hat{u}_{\lim}\hat{u}$.

\subsection{Poisson equation}
\label{sect:1.1}

We consider the scalar Poisson problem defined on a bounded domain $\Omega \subset \mathbb{R}^2$:
\begin{equation}
	\Delta U(\textbf{x})=f(\textbf{x}).
\end{equation}
\textbf{Boundary conditions.} The boundary $\partial \Omega$ is partitioned into Dirichlet boundary $\Gamma_{\text{D}}$ and a Neumann boundary $\Gamma_{\text{N}}$, such that
\begin{equation}
\begin{aligned}
        U(\textbf{x})=0,\quad \textbf{x} \in \Gamma_{\text{D}}, \\
        \nabla U(\textbf{x}) \cdot \textbf{n}=0,\quad \textbf{x} \in \Gamma_{\text{N}},
\end{aligned}
\end{equation}
where $\textbf{n}$ denotes the outward unit normal vector.

The Dirichlet boundary $\Gamma_{\text{D}}$ is explicitly specified as part of the input, while the homogeneous Neumann boundary condition on $\Gamma_{\text{N}} = \partial \Omega \backslash \Gamma_{\text{D}}$ is implicitly enforced by the weak formulation.

\subsection{Steady-state advection-diffusion equation}
\label{sect:1.2}

We consider the steady-state advection-diffusion equation
\begin{equation}
    \nabla \cdot (\kappa\nabla U(\textbf{x})-\textbf{v}U(\textbf{x}))=f(\textbf{x}),    
\end{equation}
where $\kappa > 0$ is the diffusion coefficient and $\textbf{v}=(v_x, v_y)$ is a constant advection velocity. As introduced in the main manuscript, the diffusion coefficient $\kappa$ and advection velocities $\textbf{v}$ are fixed. For two different Peclet number (Pe) studies, the parameters are as follow:
\begin{equation}
    \begin{aligned}
        \kappa &= 1.0,\quad\textbf{v}=(0.25,0.5),\qquad\text{($\text{Pe}\approx0.45$)} \\
        \kappa &= 0.2,\quad\textbf{v}=(0.5,-1.0),\qquad\text{($\text{Pe}\approx4.5$)}
    \end{aligned}
\end{equation}
\textbf{Boundary conditions.} The boundary $\partial \Omega$ is partitioned into Dirichlet boundary $\Gamma_{\text{D}}$ and a Neumann boundary $\Gamma_{\text{N}}$, such that
\begin{equation}
    \begin{aligned}
        U(\textbf{x})=0,\quad \textbf{x} \in \Gamma_{\text{D}}, \\
        \kappa \nabla U(\textbf{x}) \cdot \textbf{n}=0,\quad \textbf{x} \in \Gamma_{\text{N}}.
    \end{aligned}
\end{equation}
Here, homogeneous Neumann conditions correspond to zero diffusive flux across $\Gamma_{\text{N}}$. All advection-diffusion experiments adopt explicitly prescribed zero Dirichlet boundary segments and implicitly enforced homogeneous Neumann conditions elsewhere.

\subsection{Linear elasticity}
\label{sect:1.3}

We consider small-strain, linear elasticity under quasi-static conditions. The displacement field $\textbf{U}=(U_x,U_y):\Omega \to \mathbb{R}^2$ satisfies
\begin{equation}
    -\nabla \cdot \boldsymbol{\sigma}(\textbf{U})=\textbf{f}(\textbf{x}),\quad\textbf{x} \in \Omega,
\end{equation}
with the constitutive relation
\begin{equation}
    \begin{aligned}
        \boldsymbol{\sigma}(\textbf{U})&=\lambda \text{tr}(\boldsymbol{\epsilon}(\textbf{U}))\textbf{I}+2\mu \boldsymbol{\epsilon}(\textbf{U}), \\
        \boldsymbol{\epsilon}(\textbf{U})&=\frac{1}{2}(\nabla \textbf{U} + \nabla \textbf{U}^{\top}),
    \end{aligned}
\end{equation}
where $\lambda$ and $\mu$ are Lamé parameters, and are computed by
\begin{equation}
    \begin{aligned}
        \lambda &= \frac{E\nu}{(1+\nu)(1-2\nu)},\\
        \mu &= \frac{E}{2(1+\nu)},
    \end{aligned}
\end{equation}
where $E$ is the Young's modulus and $\nu$ denotes the Poisson's ratio. These two parameters are fixed as $E = 1\times 10^3$ and $\nu = 0.3$.

\textbf{Boundary conditions.} The boundary is decomposed into displacement and traction boundaries:
\begin{equation}
    \begin{aligned}
        \textbf{U}(\textbf{x})&=0,\qquad \textbf{x} \in \Gamma_{\text{D}}, \\
        \boldsymbol{\sigma}(\textbf{U}) \cdot \textbf{n} &= \textbf{t}(\textbf{x}),\quad \textbf{x} \in \Gamma_{\text{N}}.
    \end{aligned}
\end{equation}
Unlike the scalar problems, the Neumann boundary in elasticity may involve non-zero applied tractions. Both $\Gamma_{\text{D}}$ and $\Gamma_{\text{N}}$, as well as the traction field $\textbf{t}(\textbf{x})$, are explicitly encoded as input channels.

\subsection{Time-dependent heat equation}
\label{sect:1.4}

We consider the transient heat conduction problem
\begin{equation}
    \frac{\partial U(\textbf{x},t)}{\partial t} - \nabla \cdot (\kappa \nabla U(\textbf{x},t))=f(\textbf{x},t),\quad\textbf{x} \in \Omega,\quad t\in [0,T],
\end{equation}
where $\kappa$ is coefficient of thermal conductivity and is fixed as $\kappa =1.0$.

\textbf{Initial and boundary conditions.}
\begin{equation}
    \begin{aligned}
        U(\textbf{x},0)&=U_0(\textbf{x}),\\
        U(\textbf{x},t)&=0, \textbf{x} \in \Gamma_{\text{D}}, \\
        \kappa \nabla U(\textbf{x},t) \cdot \textbf{n} &= 0, \textbf{x} \in \Gamma_{\text{N}}
    \end{aligned}
\end{equation}
where $\kappa$ is the coefficient of conduction in this case. As in the scalar problems, Dirichlet boundaries are explicitly specified, while homogeneous Neumann conditions are implicitly enforced.

\subsection{Numerical discretization and solvers}
\label{sect:1.5}

All ground-truth solutions are generated using the finite element method (FEM) \cite{reddy2005introduction}.
Unless otherwise stated, the following settings are adopted:

\begin{itemize}
    \item \textbf{Spatial discretization}: conforming $P^1$  (linear) finite elements on structured Cartesian meshes.

    \item \textbf{Weak formulation}: standard Galerkin formulation.

    \item \textbf{Boundary treatment}: Dirichlet conditions are strongly imposed; homogeneous Neumann conditions arise naturally from the weak form.

    \item \textbf{Linear solvers}: sparse direct solvers are used for all steady-state problems.

    \item \textbf{Time integration (heat equation)}: implicit backward Euler scheme with a fixed time step.
\end{itemize}

The mesh resolution used for data generation is fixed across all training, validation, and test datasets, ensuring that all operator learning tasks are performed at a consistent discrete resolution.

\paragraph{Domain normalization.} For all experiments, the computational domain is defined on a normalized square domain $\Omega \subset [0,1]^2$, regardless of the grid resolution.
All spatial coordinates, geometric constructions, source terms, and boundary conditions are defined in this normalized coordinate system.
Different image resolutions therefore correspond only to different discrete samplings of the same physical domain, rather than changes in physical length scales.

\clearpage
\section{Data generation}
\label{sect:2}
\subsection{Geometry generation}
\label{sect:2.1}

All computational domains used in this work are generated procedurally to ensure controlled geometric variability while maintaining strict consistency between training and test distributions.

\paragraph{Single-connected smooth domains for training.} For all training datasets, the computational domain $\Omega$ is restricted to be single-connected with a smooth outer boundary. The geometry generation procedure is as follows:

\begin{itemize}
    \item [1)]
    A fixed number ($N=20$) of control points is randomly sampled in the two-dimensional plane;
    \item [2)]
    A boundary extraction algorithm is applied to identify an ordered subset of points that form a closed envelope enclosing the sampled points.
    \item [3)]
    The extracted boundary points are used as control nodes of a closed B-spline curve \cite{hollig2013approximation}, yielding a smooth, non-self-intersecting boundary.
    \item [4)]
    The resulting continuous geometry is rasterized onto a structured Cartesian grid, producing a binary geometry mask:
    \begin{equation}
        m(\textbf{x})=\left\{
        \begin{aligned}
            1,\quad \textbf{x} \in \Omega,\\
            0,\quad \textbf{x} \notin \Omega.
        \end{aligned}
        \right.
    \end{equation}
\end{itemize}

This procedure guarantees that all training geometries are topologically simple (single-connected) and free of sharp corners or discontinuities, while still exhibiting substantial shape variability.

\begin{figure}
    \centering
    \includegraphics[width=0.7\linewidth]{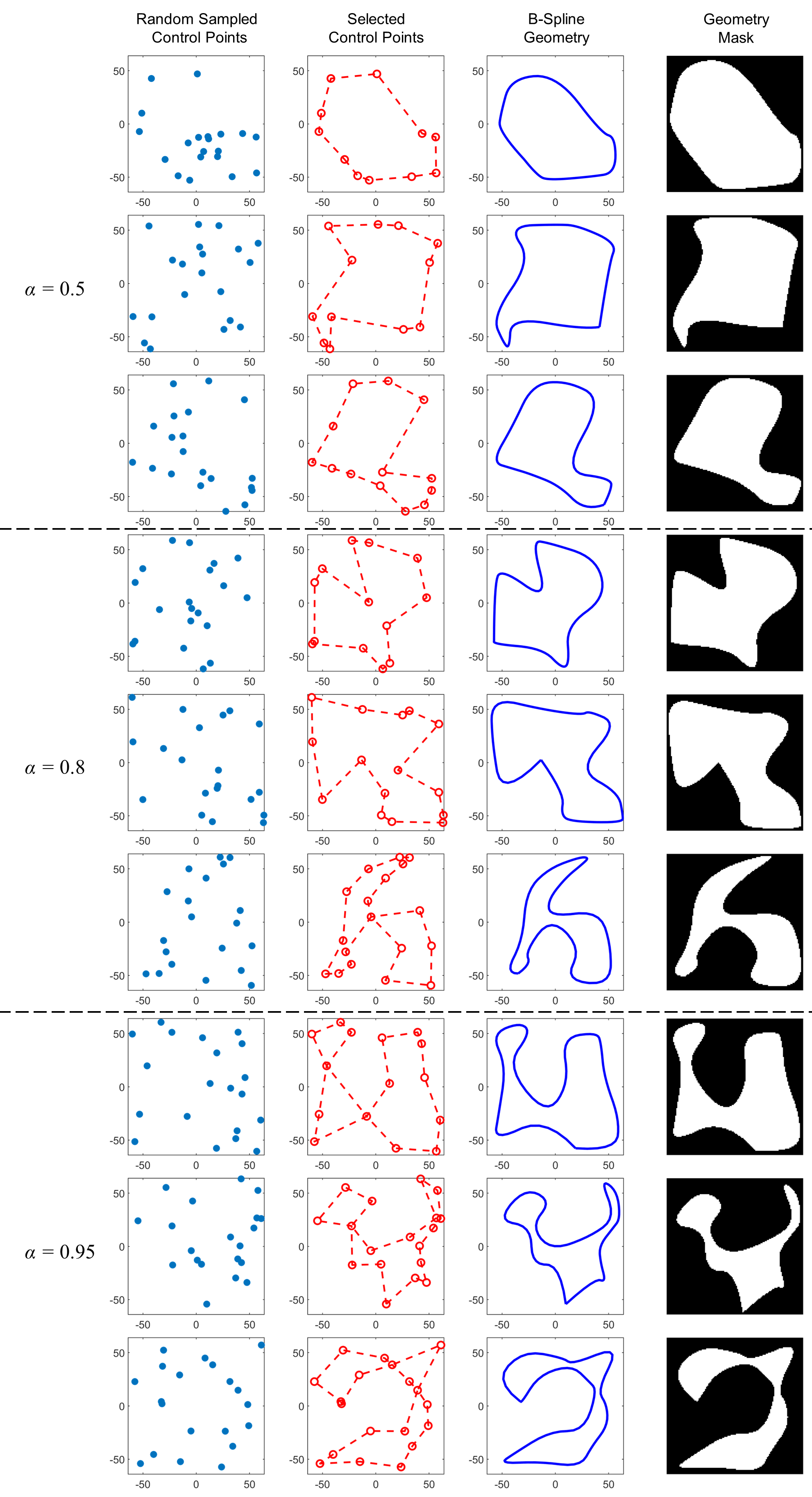}
    \caption{\textbf{Geometry generation pipeline for training and its role as the base configuration for OOD geometries.} Training geometries are generated as single-connected domains with smooth boundaries by sampling random control points, extracting an enclosing boundary, and fitting a closed B-spline curve, which is subsequently rasterized into a binary geometry mask. All training geometries are free of internal holes and topological discontinuities. Out-of-distribution (OOD) geometries used in subsequent experiments are constructed by introducing internal holes into these base geometries, while keeping the outer boundary generation procedure unchanged.}
    \label{fig:s1}
\end{figure}

\paragraph{Control of geometric complexity.} The complexity of the generated geometries is controlled by a single scalar parameter $\alpha \in (0,1)$, which governs the tightness of the boundary envelope around the sampled control points. Larger values of $\alpha$ produce geometries with more intricate boundary features and finer-scale structures, while smaller values yield simpler, more convex shapes, as shown in \textbf{Fig.~}\ref{fig:s1}.

Unless otherwise stated, the training datasets are generated using $\alpha = 0.8$, ensuring smooth boundaries without thin protrusions or near-disconnected features.

\paragraph{Rasterization and resolution.} All geometries are embedded into a normalized physical domain of unit length, with the image resolution controlling only the discrete sampling density. Accordingly, parameters defined relative to the image size (e.g., hole size fractions) correspond to relative geometric scales within the same physical domain:

\begin{itemize}
    \item \textbf{Poisson equation}: $64 \times 64$ grid
    \item \textbf{All other problems} $128 \times 128$ grid
\end{itemize}

The binary geometry mask is used consistently across all problems to define the computational domain and to mask loss evaluation and error metrics.

Importantly, the grid resolution is fixed within each task and is not treated as a refinement parameter. All operator learning tasks are therefore conducted on a consistent discrete representation of the domain.

\paragraph{Reproducibility.} The geometry generation process is fully deterministic given a random seed and the parameter $\alpha$. All datasets are generated offline prior to training, and the same procedure is applied consistently across training, validation, and test sets, with differences between ID and OOD cases introduced only through controlled variations described in subsequent sections.

\subsection{Boundary condition selection and encoding }
\label{sect:2.2}

For all problems involving Dirichlet boundary conditions, the boundary segments are selected procedurally from the outer boundary of the generated geometry. 

\paragraph{Boundary extraction.} Given a binary geometry mask defining the computational domain $\Omega$, the outer boundary $\partial \Omega$ is first identified using a standard boundary tracing algorithm applied to the binary mask. The resulting boundary is represented as an ordered sequence of discrete boundary pixels forming a closed loop. 

Only the outer boundary is considered at this stage; internal boundaries associated with holes are introduced separately in out-of-distribution (OOD) test cases. 

\paragraph{Random selection of continuous boundary segments.} From the extracted boundary, a single continuous boundary segment is randomly selected to serve as the Dirichlet boundary $\Gamma_{\text{D}}$. The length of the selected segment is controlled by a scalar parameter $l \in (0,1]$, which specifies the maximum fraction of the total boundary length that can be assigned as the Dirichlet boundary.

Specifically, the number of boundary pixels included in $\Gamma_{\text{D}}$ is bounded by $l\vert\partial\Omega\vert$, and the selected pixels form a contiguous segment along the boundary. This ensures that Dirichlet boundary conditions are spatially localized and geometrically consistent across different domain shapes.

\paragraph{Binary encoding of boundary conditions.} The selected Dirichlet boundary segment is encoded as a binary image channel:
\begin{equation}
    b(\textbf{x})=\left\{
    \begin{aligned}
        1&, \textbf{x} \in \Gamma_{\text{D}},\\
        0&, \text{otherwise},
    \end{aligned}
    \right.
\end{equation}
which is provided as part of the model input. All remaining boundary pixels are implicitly treated as Neumann boundary $\Gamma_{\text{N}}=\partial \Omega \backslash \Gamma_{\text{D}}$
, on which homogeneous Neumann conditions are enforced through the weak formulation of the governing equations.

\paragraph{Consistency across datasets.} The same boundary selection and encoding procedure is applied consistently across training, validation, and test datasets. Differences between in-distribution and out-of-distribution cases arise only from controlled variations in geometric topology or boundary complexity, as described in subsequent sections.

\subsection{Source terms and external loads}
\label{sect:2.3}

For all scalar-field problems considered in this work, including the Poisson equation, the steady-state advection-diffusion equation, and the transient heat equation, a spatially varying source term is introduced to drive the system response. Initial conditions for time-dependent problems are described separately and omitted here for clarity.

\paragraph{Source term generation for scalar problems.} The source term $f(x,y)$ is constructed as a superposition of Fourier and Gaussian basis functions with randomly sampled parameters. Specifically, an unnormalized source field $\hat f(x,y)$ is defined as
\begin{equation}
        \hat f(x,y)=\sum_i^{n_\text{F}} \xi_i\phi_i(x,y) + \sum_j^{n_\text{G}} \xi_j\psi_j(x,y),\quad(x,y) \in \Omega,
\end{equation}
where
\begin{equation}
    \begin{aligned}
        \phi_i(x,y)&=\sin{(2\pi(R_{1,i}x+R_{2,i}y)+R_{3,i})}, \\ 
        \psi_i(x,y)&=\exp(-\frac {(x-R_{1,j})^2+(y-R_{2,j})^2}{2R_{3,j}^2}).
    \end{aligned}
\end{equation}
Here, $n_\text{F}$ and $n_\text{G}$ denote the numbers of Fourier and Gaussian basis functions, respectively, and all parameters $R_{\cdot,\cdot}$ are sampled randomly. For the Fourier components, the sine function may be replaced by a cosine function with equal probability.

The coefficients {$\xi_i$} and {$\xi_j$} are normalized such that
\begin{equation}
    \sum_{i=1}^{n_\text{F}}\xi_i + \sum_{j=1}^{n_\text{G}}\xi_j = 1,
\end{equation}
ensuring a consistent magnitude across different realizations.

\paragraph{Source term normalization.} To further eliminate scale variations, the source term is normalized to the unit interval:
\begin{equation}
    f(x,y)=\frac{\hat f(x,y)-\min_\Omega \hat f(x,y)}{\max_\Omega \hat{f}(x,y) - \min_\Omega \hat{f}(x,y) },\quad(x,y) \in \Omega.
\end{equation}
This normalization is applied independently for each sample and ensures that all source fields lie within a comparable dynamic range. The resulting scalar field $f(x,y)$ is provided to the model as an input channel, masked by the geometry indicator to ensure that values outside the computational domain are ignored. Note that this input normalization of $f$ is independent of the output normalization of the solution pattern field described in Supplementary Information~\ref{sect:4}.

\paragraph{External loads for linear elasticity.} For the linear elasticity problem, external forcing is applied in the form of boundary nodal forces rather than volumetric source terms.

A continuous boundary segment is first selected using the procedure described in Supplementary Information \ref{sect:2.2}. Forces are then applied uniformly along the selected boundary segment, with a unit magnitude assigned to each boundary node. The force direction is specified independently in the horizontal and vertical directions.

Specifically, two force components $f_x$ and $f_y$ are defined as separate input channels:
\begin{equation}
    f_x(\textbf{x}), f_y(\textbf{x})\in \{-1,0,1\},
\end{equation}
where non-zero values appear only on the selected boundary nodes. The sign indicates the force direction, while all interior nodes and boundary nodes not selected for loading are assigned zero values.

This formulation allows for flexible combinations of loading configurations while maintaining a consistent and sparse representation of external forces.

\paragraph{Consistency across datasets.} The source term and load generation procedures described above are applied consistently across training, validation, and test datasets. For scalar problems, the source fields are independent of the geometry and boundary selection, while for elasticity, the loading locations are directly tied to the selected boundary segments.

Out-of-distribution cases may involve additional geometric complexity or topological changes, but the functional forms and normalization of source terms and external loads remain unchanged.

\subsection{Initial condition} 

The transient heat equation requires initial conditions as an additional input channel to the model. The random initial condition field is generated using the same strategy as the source field.

\subsection{Data size and split}

Following the data-generation protocol described above, we construct datasets for five benchmark problems.
The total numbers of samples are: Poisson equation ($N_{\mathrm{total}}=46{,}011$), advection--diffusion with low P\'eclet number ($N_{\mathrm{total}}=49{,}650$), advection--diffusion with high P\'eclet number ($N_{\mathrm{total}}=49{,}400$), linear elasticity ($N_{\mathrm{total}}=49{,}788$), and time-dependent heat conduction ($N_{\mathrm{total}}=10{,}000$).
For each benchmark, we adopt an $8{:}2$ split into training and held-out test sets:
$N_{\mathrm{train}}=\lfloor 0.8\,N_{\mathrm{total}}\rfloor$ and $N_{\mathrm{test}}=N_{\mathrm{total}}-N_{\mathrm{train}}$.
Concretely, this yields
Poisson: $N_{\mathrm{train}}=36{,}808$, $N_{\mathrm{test}}=9{,}203$;
low-Pe advection--diffusion: $N_{\mathrm{train}}=39{,}720$, $N_{\mathrm{test}}=9{,}930$;
high-Pe advection--diffusion: $N_{\mathrm{train}}=39{,}520$, $N_{\mathrm{test}}=9{,}880$;
linear elasticity: $N_{\mathrm{train}}=39{,}830$, $N_{\mathrm{test}}=9{,}958$;
time-dependent heat conduction: $N_{\mathrm{train}}=8{,}000$, $N_{\mathrm{test}}=2{,}000$.
Unless otherwise stated, all reported metrics are computed per sample and then aggregated over the corresponding split.

\subsection{Construction of out-of-distribution (OOD) geometries via hole insertion}
\label{sect:2.4}

Out-of-distribution (OOD) test cases are constructed by introducing controlled deviations from the training data distribution along three complementary dimensions: geometric topology, boundary condition complexity, and source-term frequency content.

\paragraph{Topological variation via hole insertion.} OOD geometries are generated by introducing internal holes into the single-connected base geometries described in Supplementary Information \ref{sect:2.1}.

For each base geometry with characteristic length $L = 1$ (corresponding to Supplementary Information \ref{sect:1.5}), an internal void is generated using the same procedural geometry generation pipeline, but with modified parameters to ensure a smaller characteristic scale. The effective size of the void is determined as a fraction of the domain size,
\begin{equation}
    L_{\text{void}}=[0.2L,0.4L],
\end{equation}
with the exact value sampled randomly.

The internal void is generated using a reduced number of control points ($N=10$) and a randomly sampled boundary complexity parameter $\alpha \in [0.2,0.85]$. The final OOD geometry is obtained by subtracting the void from the base geometry. Geometries that violate basic validity criteria (e.g., disconnected outer domains or degenerate boundaries) are discarded.

This procedure yields multiply connected domains with smooth outer boundaries and internal boundaries of comparable geometric regularity.

\paragraph{Internal boundary condition selection.} For OOD cases with internal holes, Dirichlet boundary conditions may be applied not only to the outer boundary but also to internal boundaries.

Boundary segments are selected independently on each boundary component using the same continuous segment selection procedure described in Supplementary Information \ref{sect:2.2}. To increase boundary complexity, multiple boundary segments may be selected, and the maximum allowable segment length is reduced to enforce shorter and more fragmented Dirichlet boundary regions.

This results in boundary configurations that are both topologically and spatially more complex than those encountered during training.

\paragraph{High-frequency source term generation.} In addition to geometric and boundary variations, OOD cases include source terms with elevated frequency content.

For training (in-distribution) datasets, the numbers of Fourier and Gaussian basis functions are sampled independently from the range $[0,3]$. The parameters of the Fourier components satisfy
\begin{equation}
    R_{1,i},R_{2,i} \in (0,1),\quad R_{3,i} \in (-\frac{\pi}{4},\frac{\pi}{4}),\qquad\text{(training)}
\end{equation}
corresponding to low-frequency spatial variations.

For high-frequency OOD cases, the numbers of Fourier and Gaussian components are increased and sampled from the range $[0,7]$. The Fourier parameters are drawn from
\begin{equation}
    R_{1,i},R_{2,i} \in (0,4),\quad R_{3,i} \in (-\pi,\pi),\qquad\text{(high-frequency OOD)}
\end{equation}
leading to substantially higher spatial frequency content.

For Gaussian components, the center locations $R_{1,j},R_{2,j}$ are sampled from the same spatial range $[0,1]$ in both training and OOD settings. However, the characteristic width parameter differs:
\begin{equation}
    \begin{aligned}
        R_{3,j} &\in (0.5,1.5),\quad \text{(training)}\\
        R_{3,j} &\in (0, 0.4),\qquad \text{(high-frequency OOD)}
    \end{aligned}
\end{equation}
resulting in more localized source patterns in OOD cases.

\paragraph{Summary of OOD perturbations.} Across all OOD test cases, the governing equations, numerical solvers, and input-output representations remain unchanged. Only the geometric topology, boundary condition complexity, and source-term frequency content are altered in a controlled manner.

This design allows for a systematic evaluation of the robustness of learned operators under distribution shifts that are common in practical scientific computing scenarios.

\clearpage
\section{Detailed architecture and training settings}

\subsection{Input-output representation}
\label{sect:3.1}

All learning tasks in this work are formulated as image-to-image operator mappings on fixed-resolution Cartesian grids. Each problem is represented using a multi-channel input tensor that encodes geometry, boundary conditions, and problem-specific driving terms, while the output corresponds to the target normalized solution pattern field $u_h$.

\paragraph{Scalar-field problems.} For scalar-field problems, including the Poisson equation, the steady-state advection-diffusion equation, and the transient heat equation, the model input consists of the following channels:

\begin{itemize}
    \item [1)] \textbf{Geometry mask}: a binary indicator of the computational domain, with value 1 inside the domain and 0 outside.
    \item [2)] \textbf{Dirichlet boundary mask}: a binary channel indicating the selected Dirichlet boundary segments.
    \item [3)] \textbf{Source term}: a scalar field representing the volumetric forcing term.
\end{itemize}

For time-dependent problems, an additional channel is provided:

\begin{itemize}
    \item [4)] \textbf{Initial condition (IC)}: the scalar field at the initial time, defined within the geometry mask.
\end{itemize}

The model output for these problems is a single-channel scalar field corresponding to the solution variable (e.g., potential, concentration, or temperature).

\paragraph{Linear elasticity.} For the linear elasticity problem, the input representation is adapted to accommodate vector-valued boundary conditions and outputs. The input tensor includes:

\begin{itemize}
    \item [1)] \textbf{Geometry mask}: a binary indicator of the solid domain.
    \item [2)] \textbf{Fixed boundary conditions}: two channels corresponding to zero-displacement (Dirichlet) constraints in the horizontal and vertical directions.
    \item [3)] \textbf{Boundary forces}: two channels corresponding to the applied Neumann boundary forces in the horizontal and vertical directions.
\end{itemize}

The model output consists of two channels representing the horizontal and vertical displacement components.

\paragraph{Masking and loss evaluation.} All models are trained using a global L1 loss computed over the full image domain. To ensure that predictions are physically meaningful only within the computational domain, the network outputs are multiplied element-wise by the geometry mask prior to loss evaluation.

As a result, predictions outside the geometry are identically zero and do not contribute to the loss, effectively restricting supervision to the interior of the domain while maintaining a consistent image-based formulation.

\subsection{Discrete Solution Operator Learning (DiSOL) architecture}
\label{sect:3.2}
\begin{figure}[h]
    \centering
    \includegraphics[width=0.8\linewidth]{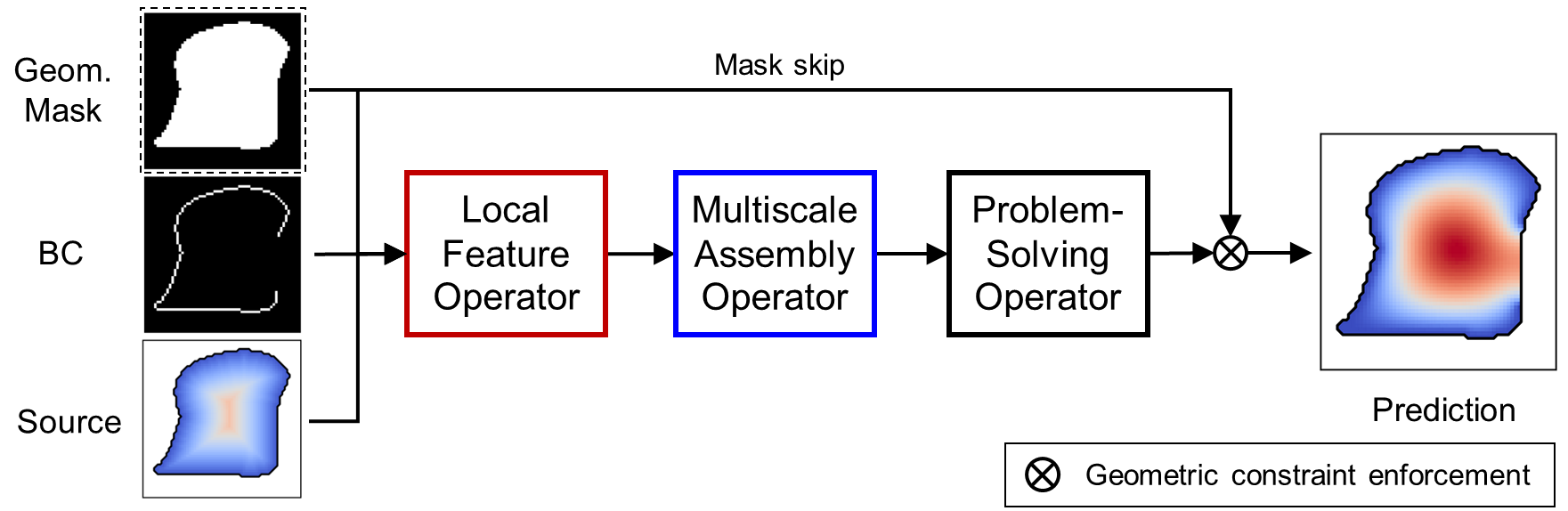}
    \caption{\textbf{Schematic illustration of the DiSOL architecture used in the ablation study.} DiSOL factorizes the solution operator into a geometry-aware local feature operator, a multiscale assembly operator (implemented using a UNet backbone), and a discrete solution readout. An explicit output mask is applied as a feasibility projection to enforce geometric constraints, and is treated as a structural component rather than a training heuristic.}
    \label{fig:s_DiSOL_struct}
\end{figure}

The proposed DiSOL model is designed as a fully convolutional, image-based discrete operator that maps problem-specific input fields to solution pattern fields on a fixed-resolution grid. The architecture emphasizes locality, multi-scale information assembly, and explicit conditioning on geometry and boundary information. A schematic overview of the DiSOL operator decomposition and the role of output masking is provided in \textbf{Fig.} \ref{fig:s_DiSOL_struct}.

\paragraph{Overall architecture.} DiSOL adopts a UNet-like encoder-decoder structure \cite{UNetRonnebergerFB15} with skip connections to aggregate local and multi-scale features. The same backbone architecture is used across all spatial problems, including Poisson, advection-diffusion, and linear elasticity, with task-specific input and output channels.

The network consists of:

\begin{itemize}
    \item [1)] An input embedding stage that separates geometry-related and physics-related channels.
    \item [2)] A multi-scale encoder-decoder backbone that performs local operator learning and information aggregation.
    \item [3)] A lightweight output head that maps latent features to the target solution pattern fields.
\end{itemize}

\paragraph{Local feature operator.} At each resolution level, DiSOL employs lightweight local operator blocks composed of pointwise convolutions, normalization, nonlinearity, and gated feature modulation. Specifically, each block follows a residual formulation to promote stable training and preserve locality.

Channel-wise interactions are enhanced through gated pointwise transformations and optional squeeze-and-excitation mechanisms \cite{hu2018squeeze}, allowing the network to adaptively emphasize physically relevant features while maintaining a compact parameterization.

\paragraph{Geometry-aware conditioning via FiLM.} To explicitly incorporate geometric and boundary information, DiSOL uses feature-wise linear modulation (FiLM) \cite{perez2018film}. Geometry- and boundary-related channels are processed through a dedicated branch and globally pooled to form a low-dimensional conditioning vector.

This conditioning vector is used to generate spatially invariant modulation parameters that scale and shift intermediate feature maps:
\begin{equation}
    \textbf{z}'=(1+\gamma)\odot \textbf{z} + \beta,
\end{equation}
where $\gamma$ and $\beta$ are learned functions of the geometry and boundary configuration.

This design enables DiSOL to adapt its local operator behavior based on global geometric context while preserving translational equivariance within the domain.

\paragraph{Multi-scale assembly operator.} Multi-scale information is aggregated through the encoder-decoder pathway using strided convolutions for downsampling and interpolation-based upsampling. Skip connections between corresponding resolution levels ensure that fine-scale geometric details and boundary information are retained in the reconstruction process.

This multi-scale assembly allows DiSOL to capture both localized effects near boundaries and global structural responses induced by geometry and forcing.

\paragraph{Problem-solving operator.} For scalar-field problems, the network outputs a single-channel solution pattern field.
For linear elasticity, the output consists of two channels corresponding to the horizontal and vertical displacement components.

\paragraph{Mask-skip routing (geometry-consistent feature gating).} 
Geometry-dependent PDEs are posed on an embedded Cartesian grid, where only the pixels inside the active domain $\Omega_h$ are physically meaningful. Beyond applying output masking for feasibility, DiSOL further introduces a \emph{mask-skip} mechanism that injects the geometry mask into the skip pathways to gate multi-scale feature propagation. Concretely, at each resolution level, the skip features are modulated by the corresponding downsampled mask, suppressing out-of-domain activations and preventing spurious feature leakage across inactive regions. This geometry-consistent routing aligns the internal information flow with the discrete computational structure induced by $\Omega_h$, reducing the burden on the network to implicitly learn domain support and improving stability under geometry shifts.

\subsection{Baseline architectures}
\label{sect:3.3}

To evaluate the performance of DiSOL, we compare against five representative neural-operator baselines: DeepONet and the Fourier Neural Operator (FNO), as well as three geometry-aware operator-learning methods:
\begin{itemize}
    \item Diffeomorphic Mapping Operator Learning (DIMON) \cite{yin2024scalable},
    \item Geo-FNO \cite{li2023fourier},
    \item Graph Neural Operator \cite{li2020neural},
\end{itemize}
which augment standard operators with explicit geometric mappings. All baselines are implemented following their original formulations and are trained under the same data representation, loss functions, evaluation protocol, and optimization settings as DiSOL (unless otherwise stated), to ensure fair comparisons.

\paragraph{DeepONet.} \cite{DeepONetLu2021} DeepONet is implemented using a standard branch-trunk architecture. The branch network processes the discretized input PDE dependencies, while the trunk network encodes spatial coordinates.

To enable a fair comparison with image-based operator learning methods, the branch network receives the same input channels as DiSOL and FNO, flattened into a vector representation. The trunk network takes normalized spatial coordinates as input.

The outputs of the branch and trunk networks are combined through an inner-product operation to produce the predicted solution pattern field at each spatial location.

Network depth and width for both branch and trunk components are selected according to the configuration files and are kept fixed for each problem setting.

\paragraph{\textbf{Diffeomorphic Mapping Operator Learning (DIMON).}} DIMON \cite{yin2024scalable} is a geometry-aware neural-operator framework that handles PDE learning on varying geometries by introducing an explicit diffeomorphic mapping between each physical domain and a shared template (reference) domain. The key idea is to use change-of-variables so that the PDE on a geometry-dependent domain can be consistently represented and learned on a fixed template domain, with the learned/constructed diffeomorphism accounting for geometric variability. In this work, we include DIMON as a representative mapping-based operator-learning baseline and follow the formulation and implementation described in the original paper \cite{yin2024scalable} and the authors’ open-source code. Notably, DIMON assumes that the family of domains is diffeomorphic to the template; thus, topology-changing domain shifts (e.g., the OOD tests used in this work) fall outside the original applicability of DIMON. 

In this work, the geometry is represented using PCA of a displacement field computed between a reference circle and the target shape, where corresponding boundary points are established via a concentric-circle mapping. The $x$- and $y$-components of the displacement field are decomposed separately, yielding a compact shape descriptor $[\mathbf{c}_x, \mathbf{c}_y] \in \mathbb{R}^{2M}$ of PCA coefficients used as input to the network. Readers are referred to the DIMON paper and repository for full details.

\paragraph{Fourier Neural Operator (FNO).} \cite{FNOLi2020} The Fourier Neural Operator is implemented following the standard spectral convolution framework, where global interactions are captured through truncated Fourier modes in the frequency domain.

For all problems, FNO operates directly on the same multi-channel input representation used by DiSOL, including geometry masks, boundary condition channels, and source or load fields. No additional geometric preprocessing or mesh-based information is introduced.

Each FNO layer consists of:

\begin{itemize}
    \item [1)] A spectral convolution with a fixed number of retained Fourier modes in each spatial dimension.
    \item [2)] A pointwise linear transformation in physical space.
    \item [3)] A nonlinear activation.
\end{itemize}

The number of Fourier modes, channel width, and network depth are selected according to the corresponding configuration files and kept consistent across training, validation, and test experiments for each problem.

For scalar-field problems, FNO outputs a single-channel solution pattern field, while for linear elasticity, the output consists of two channels corresponding to the displacement components.

\paragraph{\textbf{Geo-FNO.}} Geo-FNO \cite{li2023fourier} extends Fourier Neural Operators (FNOs) to PDEs posed on general geometries by augmenting the operator backbone with a learned geometric deformation (coordinate warping) module. The deformation module aims to map geometry-dependent inputs to a latent coordinate system where FFT-based spectral convolutions can be applied effectively, after which predictions are mapped back to the physical domain. We use Geo-FNO as a representative geometry-mapping enhancement of FNO and adopt a formulation consistent with the original Geo-FNO paper; readers are referred to that work for the detailed architecture, training procedure, and theoretical motivation.

\paragraph{Graph Neural Operator (GNO).} We additionally report results for a Graph Neural Operator \cite{li2020neural} baseline, which represents continuous operator learning methods that incorporate explicit geometric structure via graph-based message passing or kernel aggregation. In our implementation, each grid cell (pixel) is treated as a node, and the graph is fully connected to enable geometry-aware information propagation. Importantly, although the computation is executed on a discrete grid, the underlying modeling paradigm remains that of a continuous neural operator, i.e., approximating a mapping between function spaces using globally shared kernels/aggregators. We include GNO to test whether adding explicit geometry encoding to a continuous-operator formulation alleviates the geometry-induced generalization failures observed for FNO/DeepONet under discrete, geometry-dependent PDE settings.

\paragraph{Fairness and consistency.} All baseline models share the same:

\begin{itemize}
    \item Input channel definitions and normalization,

    \item Geometry masking strategy,

    \item Loss function and evaluation metrics,

    \item Training and validation splits.
\end{itemize}

No baseline model is provided with additional geometric information, adaptive meshing, or problem-specific heuristics beyond what is available to DiSOL. Differences in performance therefore reflect differences in operator representation rather than disparities in input information or training procedures.

\subsection{Training settings}
\label{sect:3.4}

All models are trained using identical optimization and training settings across different problem types, except for the total number of training epochs. This design ensures a fair comparison between models and isolates performance differences from task-specific training heuristics.

\paragraph{Optimization.} All networks are trained using the same optimizer, learning rate, batch size, and learning rate scheduling strategy, as specified in the corresponding training scripts and configuration files. The loss function for all tasks is the global L1 loss. No task-specific tuning of optimization hyperparameters is performed.

\paragraph{Training epochs.} The total number of training epochs is adjusted according to the complexity of each problem:
\begin{itemize}
    \item Poisson equation: 500 epochs
    \item Advection-diffusion equation: 300 epochs
    \item Linear elasticity: 500 epochs
    \item Spatio-temporal heat equation: 500 epochs
\end{itemize}

All models are trained until convergence within the specified epoch budget, and the best-performing model on the validation set is selected for evaluation.

\paragraph{Consistency across models.} For each problem, DiSOL, FNO, and DeepONet are trained using the same training-validation splits and identical optimization settings. This ensures that differences in performance reflect differences in operator representations rather than disparities in training procedures.

\subsection{Model size}
\label{sect:3.5}

To ensure a fair comparison across different operator learning approaches, we report the total number of trainable parameters for all models and problem settings considered in this work.

Table \ref{tab:1} summarizes the parameter counts of DiSOL, FNO, and DeepONet for each problem. For all tasks, the model sizes are of comparable magnitude across different architectures.

In particular, DiSOL achieves competitive or superior performance without relying on larger parameter budgets, indicating that the observed improvements are primarily due to architectural design rather than increased model capacity.
\begin{table}
    \centering
    \caption{Number of trainable parameters}
    \begin{tabular}{|c|c|c|c|c|c|c|}
    \hline
        \textbf{Problem} & \textbf{DiSOL} & \textbf{DeepONet} & \textbf{DIMON} & \textbf{FNO} & \textbf{Geo-FNO} & \textbf{GNO} \\ \hline
        Poisson & 0.13M & 0.14M & 0.16M & 0.13M & 0.18M & 0.13M \\ \hline
        Advection-Diffusion & 0.49M & 0.66M & 0.64M & 0.53M & 0.51M & 0.53M \\ \hline
        Linear Elasticity & 7.52M & 7.49M & -- & 8.43M & -- & -- \\ \hline
        Thermal Conduction & 0.74M & N/A & -- & N/A & -- & -- \\ \hline
    \end{tabular}
    \label{tab:1}
\end{table}

\clearpage
\section{Output normalization and amplitude recovery}
\label{sect:4}
\subsection{Separation of pattern learning and amplitude recovery}

Given a physical-amplitude solution pattern field $U(x)$, we define a per-sample amplitude
\begin{equation}
    u_{\text{lim}}=\max_{x\in\Omega_h}|U(x)|
\end{equation}
and the normalized solution pattern
\begin{equation}
    u(x)=\frac{U(x)}{u_{\text{lim}}},\qquad \max_{x\in\Omega_h}|u(x)|=1.
\end{equation}
All models in the main text are trained to predict the pattern $\hat{u}(x)$. The recovery of the absolute amplitude $u_{\text{lim}}$ is treated as a secondary problem and handled separately.

\subsection{Log-scale regression of the amplitude}

In practice, the amplitude $u_{\text{lim}}$  exhibits a wide dynamic range across samples. Direct regression of $u_{\text{lim}}$ leads to unstable training and skewed error distributions.

To address this issue, we perform regression on the logarithmic scale and train a network to predict
\begin{equation}
    y=\ln(u_{\text{lim}}),
\end{equation}
and denote the prediction by $\hat{y}=\ln(\hat{u}_{\lim})$. The amplitude predictor shares the same backbone architecture as DiSOL, with the final multilayer perceptron modified to output a single scalar. The network is trained using a mean-squared error loss on $\ln(u_{\text{lim}})$. No coupling between pattern prediction and amplitude regression is introduced during training.

\subsection{Representative Poisson example}

We demonstrate amplitude recovery using the Poisson problem as a representative example. The model is trained and evaluated under in-distribution (ID) settings only. Our goal here is not to study geometric out-of-distribution generalization of the amplitude itself, but to verify that amplitude recovery can be achieved independently once the normalized solution pattern is accurately learned.

\begin{figure}
    \centering
    \includegraphics[width=0.5\linewidth]{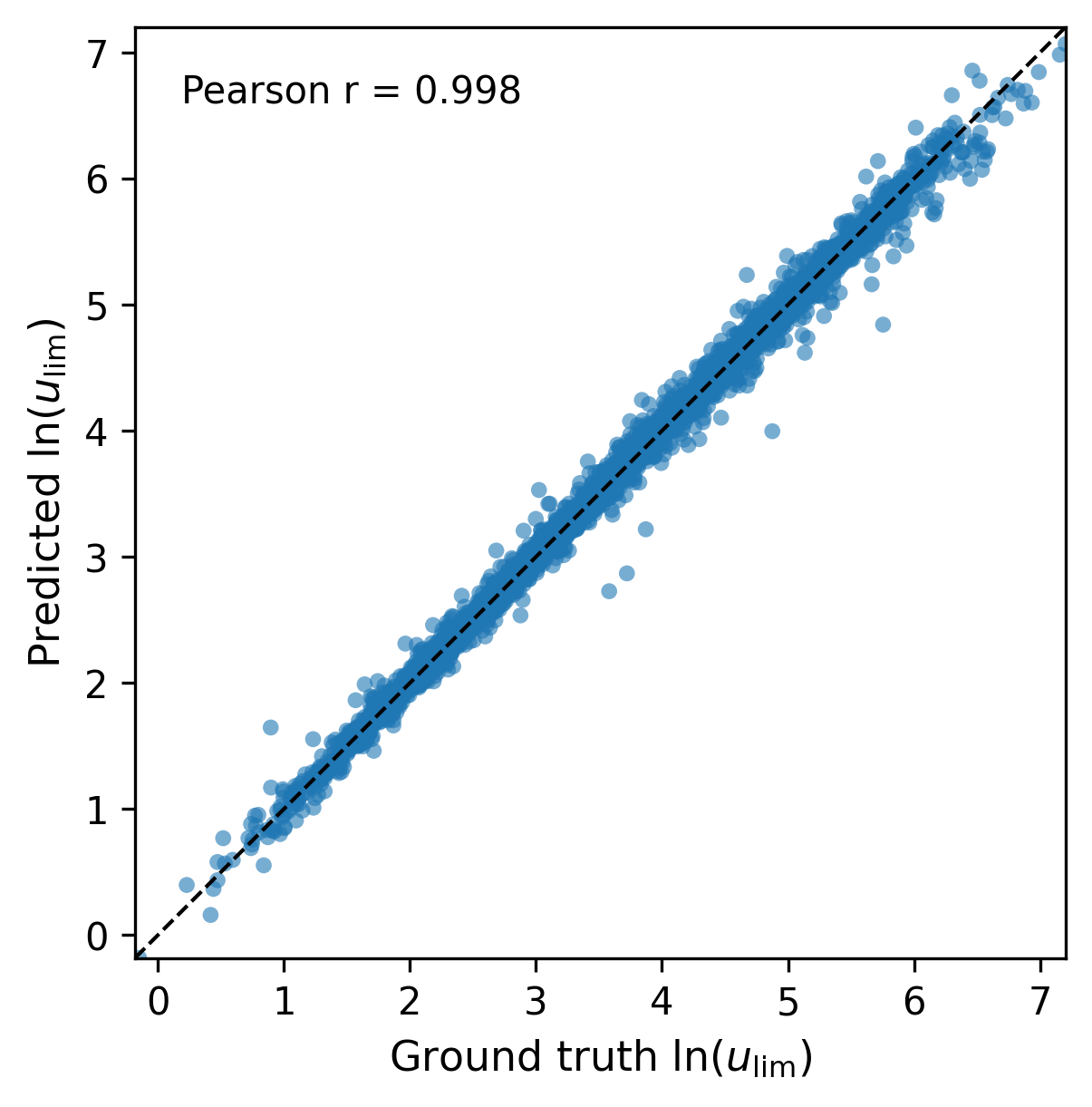}
    \caption{Predicted versus ground-truth values of $\ln(u_{\text{lim}})$ for the Poisson problem under in-distribution test settings. The global amplitude $u_{\text{lim}}$ is predicted on a logarithmic scale to account for its wide dynamic range. The dashed line indicates the ideal prediction $y=x$, and the strong Pearson correlation ($r=0.998$) confirms that amplitude recovery can be handled stably as an auxiliary task once the normalized solution pattern is learned.}
    \label{fig:s_ulim}
\end{figure}

For this Poisson case, the validation loss for $\ln(u_{\text{lim}})$ stabilizes at approximately $4\times10^{-2}$, indicating accurate and stable amplitude prediction within the training distribution. \textbf{Fig.} \ref{fig:s_ulim} further shows predicted versus ground-truth values of $\ln(u_{\text{lim}})$. The strong linear correlation indicates that the global amplitude can be recovered stably on the logarithmic scale. This auxiliary prediction enables reconstruction of the physical-amplitude solution field once the normalized solution pattern is obtained.

Together with the normalized pattern prediction provided by DiSOL, the physical-amplitude field can be reconstructed as
\begin{equation}
    U(x)=u_{\text{lim}} u(x).
\end{equation}
This experiment confirms that output normalization does not pose a fundamental limitation to solution recovery and that amplitude prediction can be treated as an auxiliary task without affecting the main conclusions of this work.

\clearpage
\section{Results and comparative study}
This section evaluates the performance of DiSOL on various PDE problems, along with comparative studies against other models. The PDE parameters for each problem are stated in Supplementary \ref{sect:1}.

The results are summarized in Tables~\ref{tab:train_errors}, \ref{tab:val_errors} and \ref{tab:test_errors}. For training, it shows that DiSOL, DeepONet, FNO, Geo-FNO, and DIMON have comparable computational costs for both training and inference. Overall, DiSOL achieves the best performance on most problems in the geometry generalization task for solving PDEs. Moreover, Geo-FNO can improve over FNO on several datasets, demonstrating that the geometric encoder provides a moderate improvement. DIMON performs comparably to or slightly better than DeepONet on regular geometries within the training distribution, but degrades on complex shapes due to its PCA-based geometric representation. Notably, DIMON is excluded from OOD test evaluation, as the test geometries contain internal holes that break the diffeomorphic requirement between the test shapes and the reference domain, rendering DIMON inapplicable. In contrast, GNO is significantly more expensive in both stages, mainly because the automatic GNO constructs a global fully connected graph that encodes the connections between all pixels, which is very costly to compute during both training and prediction. 

More detailed visualizations and comparisons of the models are presented in the following subsections, organized by PDE type.
\begin{table}[h]
    \centering
    \caption{Training errors (relative $\ell_2$-norm)}
    \label{tab:train_errors_1}
    \begin{tabular*}{\textwidth}{@{\extracolsep{\fill}}cccc}
        \hline
        Dataset (Training) & DiSOL & DeepONet & DIMON \\
        \hline
        Poisson Equation
           & $\textbf{2.81e-3} \boldsymbol{\pm} \textbf{3.6e-5}$ 
           & $1.61\text{e-2} \pm 5.8\text{e-4}$ 
           & $5.52\text{e-2} \pm 4.8\text{e-3}$ \\
        Advection-Diffusion ($Pe \approx 0.45$)
           & $\textbf{1.96e-3} \boldsymbol{\pm} \textbf{3.5e-5}$ 
           & $1.15\text{e-1} \pm 4.2\text{e-3}$
           & $2.05\text{e-2} \pm 1.9\text{e-3}$ \\
        Advection-Diffusion ($Pe \approx 4.5$)
           & $\textbf{1.86e-3} \boldsymbol{\pm} \textbf{3.3e-5}$ 
           & $1.12\text{e-1} \pm 4.0\text{e-3}$
           & $3.00\text{e-2} \pm 2.8\text{e-3}$ \\
        Linear Elasticity
           & $\textbf{1.29e-3} \boldsymbol{\pm} \textbf{2.3e-5}$ 
           & $5.92\text{e-2} \pm 2.1\text{e-3}$
           & -- \\
        \hline
    \end{tabular*}
    \begin{tabular*}{\textwidth}{@{\extracolsep{\fill}}cccc}
        \hline
        Dataset (Training) & FNO & Geo-FNO & GNO \\
        \hline
        Poisson Equation
           & $1.73\text{e-2} \pm 6.9\text{e-4}$
           & $1.65\text{e-2} \pm 4.4\text{e-4}$
           & $2.41\text{e-2} \pm 4.8\text{e-3}$ \\
        Advection-Diffusion ($Pe \approx 0.45$)
           & $6.47\text{e-2} \pm 1.8\text{e-3}$
           & $1.93\text{e-2} \pm 1.3\text{e-3}$
           & $7.09\text{e-2} \pm 1.9\text{e-3}$ \\
        Advection-Diffusion ($Pe \approx 4.5$)
           & $6.14\text{e-2} \pm 1.7\text{e-3}$
           & $2.02\text{e-2} \pm 1.5\text{e-3}$
           & $6.94\text{e-2} \pm 2.8\text{e-3}$ \\
        Linear Elasticity
           & $3.22\text{e-2} \pm 8.9\text{e-4}$
           & --
           & -- \\
        \hline
    \end{tabular*}
    \label{tab:train_errors}
\end{table}
\begin{table}[h]
    \centering
    \caption{ID validation errors (relative $\ell_2$-norm)}
    \label{tab:val_errors_1}
    \begin{tabular*}{\textwidth}{@{\extracolsep{\fill}}cccc}
        \hline
        Dataset (Validation) & DiSOL & DeepONet & DIMON \\
        \hline
        Poisson Equation
           & $\textbf{3.93e-3} \boldsymbol{\pm} \textbf{1.5e-5}$ 
           & $1.95\text{e-2} \pm 1.9\text{e-4}$
           & $5.39\text{e-2} \pm 1.5\text{e-3}$ \\
        Advection-Diffusion ($Pe \approx 0.45$)
           & $\textbf{3.52e-3} \boldsymbol{\pm} \textbf{1.3e-5}$ 
           & $1.17\text{e-1} \pm 1.2\text{e-3}$
           & $4.17\text{e-2} \pm 3.1\text{e-3}$ \\
        Advection-Diffusion ($Pe \approx 4.5$)
           & $\textbf{4.32e-3} \boldsymbol{\pm} \textbf{1.6e-5}$ 
           & $1.09\text{e-1} \pm 1.1\text{e-3}$
           & $6.06\text{e-2} \pm 3.6\text{e-3}$ \\
        Linear Elasticity
           & $\textbf{5.44e-3} \boldsymbol{\pm} \textbf{2.1e-5}$ 
           & $5.95\text{e-2} \pm 6.0\text{e-4}$
           & -- \\
        \hline
    \end{tabular*}
    \begin{tabular*}{\textwidth}{@{\extracolsep{\fill}}cccc}
        \hline
        Dataset (Validation) & FNO & Geo-FNO & GNO \\
        \hline
        Poisson Equation
           & $3.11\text{e-2} \pm 5.8\text{e-4}$
           & $2.26\text{e-2} \pm 4.2\text{e-4}$
           & $2.57\text{e-2} \pm 1.5\text{e-3}$ \\
        Advection-Diffusion ($Pe \approx 0.45$)
           & $6.59\text{e-2} \pm 1.4\text{e-3}$
           & $2.16\text{e-2} \pm 1.1\text{e-3}$
           & $7.13\text{e-2} \pm 6.1\text{e-3}$ \\
        Advection-Diffusion ($Pe \approx 4.5$)
           & $6.19\text{e-2} \pm 1.4\text{e-3}$
           & $2.39\text{e-2} \pm 1.0\text{e-3}$
           & $6.81\text{e-2} \pm 6.6\text{e-3}$ \\
        Linear Elasticity
           & $3.32\text{e-2} \pm 7.3\text{e-4}$
           & --
           & -- \\
        \hline
    \end{tabular*}
    \label{tab:val_errors}
\end{table}
\begin{table}[h]
    \centering
    \caption{OOD test errors (relative $\ell_2$-norm)}
    \label{tab:test_errors_1}
    \begin{tabular*}{\textwidth}{@{\extracolsep{\fill}}cccc}
        \hline
        Dataset (Test) & DiSOL & DeepONet & DIMON \\
        \hline
        Poisson Equation
           & $\textbf{9.98e-3} \boldsymbol{\pm} \textbf{6.0e-4}$ 
           & $2.80\text{e-2} \pm 6.7\text{e-4}$
           & -- \\
        Advection-Diffusion ($Pe \approx 0.45$)
           & $\textbf{7.79e-3} \boldsymbol{\pm} \textbf{4.7e-4}$ 
           & $7.67\text{e-2} \pm 1.8\text{e-3}$
           & -- \\
        Advection-Diffusion ($Pe \approx 4.5$)
           & $\textbf{1.35e-2} \boldsymbol{\pm} \textbf{8.1e-4}$ 
           & $8.66\text{e-2} \pm 2.1\text{e-3}$
           & -- \\
        Linear Elasticity
           & $\textbf{9.98e-2} \boldsymbol{\pm} \textbf{6.0e-3}$ 
           & $1.54\text{e-1} \pm 3.7\text{e-3}$
           & -- \\
        \hline
    \end{tabular*}
    \begin{tabular*}{\textwidth}{@{\extracolsep{\fill}}cccc}
        \hline
        Dataset (Test) & FNO & Geo-FNO & GNO \\
        \hline
        Poisson Equation
           & $2.76\text{e-2} \pm 1.1\text{e-3}$
           & $2.45\text{e-2} \pm 9.4\text{e-4}$
           & $2.93\text{e-2} \pm 8.0\text{e-3}$ \\
        Advection-Diffusion ($Pe \approx 0.45$)
           & $3.89\text{e-2} \pm 1.6\text{e-3}$
           & $1.64\text{e-2} \pm 1.3\text{e-3}$
           & $4.34\text{e-2} \pm 1.9\text{e-2}$ \\
        Advection-Diffusion ($Pe \approx 4.5$)
           & $5.59\text{e-2} \pm 2.3\text{e-3}$
           & $3.69\text{e-2} \pm 1.7\text{e-3}$
           & $5.83\text{e-2} \pm 1.1\text{e-2}$ \\
        Linear Elasticity
           & $1.23\text{e-1} \pm 5.1\text{e-3}$
           & --
           & -- \\
        \hline
    \end{tabular*}
    \label{tab:test_errors}
\end{table}
\clearpage
\subsection{Poisson Equation}
\begin{figure}[hbp]
    \centering
    \includegraphics[width=0.65\linewidth]{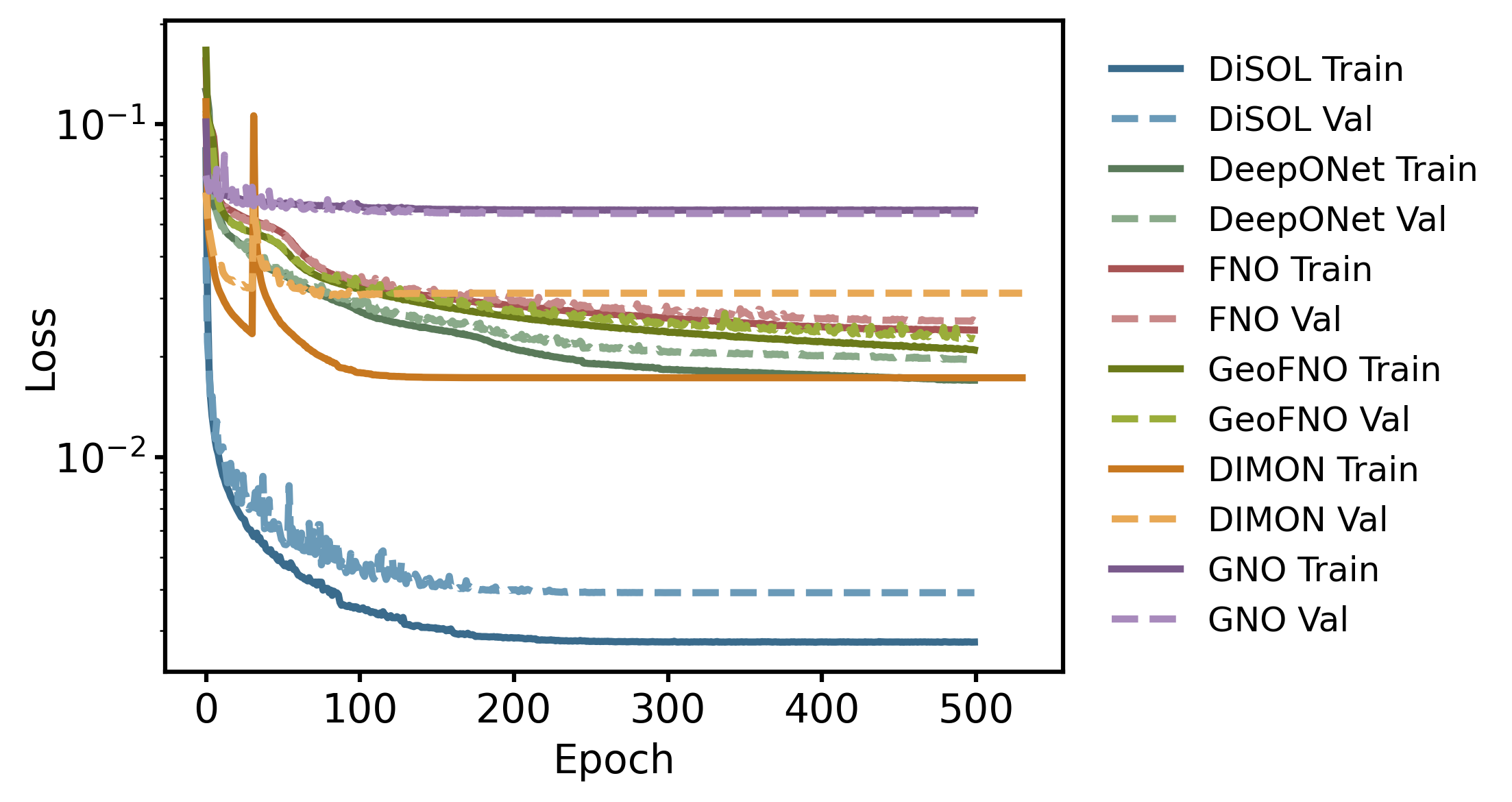}
    \caption{Training and validation loss evolution for the Poisson equation problem.}
    \label{fig:SI_PE_1}
\end{figure}

\begin{figure}
    \centering
    \includegraphics[width=0.999\linewidth]{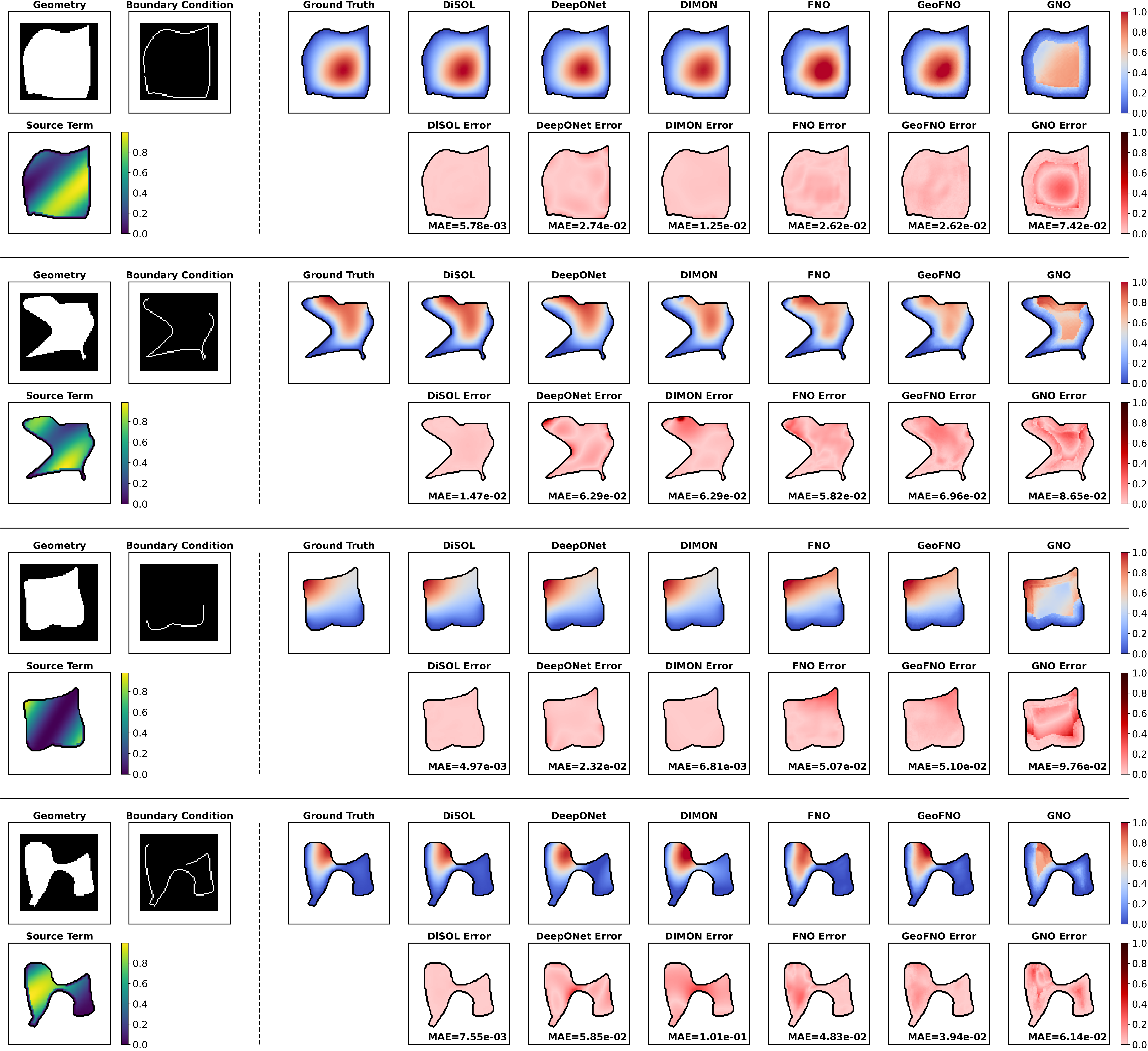}
    \caption{Validation results for the Poisson equation problem across four representative samples. For each case, the left panel displays the input fields (geometry, boundary condition, and source term), while the right panel shows the ground truth solution pattern alongside predictions from DiSOL, DeepONet, DIMON, FNO, Geo-FNO and GNO with corresponding absolute error maps.}
    \label{fig:SI_PE_2}
\end{figure}

\begin{figure}
    \centering
    \includegraphics[width=0.999\linewidth]{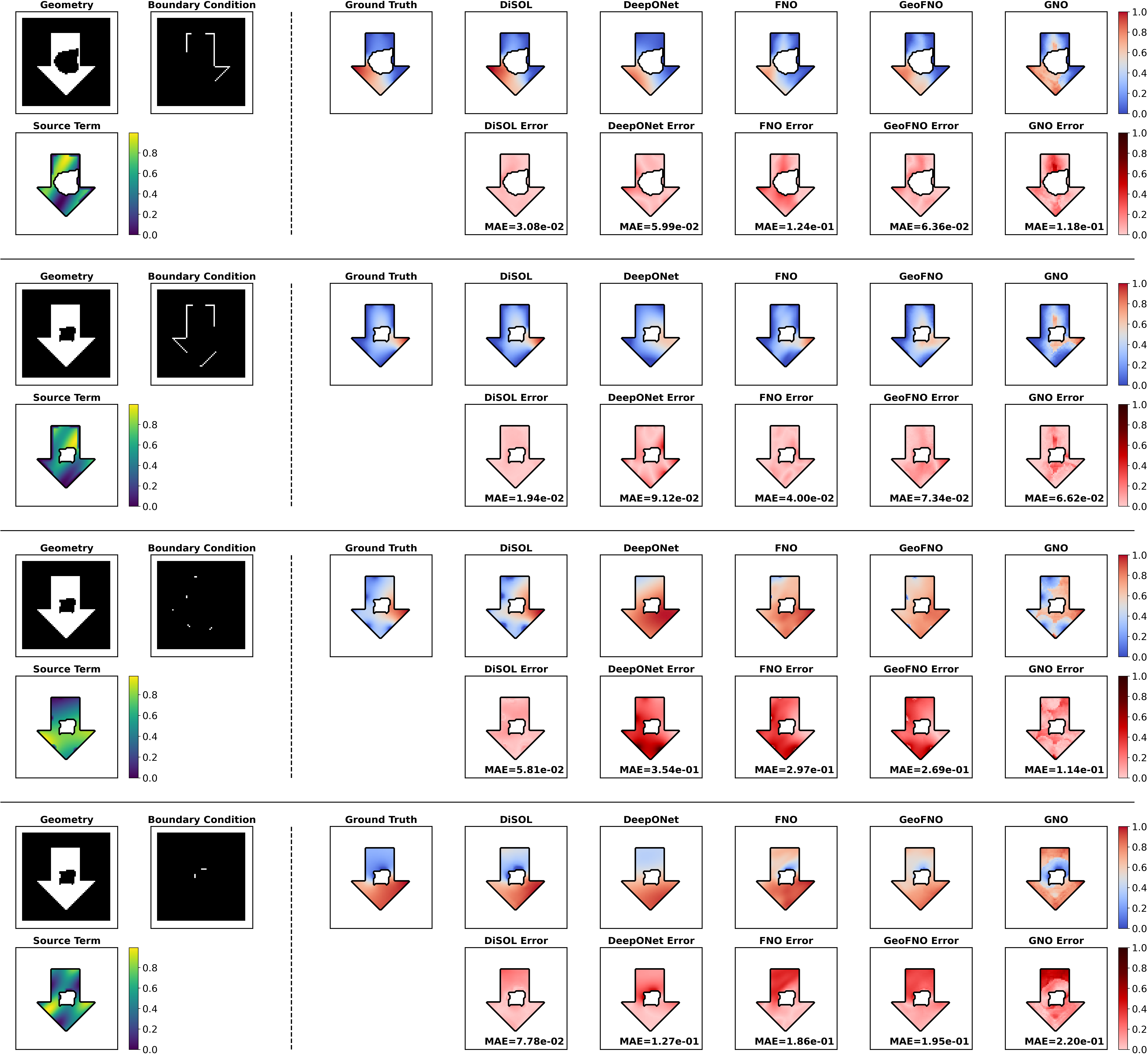}
    \caption{Test results for the Poisson equation problem across four representative samples. For each case, the left panel displays the input fields (geometry, boundary condition, and source term), while the right panel shows the ground truth solution pattern alongside predictions from DiSOL, DeepONet, FNO, Geo-FNO and GNO with corresponding absolute error maps.}
    \label{fig:SI_PE_3}
\end{figure}

The Poisson equation problem defined in Supplementary Information \ref{sect:1.1} is investigated. Results for the loss evolution, validation visualization and test visualization are shown in \textbf{Figs.} \ref{fig:SI_PE_1}, \ref{fig:SI_PE_2} and \ref{fig:SI_PE_3}, respectively.

\textbf{Figure}~\ref{fig:SI_PE_1} illustrates the training and validation loss evolution for this operator learning task. The DiSOL model achieves the lowest loss among all models, reaching approximately $2 \times 10^{-3}$ after 500 epochs. DeepONet converges to a higher loss around $2$--$3 \times 10^{-2}$, while FNO and GNO plateau at even higher values near $3$--$5 \times 10^{-2}$. Notably, GNO shows minimal improvement throughout training, with both training and validation losses remaining nearly flat after the initial epochs. The small gap between training and validation losses for DiSOL indicates good generalization without overfitting.

This observation is further supported by the validation set visualization presented in \textbf{Fig.}~\ref{fig:SI_PE_2}, where DiSOL consistently achieves the lowest MAE values across all four samples (ranging from $4.97 \times 10^{-3}$ to $1.47 \times 10^{-2}$). The competing models exhibit substantially higher prediction errors: DeepONet produces MAE values between $2.32 \times 10^{-2}$ and $6.29 \times 10^{-2}$, FNO yields errors from $2.62 \times 10^{-2}$ to $5.82 \times 10^{-2}$, and GNO shows the highest errors ranging from $6.14 \times 10^{-2}$ to $9.75 \times 10^{-2}$. The error maps reveal that GNO struggles particularly with capturing the overall solution distribution, producing visible artifacts throughout the domain. DIMON demonstrates marginally better performance than the standard DeepONet for geometries that are relatively regular and close to the reference shape (a circle), as observed in the first and third cases. However, for complex geometries, particularly those exhibiting strongly concave features, DIMON's accuracy degrades considerably. This limitation is from the PCA-based geometric representation employed in DIMON: the displacement field encoding used to parameterise the geometry struggles to represent target shapes that deviate significantly from the reference domain, as the principal components are derived from a training distribution centred around near-circular configurations. When the target geometry is highly irregular or concave, the low-dimensional PCA representation cannot adequately capture the shape, leading to elevated prediction errors most visibly at sharp corners and concave boundary regions where the geometric encoding is least accurate. For Geo-FNO, the introduction of a geometric encoder prior to the FNO spectral blocks yields a modest improvement in predictive accuracy, suggesting that explicitly encoding geometric information into the frequency-domain operator helps the model better adapt to irregular boundaries. Nevertheless, this improvement remains limited compared to the gains achieved by DiSOL, indicating that a more fundamental treatment of geometry, rather than a preprocessing encoding step, is necessary for robust performance across the full range of shape complexity encountered in this benchmark.

Beyond validation performance, we also investigate the out-of-distribution (OOD) zero-shot generalization capability of the trained models. For this test, we construct arrow-shaped geometries with randomly generated holes at their centres, apply discontinuous boundary conditions, and impose source fields with frequencies exceeding those used during training. The results are presented in \textbf{Fig.}~\ref{fig:SI_PE_3}. The first two cases feature different hole positions with discontinuous boundary conditions applied along the outer edges. DiSOL maintains accurate predictions in both cases with MAE values of $3.08 \times 10^{-2}$ and $1.94 \times 10^{-2}$, whereas DeepONet produces considerable errors ($5.99 \times 10^{-2}$ and $9.12 \times 10^{-2}$), and FNO and GNO exhibit even larger errors exceeding $10^{-1}$ in some cases.

The last two cases in \textbf{Fig.}~\ref{fig:SI_PE_3} involve short, disconnected boundary condition segments, which pose a greater learning challenge due to the highly localized information they introduce. In the third case, only small portions of the boundary are prescribed with non-zero Dirichlet conditions, resulting in a spatially concentrated solution pattern field. In the fourth case, the discontinuous boundary condition is applied to the internal hole boundary rather than the outer edges. In these more demanding scenarios, all models show increased errors compared to the first two cases. However, DiSOL continues to deliver reasonable predictions with MAE values of $5.81 \times 10^{-2}$ and $7.78 \times 10^{-2}$, while DeepONet, FNO, and GNO produce substantially larger errors ranging from $1.15 \times 10^{-1}$ to $3.54 \times 10^{-1}$. The Geo-FNO model is also presented with large discrepancies. The error maps clearly show that these competing methods fail to capture the localized solution behavior induced by the discontinuous boundary conditions.

\clearpage
\subsection{Advection-Diffusion Equation}
\begin{figure}[htbp]
    \centering
    \includegraphics[width=0.999\linewidth]{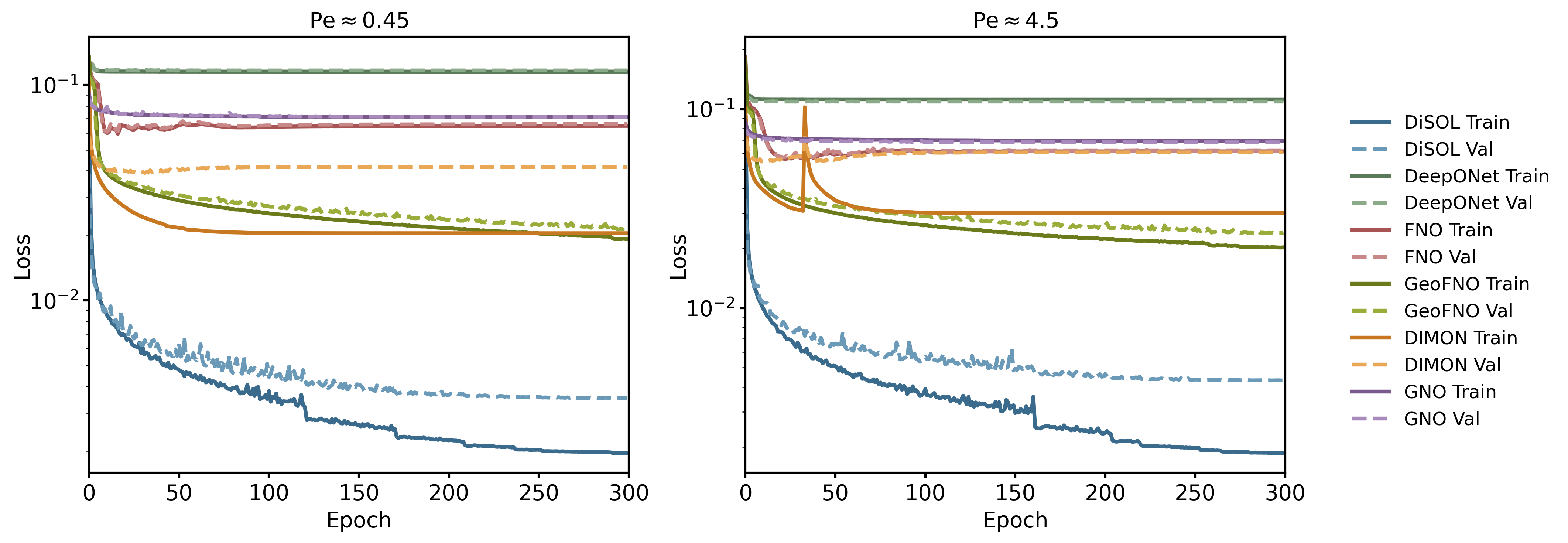}
    \caption{Training and validation loss evolution for the advection-diffusion problem at two Péclet numbers: $Pe \approx 0.45$ (left) and $Pe \approx 4.5$ (right).}
    \label{fig:SI_AD_1}
\end{figure}

\begin{figure}
    \centering
    \includegraphics[width=0.999\linewidth]{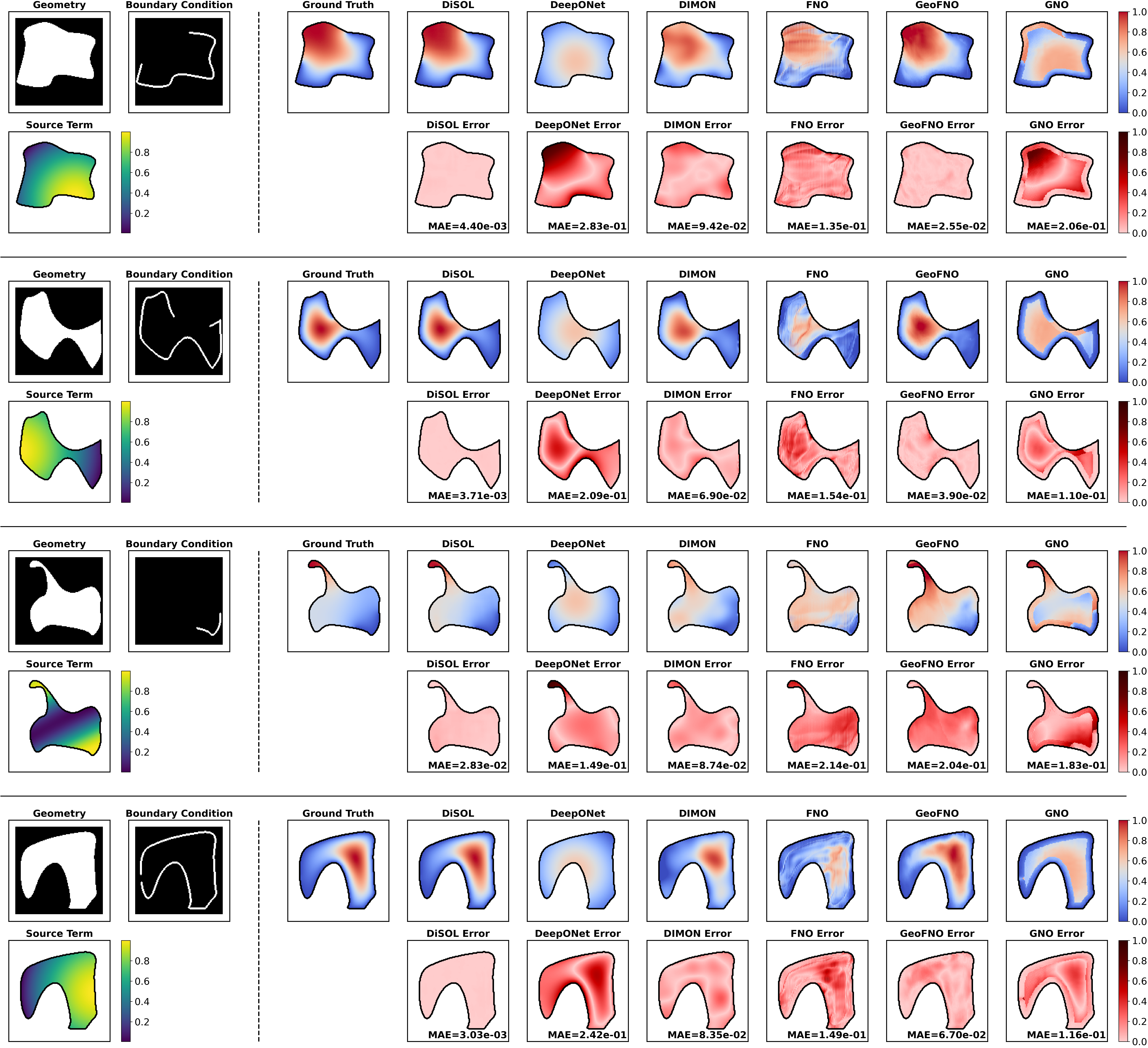}
    \caption{Validation results for the advection-diffusion equation problem with $Pe \approx 0.45$ across four representative samples. For each case, the left panel displays the input fields (geometry, boundary condition, and source term), while the right panel shows the ground truth solution pattern alongside predictions from DiSOL, DeepONet, DIMON, FNO, Geo-FNO and GNO with corresponding absolute error maps.}
    \label{fig:SI_AD_2_low}
\end{figure}

\begin{figure}
    \centering
    \includegraphics[width=0.999\linewidth]{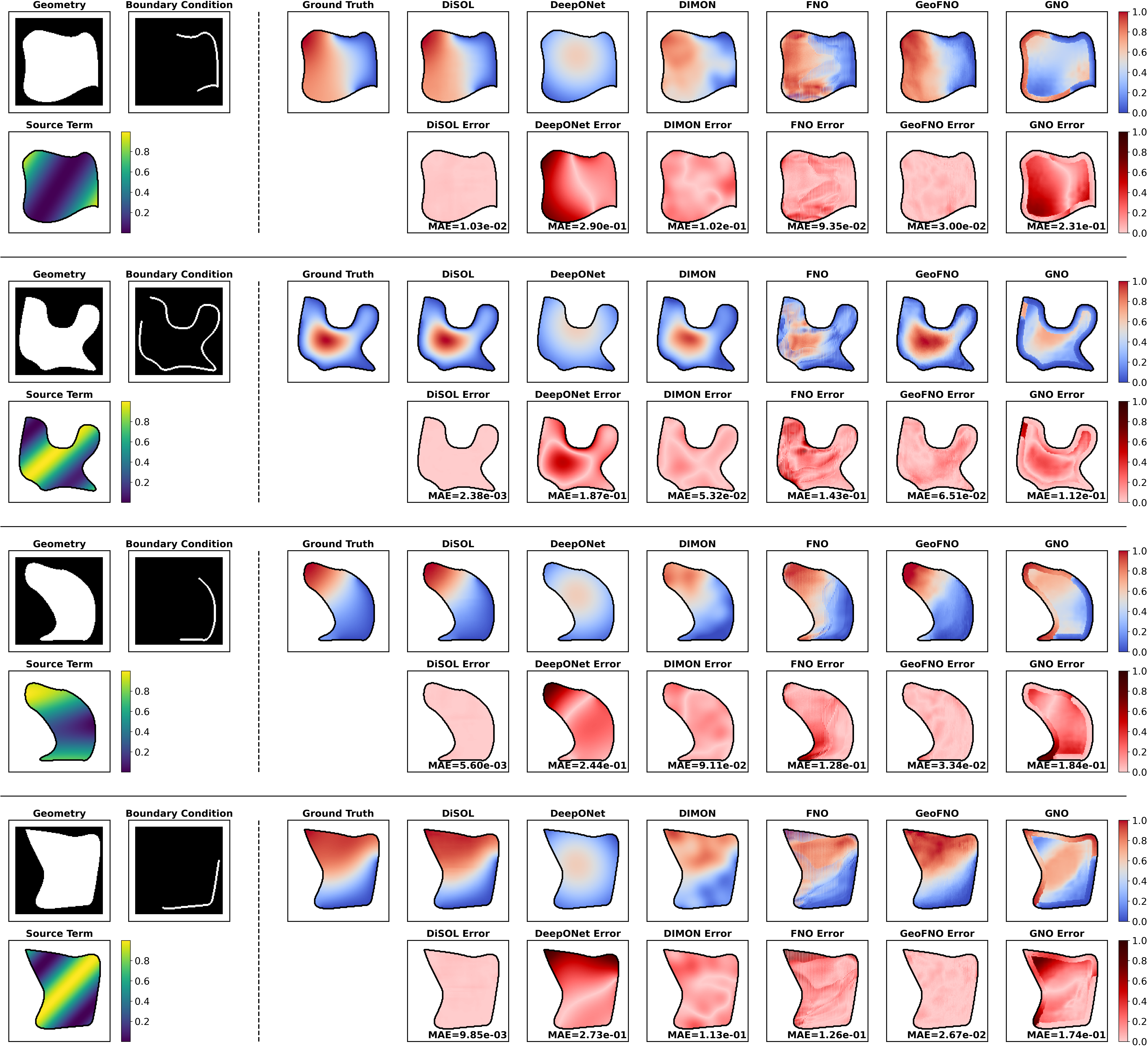}
    \caption{Validation results for the advection-diffusion equation problem with $Pe \approx 4.5$ across four representative samples. For each case, the left panel displays the input fields (geometry, boundary condition, and source term), while the right panel shows the ground truth solution pattern alongside predictions from DiSOL, DeepONet, DIMON, FNO, Geo-FNO and GNO with corresponding absolute error maps.}
    \label{fig:SI_AD_2_high}
\end{figure}

\begin{figure}
    \centering
    \includegraphics[width=0.999\linewidth]{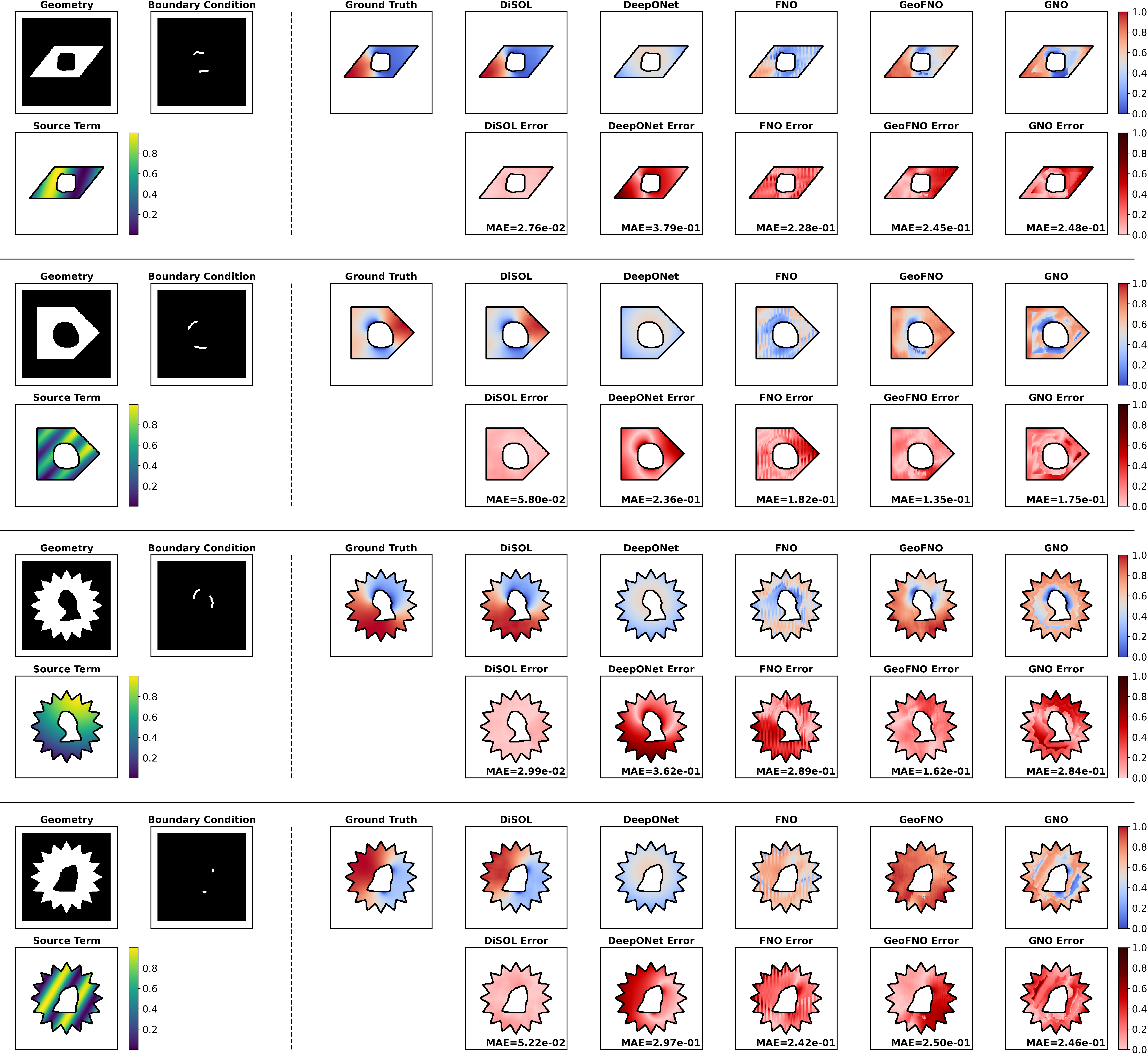}
    \caption{Test results for the advection-diffusion equation problem with $Pe \approx 4.5$ across four representative samples. For each case, the left panel displays the input fields (geometry, boundary condition, and source term), while the right panel shows the ground truth solution pattern alongside predictions from DiSOL, DeepONet, FNO, Geo-FNO and GNO with corresponding absolute error maps.}
    \label{fig:SI_AD_test}
\end{figure}

The advection-diffusion problem defined in Supplementary Information \ref{sect:1.2} is investigated. Results are shown in \textbf{Figs.}~\ref{fig:SI_AD_1}, \ref{fig:SI_AD_2_low}, \ref{fig:SI_AD_2_high} and \ref{fig:SI_AD_test}, respectively.

The advection-diffusion problem is more challenging than the Poisson equation because the solution behavior depends on the Péclet number ($Pe$), which characterizes the ratio of advective to diffusive transport. At low $Pe$, diffusion dominates and the solution varies smoothly across the domain. At high $Pe$, advection dominates, leading to sharper gradients and more localized features that are generally harder to learn. We study the operator learning task at both a low ($Pe \approx 0.45$) and a high ($Pe \approx 4.5$) Péclet number to examine how different models perform under these two regimes.

\textbf{Figure}~\ref{fig:SI_AD_1} shows the training and validation loss evolution for both cases. For the low Péclet number case ($Pe \approx 0.45$), DiSOL achieves the lowest loss, converging to approximately $2 \times 10^{-3}$ after 300 epochs. FNO converges to a higher loss around $6 \times 10^{-2}$, while DeepONet and GNO show minimal improvement throughout training, with losses remaining near $10^{-1}$. For the high Péclet number case ($Pe \approx 4.5$), the overall training is more difficult for all models. DiSOL still achieves the best performance, reaching a loss around $2 \times 10^{-3}$, though with a slightly larger gap between training and validation losses compared to the low $Pe$ case. FNO, DeepONet, and GNO all plateau at higher loss values between $5 \times 10^{-2}$ and $10^{-1}$.

The validation results for $Pe \approx 0.45$ are shown in \textbf{Fig.}~\ref{fig:SI_AD_2_low}. DiSOL consistently achieves the lowest MAE values across all four samples, ranging from $3.03 \times 10^{-3}$ to $2.83 \times 10^{-2}$. The competing models exhibit substantially higher errors: Geo-FNO produces MAE values between $2.55 \times 10^{-2}$ and $2.04 \times 10^{-1}$, DIMON yields errors from $8.35 \times 10^{-2}$ to $9.42 \times 10^{-2}$, FNO produces MAE values between $1.35 \times 10^{-1}$ and $2.14 \times 10^{-1}$, GNO yields errors from $1.10 \times 10^{-1}$ to $2.06 \times 10^{-1}$, and DeepONet shows the highest errors ranging from $1.49 \times 10^{-1}$ to $2.83 \times 10^{-1}$. Notably, Geo-FNO outperforms the standard FNO by a modest margin, consistent with the benefit of its geometric encoder. DIMON performs comparably to FNO on regular geometries but degrades on more complex shapes, as its PCA-based geometric representation struggles with geometries that deviate significantly from the reference domain. The error maps reveal that DeepONet and GNO fail to capture the overall solution distribution, while FNO produces visible stripe-like artifacts throughout the domain.

For the more challenging high P\'{e}clet number case ($Pe \approx 4.5$), the validation results are shown in \textbf{Fig.}~\ref{fig:SI_AD_2_high}. As expected, all models show increased errors compared to the low $Pe$ case. Nevertheless, DiSOL maintains good accuracy with MAE values ranging from $2.38 \times 10^{-3}$ to $1.03 \times 10^{-2}$, comparable to its performance at low $Pe$. In contrast, the competing models struggle more significantly: Geo-FNO produces errors from $3.00 \times 10^{-2}$ to $6.51 \times 10^{-2}$, remaining the best-performing baseline and again outperforming the standard FNO, which yields errors from $9.35 \times 10^{-2}$ to $1.43 \times 10^{-1}$. DIMON achieves errors between $5.32 \times 10^{-2}$ and $1.02 \times 10^{-1}$, performing better than FNO on smoother geometries but deteriorating on complex concave shapes where its geometric encoding is insufficient. GNO yields errors from $1.12 \times 10^{-1}$ to $2.31 \times 10^{-1}$, and DeepONet shows the highest errors from $1.87 \times 10^{-1}$ to $2.90 \times 10^{-1}$. The error maps show that FNO exhibits pronounced stripe artifacts, particularly in regions with sharp gradients, while DeepONet and GNO fail to capture the asymmetric solution patterns induced by advection.

Beyond validation performance, we also investigate the out-of-distribution (OOD) zero-shot generalization capability at the high Péclet number. For this test, we construct geometries with internal holes of various shapes (parallelograms, circles, and star-shaped boundaries), apply discontinuous or localized boundary conditions, and use source fields not seen during training. The results are presented in \textbf{Fig.}~\ref{fig:SI_AD_test}. The first two cases feature parallelogram and D-shaped geometries with rectangular and circular holes, respectively, combined with short discontinuous boundary conditions. DiSOL produces accurate predictions with MAE values of $2.76 \times 10^{-2}$ and $5.80 \times 10^{-2}$, while the competing models show errors exceeding $1.5 \times 10^{-1}$. The last two cases involve star-shaped outer boundaries with irregular internal holes, representing more complex geometries. In these cases, DiSOL maintains reasonable accuracy with MAE values of $2.99 \times 10^{-2}$ and $5.22 \times 10^{-2}$, whereas DeepONet, FNO, and GNO produce substantially larger errors ranging from $2.42 \times 10^{-1}$ to $3.62 \times 10^{-1}$. The error maps clearly show that these competing methods fail to capture the solution behavior in the presence of complex geometries and advection-dominated transport.

\clearpage
\subsection{Linear Elasticity}
\begin{figure}[h]
    \centering
    \includegraphics[width=0.65\linewidth]{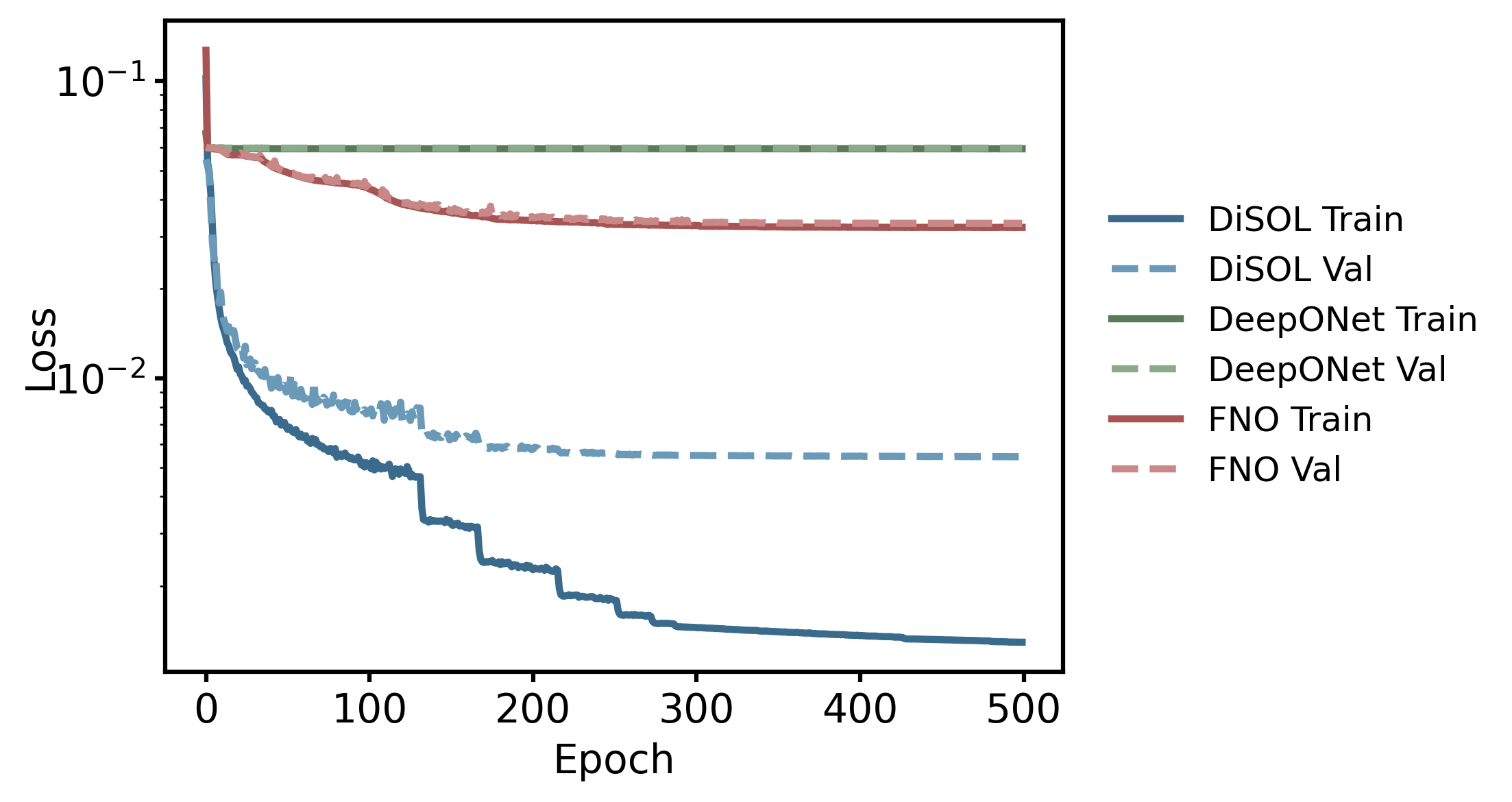}
    \caption{Training and validation loss evolution for the linear elasticity problem.}
    \label{fig:SI_EL_1}
\end{figure}

\begin{figure}
    \centering
    \includegraphics[width=0.999\linewidth]{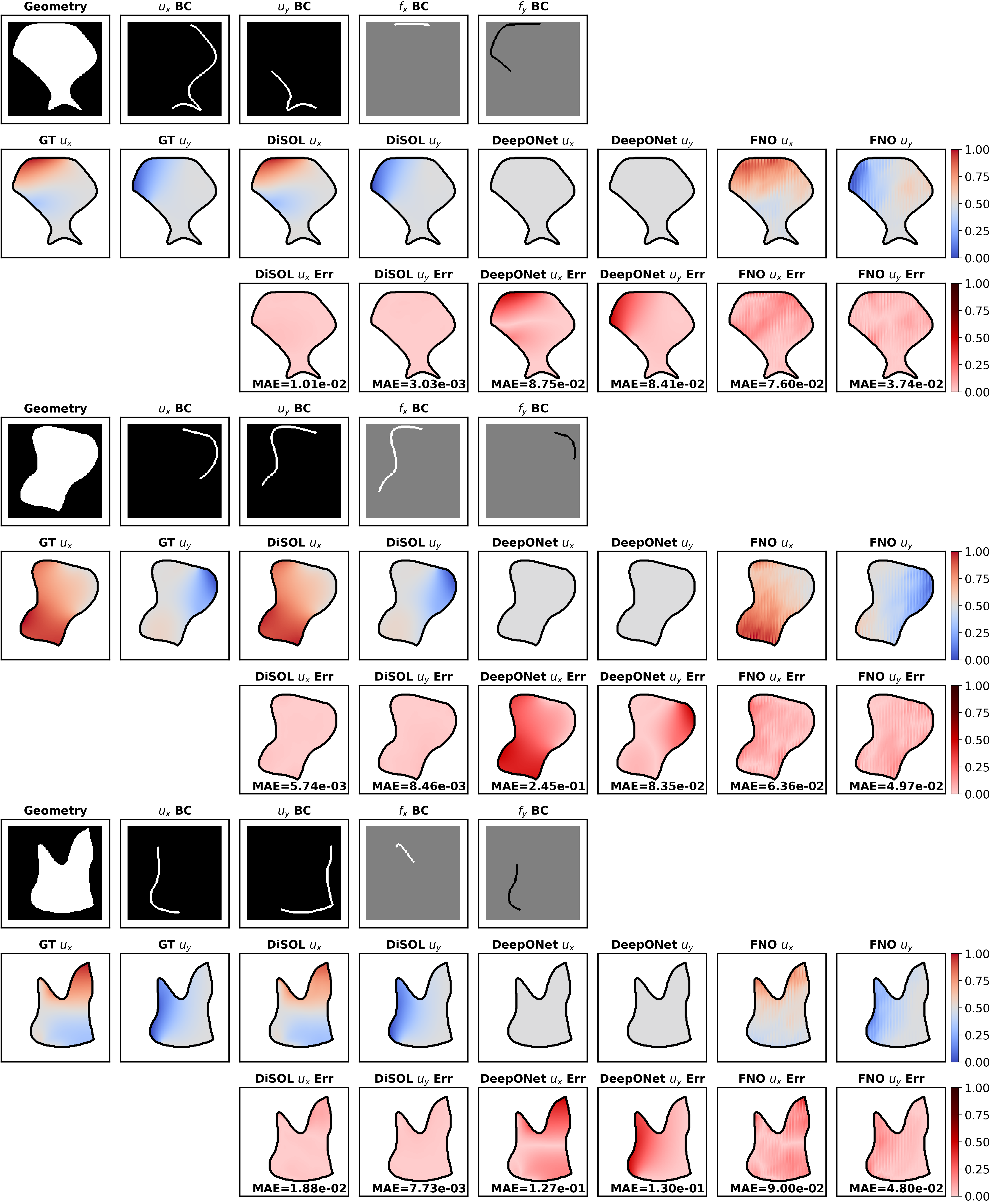}
    \caption{Validation results for the linear elasticity problem across three representative samples. For each case, the top panel displays the input fields (geometry and boundary conditions), while the bottom two panels show the ground truth solution pattern alongside predictions from DiSOL, DeepONet, and FNO with corresponding absolute error maps. The $u_x$ and $u_y$ BCs denote Dirichlet boundary conditions (prescribed displacements), while $f_x$ and $f_y$ represent Neumann boundary conditions (applied tractions) in the $x$ and $y$ directions, respectively. Black and white indicate values of $-1$ and $+1$, respectively.}
    \label{fig:SI_EL_2}
\end{figure}

\begin{figure}
    \centering
    \includegraphics[width=0.999\linewidth]{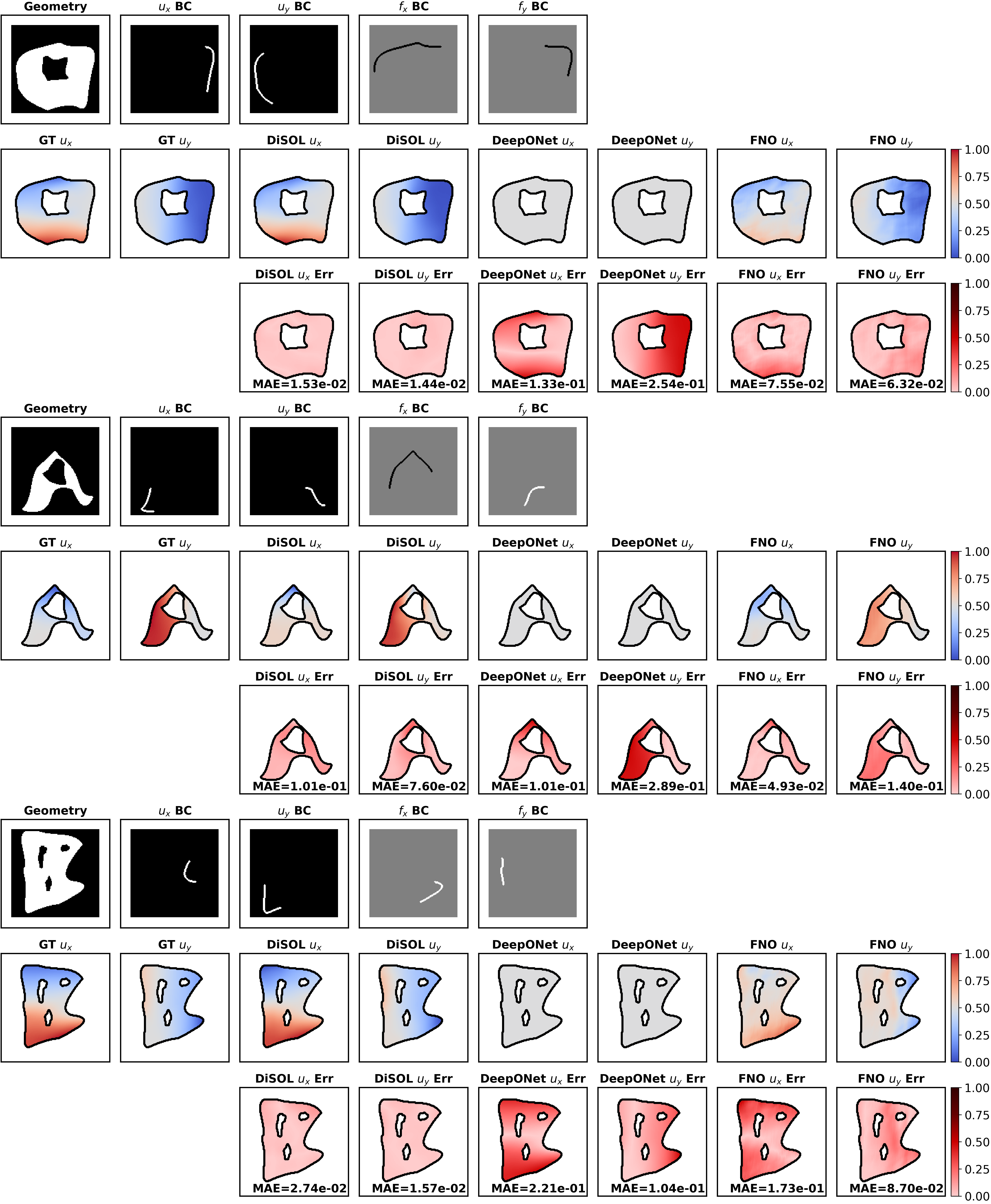}
    \caption{Test results for the linear elasticity problem across three representative samples. For each case, the top panel displays the input fields (geometry and boundary conditions), while the bottom two panels show the ground truth solution pattern alongside predictions from DiSOL, DeepONet, and FNO with corresponding absolute error maps. The $u_x$ and $u_y$ BCs denote Dirichlet boundary conditions (prescribed displacements), while $f_x$ and $f_y$ represent Neumann boundary conditions (applied tractions) in the $x$ and $y$ directions, respectively. Black and white indicate values of $-1$ and $+1$, respectively.}
    \label{fig:SI_EL_3}
\end{figure}

The linear elasticity problem defined in Supplementary Information \ref{sect:1.3} is investigated. Results for the loss evolution, validation visualization and test visualization are shown in \textbf{Figs.}~\ref{fig:SI_EL_1}, \ref{fig:SI_EL_2} and \ref{fig:SI_EL_3}, respectively.

The linear elasticity problem is a vector field problem, which is more complex than the scalar field problems considered previously. The inputs are defined in five channels: the first channel represents the geometry, two channels represent Dirichlet boundary conditions (prescribed displacements $u_x$ and $u_y$), and two channels represent Neumann boundary conditions (applied tractions $f_x$ and $f_y$). The outputs are two channels representing the displacement components $u_x$ and $u_y$. Note that GNO is excluded from this comparison as it does not support multi-channel outputs in its current implementation.

\textbf{Figure}~\ref{fig:SI_EL_1} shows the training and validation loss evolution for this problem. DiSOL achieves the lowest loss among all models, with the training loss reaching approximately $2 \times 10^{-3}$ after 500 epochs and the validation loss converging to around $5 \times 10^{-3}$. FNO converges to a higher loss around $3 \times 10^{-2}$, while DeepONet shows minimal improvement throughout training, with both training and validation losses remaining nearly flat at approximately $5 \times 10^{-2}$.

The validation results are shown in \textbf{Fig.}~\ref{fig:SI_EL_2}. DiSOL consistently achieves the lowest MAE values across all three samples for both displacement components. For $u_x$, DiSOL produces MAE values ranging from $5.74 \times 10^{-3}$ to $1.88 \times 10^{-2}$, while for $u_y$, the errors range from $3.03 \times 10^{-3}$ to $8.46 \times 10^{-3}$. In contrast, DeepONet exhibits substantially higher errors, with MAE values ranging from $8.75 \times 10^{-2}$ to $2.45 \times 10^{-1}$ for $u_x$ and $8.35 \times 10^{-2}$ to $1.30 \times 10^{-1}$ for $u_y$. FNO performs better than DeepONet but still produces errors significantly higher than DiSOL, with MAE values from $6.36 \times 10^{-2}$ to $9.00 \times 10^{-2}$ for $u_x$ and $3.74 \times 10^{-2}$ to $4.97 \times 10^{-2}$ for $u_y$. The error maps reveal that DeepONet fails to capture the overall displacement distribution, while FNO shows moderate errors distributed throughout the domain.

Beyond validation performance, we also investigate the out-of-distribution (OOD) zero-shot generalization capability. For this test, we construct geometries with internal holes of various shapes, including single rectangular holes, triangular holes, and geometries with multiple elongated holes. We also apply discontinuous boundary conditions, where Dirichlet and Neumann conditions are prescribed on different segments of the boundary. The results are presented in \textbf{Fig.}~\ref{fig:SI_EL_3}. In the first case, a geometry with a rectangular hole is subjected to both displacement and traction boundary conditions. DiSOL produces accurate predictions with MAE values of $1.53 \times 10^{-2}$ and $1.44 \times 10^{-2}$ for $u_x$ and $u_y$, respectively. DeepONet shows substantially higher errors ($1.33 \times 10^{-1}$ and $2.54 \times 10^{-1}$), while FNO yields errors of $7.55 \times 10^{-2}$ and $6.32 \times 10^{-2}$. The second case features a triangular geometry with a triangular hole, representing a more complex shape. DiSOL maintains reasonable accuracy with MAE values of $1.01 \times 10^{-1}$ and $7.60 \times 10^{-2}$, while DeepONet fails with errors of $1.01 \times 10^{-1}$ and $2.89 \times 10^{-1}$, and FNO produces errors of $4.93 \times 10^{-2}$ and $1.40 \times 10^{-1}$. The third case involves a geometry with multiple elongated holes and discontinuous boundary conditions applied to both the outer boundary and the internal hole boundaries. In this challenging scenario, DiSOL continues to deliver accurate predictions with MAE values of $2.74 \times 10^{-2}$ and $1.57 \times 10^{-2}$, whereas DeepONet produces errors of $2.21 \times 10^{-1}$ and $1.04 \times 10^{-1}$, and FNO yields errors of $1.73 \times 10^{-1}$ and $8.70 \times 10^{-2}$. These results demonstrate the ability of DiSOL to solve vector field problems with complex geometries and mixed boundary conditions.

\clearpage
\subsection{Heat Equation}
The thermal conduction problem defined in Supplementary Information \ref{sect:1.4} is investigated. The governing PDE shares the same spatial structure as the Poisson equation, with an additional first-order time derivative that governs the temporal evolution. We employ the same model configuration as used for the Poisson equation, with the only modification being the inclusion of the time dimension as an extra input channel. Since the comparative study between DiSOL and other operator learning methods has already been conducted for the Poisson equation in the context of spatial operator learning, we do not repeat this comparison here. Instead, we focus on demonstrating DiSOL's capability in handling dynamic problems using the simplest treatment of the time dimension as an additional model input.

\begin{figure}
    \centering
    \includegraphics[width=0.999\linewidth]{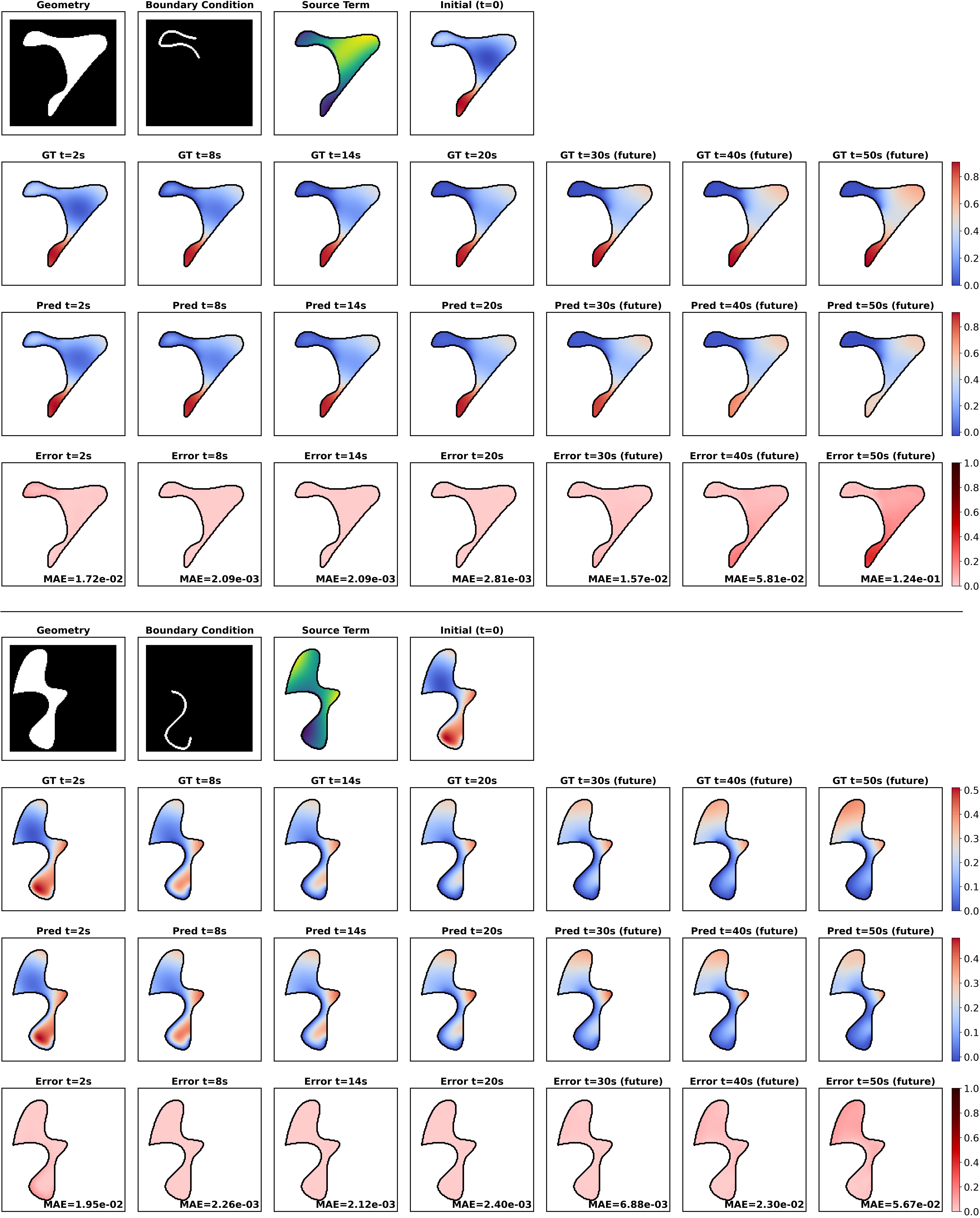}
    \caption{Validation results for the thermal conduction problem across two representative samples. For each case, the top row displays the input fields (geometry, boundary condition, source term, and initial condition), while the subsequent rows show the ground truth solution pattern, DiSOL predictions, and corresponding absolute error maps at various time steps. The final three columns (labeled "future") represent predictions at $t = 30$s, $40$s, and $50$s, which are beyond the training time horizon of $0$--$20$s.}
    \label{fig:SI_TC_1}
\end{figure}

\begin{figure}
    \centering
    \includegraphics[width=0.999\linewidth]{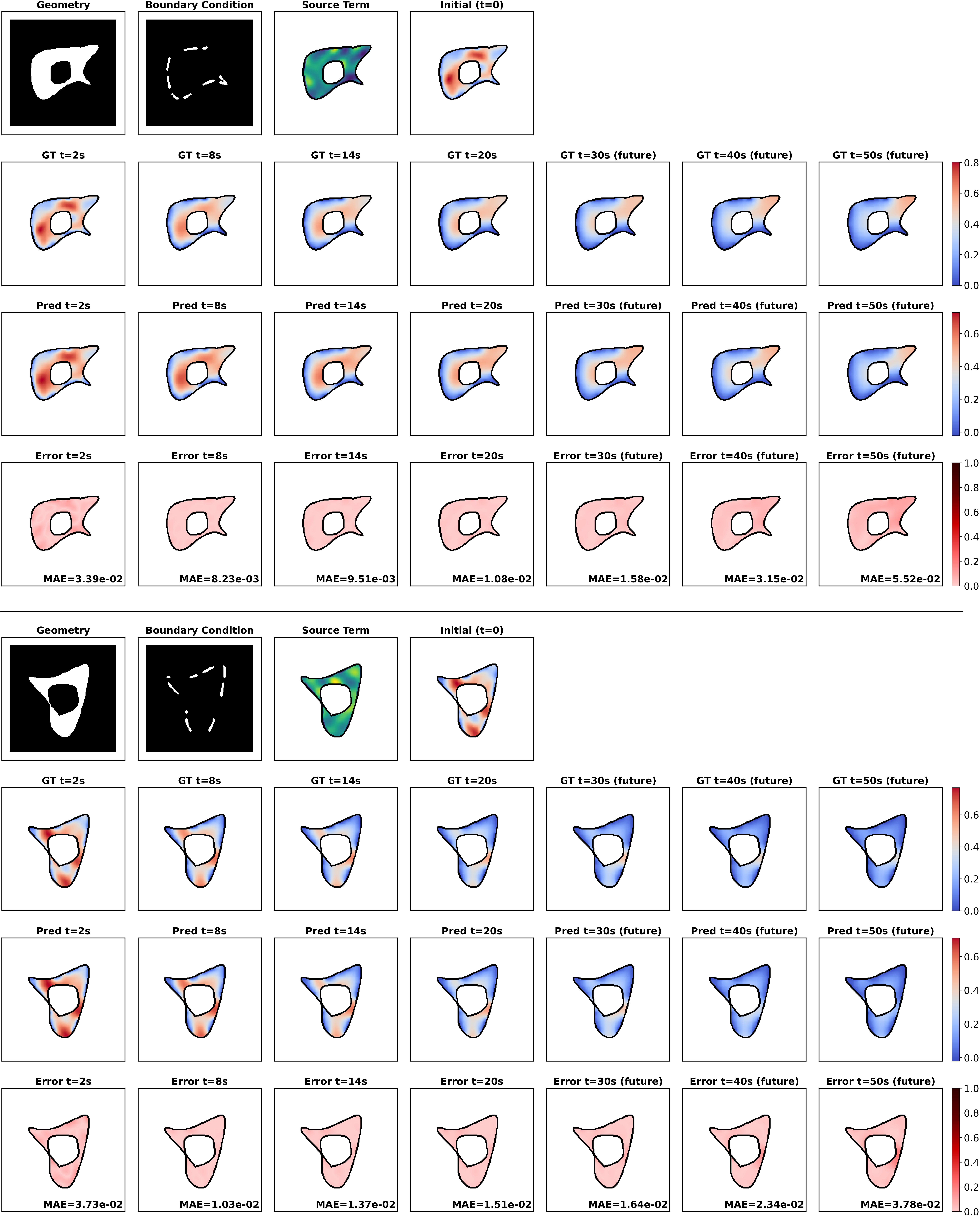}
    \caption{Test results for the thermal conduction problem across two representative samples with out-of-distribution geometries. For each case, the top row displays the input fields (geometry, boundary condition, source term, and initial condition), while the subsequent rows show the ground truth solution pattern, DiSOL predictions, and corresponding absolute error maps at various time steps. The final three columns (labeled "future") represent predictions at $t = 30$s, $40$s, and $50$s, which are beyond the training time horizon of $0$--$20$s.}
    \label{fig:SI_TC_2}
\end{figure}

The results are presented in \textbf{Fig.}~\ref{fig:SI_TC_1} for validation and \textbf{Fig.}~\ref{fig:SI_TC_2} for out-of-distribution (OOD) zero-shot testing. Each case shows the input fields (geometry, boundary condition, source term, and initial condition) along with the ground truth solution pattern, DiSOL predictions, and corresponding error maps at seven time steps: $t = 2$s, $8$s, $14$s, $20$s, $30$s, $40$s, and $50$s. Notably, training is conducted only on the interval from $0$s to $20$s, while the final three time steps ($t = 30$s, $40$s, and $50$s, labeled as "future") represent predictions at time points not seen during training. This setup allows us to evaluate both the model's interpolation accuracy within the training time horizon and its extrapolation capability beyond it.

For the validation cases shown in \textbf{Fig.}~\ref{fig:SI_TC_1}, DiSOL demonstrates strong performance across the entire temporal evolution. In the first case, which features a geometry with a rectangular hole and localized initial hot spots, the MAE values within the training time horizon range from $8.23 \times 10^{-3}$ to $1.08 \times 10^{-2}$ (at $t = 8$s to $20$s), with a slightly higher error of $3.39 \times 10^{-2}$ at $t = 2$s due to the sharp initial transient. For the future predictions, the errors gradually increase but remain reasonable: $1.58 \times 10^{-2}$ at $t = 30$s, $3.15 \times 10^{-2}$ at $t = 40$s, and $5.52 \times 10^{-2}$ at $t = 50$s. The second validation case involves a geometry with a triangular hole and discontinuous boundary conditions. Similar behavior is observed, with MAE values ranging from $1.03 \times 10^{-2}$ to $1.51 \times 10^{-2}$ within the training horizon and increasing to $1.64 \times 10^{-2}$, $2.34 \times 10^{-2}$, and $3.78 \times 10^{-2}$ for the future predictions.

The OOD zero-shot test results are shown in \textbf{Fig.}~\ref{fig:SI_TC_2}, where the model is evaluated on unseen geometries and source fields with frequencies outside the training distribution. The first test case features a Y-shaped geometry with a continuous boundary condition applied along one edge. DiSOL maintains good accuracy within the training time horizon, with MAE values ranging from $2.09 \times 10^{-3}$ to $2.81 \times 10^{-3}$ (at $t = 8$s to $20$s). The future predictions show gradually increasing errors: $1.57 \times 10^{-2}$ at $t = 30$s, $5.81 \times 10^{-2}$ at $t = 40$s, and $1.24 \times 10^{-1}$ at $t = 50$s. The error maps indicate that the largest deviations occur near the boundary where the Dirichlet condition is applied. The second test case involves an irregular geometry with a continuous boundary condition. DiSOL again performs well within the training horizon, with MAE values from $2.12 \times 10^{-3}$ to $2.40 \times 10^{-3}$ (at $t = 8$s to $20$s), and the future prediction errors increase to $6.88 \times 10^{-3}$, $2.30 \times 10^{-2}$, and $5.67 \times 10^{-2}$ at $t = 30$s, $40$s, and $50$s, respectively.

These results demonstrate that DiSOL effectively handles spatiotemporal operator learning tasks using a spatially discrete learning strategy with time as an additional input channel. The model achieves accurate predictions both for in-distribution validation cases and OOD zero-shot scenarios. Furthermore, the future forecasting results indicate that DiSOL can extrapolate beyond the training time horizon with reasonable accuracy, although the errors gradually accumulate as the prediction horizon extends further into the future.

\clearpage
\subsection{Zero-shot cross-resolution generalization}

We test whether DiSOL exhibits cross-resolution transfer when trained at a single resolution, motivated by its locality-preserving operator decomposition. For linear PDEs, the solution patterns can exhibit approximately resolution-consistent local structures when the underlying geometry/BC/source are represented on an embedded grid in a comparable manner. This motivates testing whether a model trained at one grid resolution can be used as-is at other resolutions, despite the change in discrete scale. Herein, we investigate this by training the model on $64 \times 64$ grids and evaluating on higher resolutions ($96 \times 96$, $128 \times 128$, and $256 \times 256$). For each target resolution, we re-discretize the same continuous problem specification onto the corresponding embedded grid: the geometry mask and boundary-selection map are rasterized at the target resolution, and source fields are evaluated on the target grid. Reference solutions are generated by solving the PDE at the same target resolution using the same numerical procedure as in Supplementary Information \ref{sect:2}, and the per-sample amplitude $u_{\text{lim}}$ is computed from the target-resolution physical solution $U_h$ before forming the normalized pattern $u_h=U_h/u_{\text{lim}}$.

\begin{figure}[h]
    \centering
    \includegraphics[width=\textwidth]{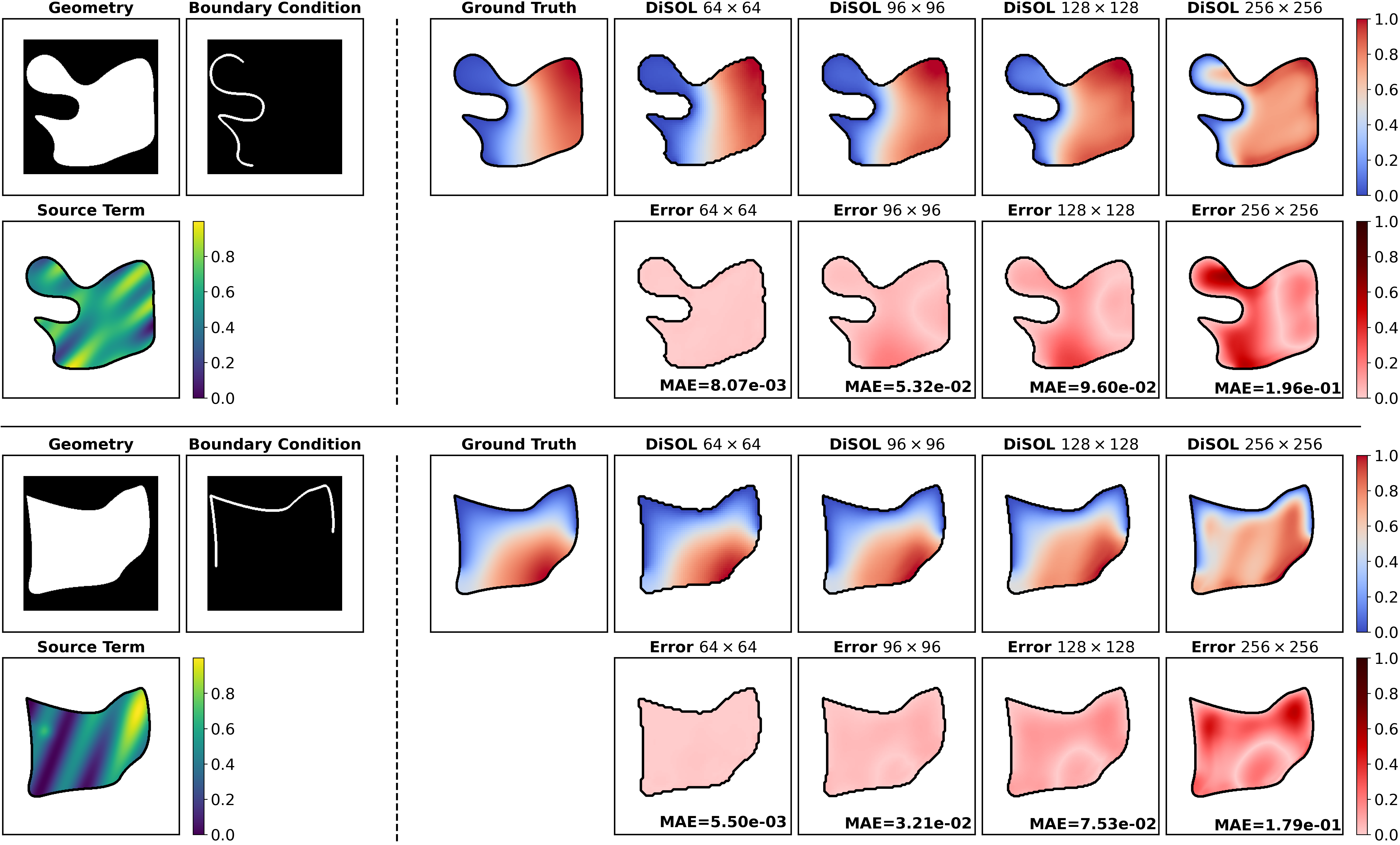}
    \caption{Zero-shot cross-resolution on ID validation cases for the Poisson equation problem across two representative samples. For each case, the left panel displays the input fields (geometry, boundary condition, and source term), while the right panel shows the ground truth solution pattern alongside DiSOL predictions at different resolutions ($64 \times 64$, $96 \times 96$, $128 \times 128$, and $256 \times 256$) with corresponding absolute error maps and MAE values.}
    \label{fig:superres_val}
\end{figure}

\begin{figure}[h]
    \centering
    \includegraphics[width=\textwidth]{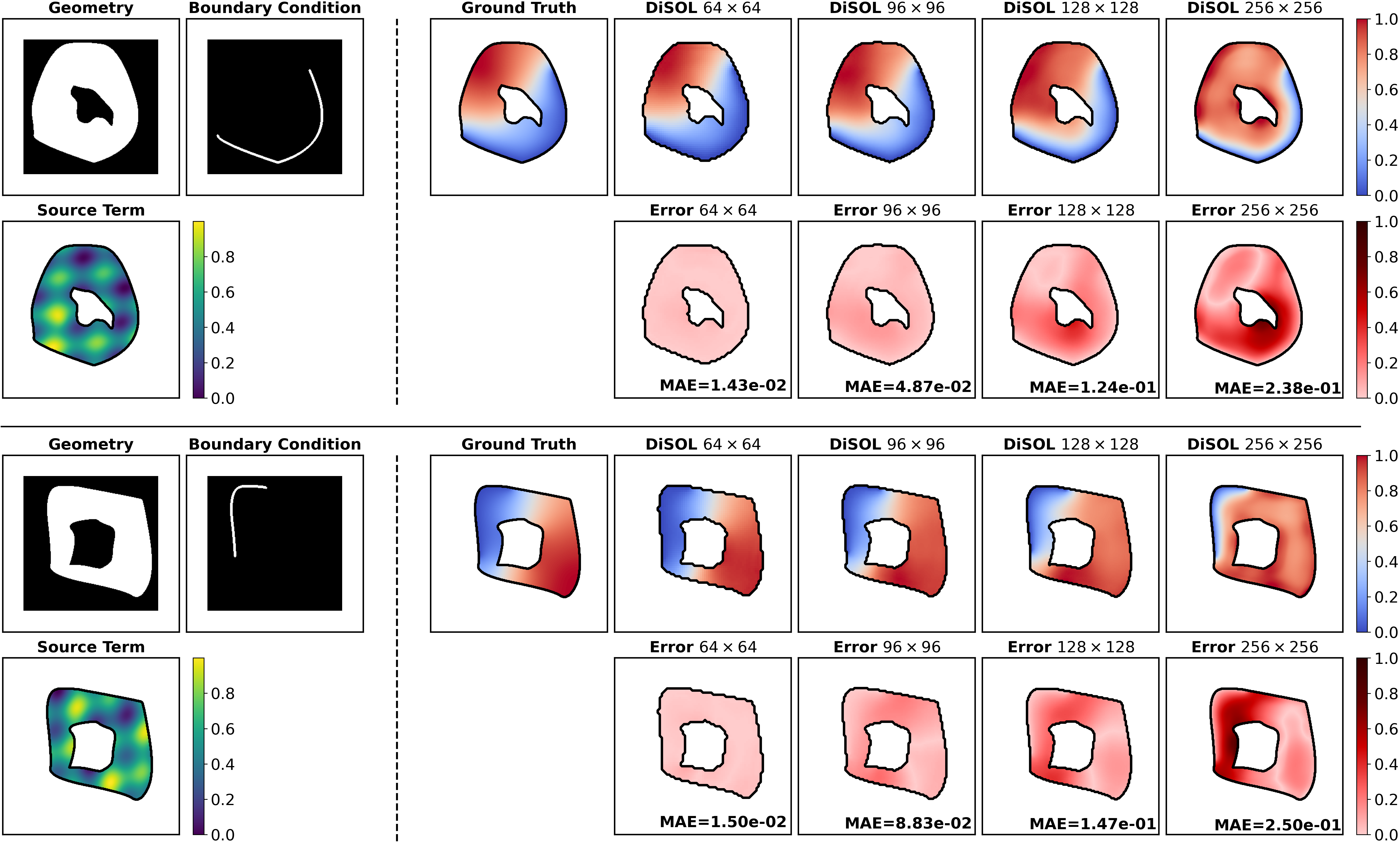}
    \caption{Zero-shot cross-resolution on OOD test cases for the Poisson equation problem across two representative samples. For each case, the left panel displays the input fields (geometry, boundary condition, and source term), while the right panel shows the ground truth solution pattern alongside DiSOL predictions at different resolutions ($64 \times 64$, $96 \times 96$, $128 \times 128$, and $256 \times 256$) with corresponding absolute error maps and MAE values.}
    \label{fig:superres_ood}
\end{figure}

Figures \ref{fig:superres_val} and \ref{fig:superres_ood} present the cross-resolution evaluation results across four representative cases. When evaluated at the training resolution ($64 \times 64$), DiSOL produces accurate predictions with MAE values ranging from $5.50\times 10^{-3}$ to $1.50\times10^{-2}$, consistent with the results reported in previous sections. At $96 \times 96$ resolution, which represents a $1.5\times$ resolution increase, the model maintains reasonable accuracy with MAE values between $3.21\times10^{-2}$ and $8.83\times10^{-2}$. The predicted fields preserve the overall solution structure and capture the essential features of the solution pattern, though some smoothing of sharp gradients becomes visible. As the resolution increases to $128 \times 128$ ($2\times$ resolution increase), the errors grow more noticeably, with MAE values reaching $7.53\times10^{-2}$ to $1.47\times10^{-1}$. The error maps reveal that the largest discrepancies occur near boundaries and in regions with steep gradients, where the local features learned at the training resolution become insufficient to resolve finer details. At $256 \times 256$ resolution ($4\times$ resolution increase), the model struggles to maintain accuracy, with MAE values of $1.79\times10^{-1}$ to $2.50\times10^{-1}$. Despite these increased errors, the model still captures the qualitative behavior of the solution, including the correct identification of high and low value regions. This degradation pattern is expected since the receptive field of the local convolution operators, which was optimized for $64 \times 64$ grids, becomes relatively smaller at higher resolutions and thus fails to capture the necessary contextual information. These results suggest that DiSOL exhibits some inherent ability to generalize across resolutions due to its local operator learning approach, though the practical applicability is limited to moderate resolution changes without additional fine-tuning or architectural modifications.

\clearpage
\section{Failure cases}

\begin{figure}[h]
    \centering
    \includegraphics[width=0.95\linewidth]{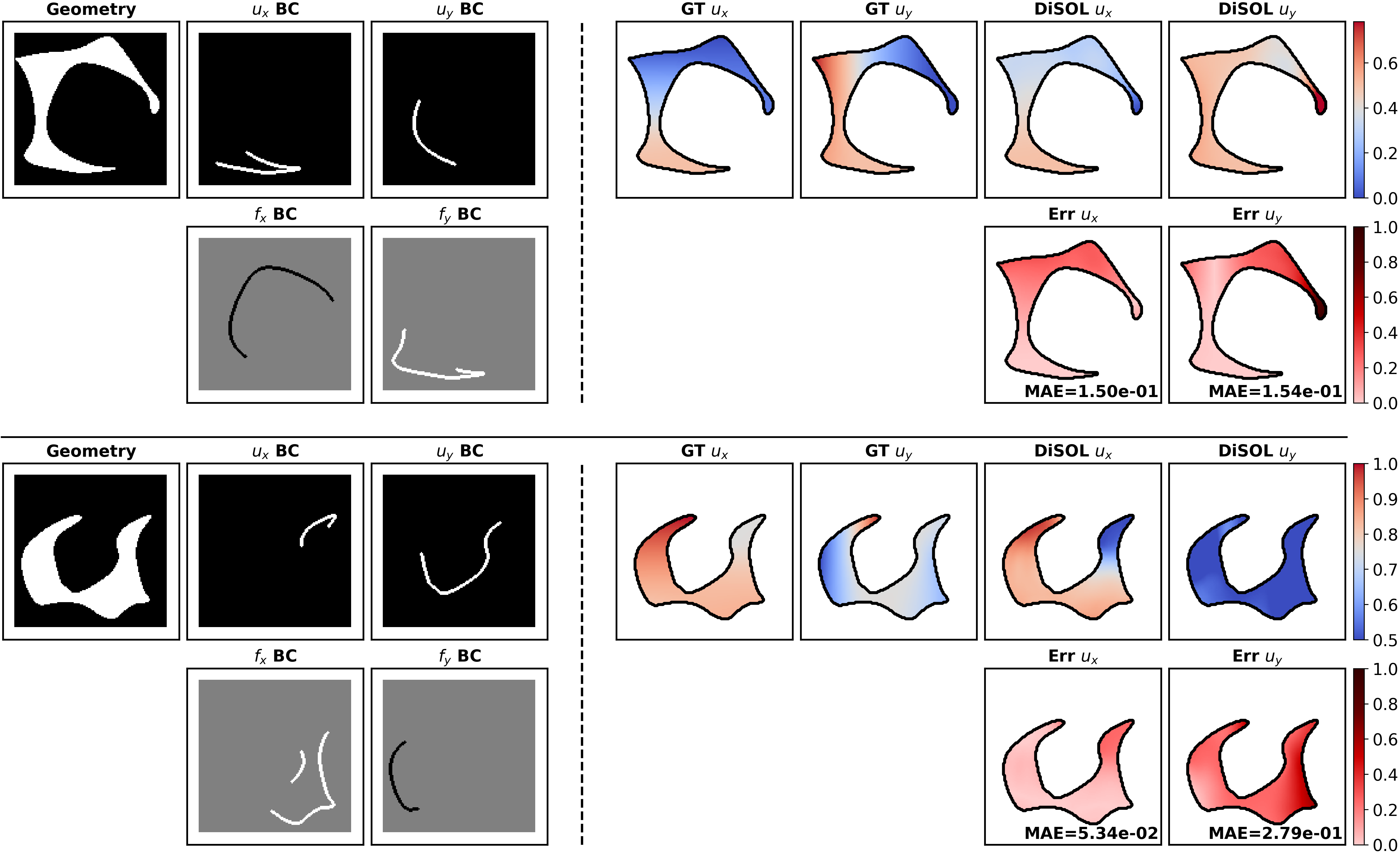}
    \caption{Representative failure cases for the linear elasticity problem. For each case, the left part show the input fields (geometry, displacement boundary conditions $u_x$~BC and $u_y$~BC, and traction boundary conditions $f_x$~BC and $f_y$~BC), and the right part depicts the ground-truth displacement components (GT~$u_x$, GT~$u_y$), the corresponding DiSOL predictions (DiSOL~$u_x$, DiSOL~$u_y$), and the absolute error maps for each displacement component together with the MAE values. The largest errors occur near thin, curved regions of the geometry and along localized boundary-condition segments, where the vector-valued solution exhibits strong anisotropy and sharp spatial variations, making these cases substantially more challenging.}
    \label{fig:SI_EL_failure}
\end{figure}

While DiSOL performs robustly across geometries and boundary conditions, especially for scalar problems such as Poisson’s equation and advection-diffusion, there remain challenging cases for the linear elasticity setting where the model exhibits relatively larger errors. \textbf{Fig.}~\ref{fig:SI_EL_failure} shows two representative failure cases for plane elasticity. In the first case, both displacement components reach mean absolute errors on the order of $1.5\times10^{-1}$, while in the second case the error in $u_y$ is particularly high (about $2.8\times10^{-1}$).

Compared to the scalar Poisson and advection-diffusion problems, where DiSOL consistently provides accurate predictions over a wide range of geometries, the linear elasticity problem is inherently more demanding: the output is a \emph{vector field} $(u_x,u_y)$, and the two components are strongly coupled by equilibrium and compatibility relations. Any local mismatch in one component or near the boundary can propagate globally and affect the overall deformation pattern. In both failure cases of \textbf{Fig.}~\ref{fig:SI_EL_failure}, the geometries contain highly curved, nonconvex regions and slender ligaments. The Dirichlet and Neumann boundary conditions are applied along relatively short, curved segments, often at the tips or along thin parts of the structure. These configurations generate complex bending and shear-dominated deformation modes, with sharp spatial variations in both $u_x$ and $u_y$ that are difficult for the convolution-based architecture to resolve.

The error maps reveal that the largest discrepancies occur near these curved boundary segments and around the thin arms of the geometry, where the ground-truth displacement fields vary rapidly in magnitude and direction. This suggests that, although DiSOL captures the global deformation trends, it underestimates small-scale anisotropic features and local gradients in vector-valued solutions. In contrast, for Poisson’s equation and advection-diffusion, the model deals with a single scalar field and generally smoother solution structure, which makes the learning task comparatively easier and leads to far fewer pronounced failure cases. Future work could address these limitations in elasticity by incorporating cross-resolution techniques, and by enforcing stronger coupling between displacement components directly within the network architecture.

\clearpage

\section{Mechanistic interpretation of discrete operator learning}

This section provides a conceptual interpretation of the proposed DiSOL framework from the perspective of discrete operator learning. The goal is not to establish formal mathematical guarantees, but to clarify why the architectural principles adopted in DiSOL naturally lead to robust generalization across geometries, boundary configurations, and out-of-distribution settings.

\subsection{Local operators as discrete stencil learners}

At its core, DiSOL is built upon the principle of locality. Each convolutional block acts as a local operator that maps neighborhood-level input information—geometry indicators, boundary conditions, and source terms—to local solution responses.

This behavior closely resembles classical numerical discretizations, such as finite difference \cite{thomas2013numerical} or finite element methods, where local stencils define how solution values depend on nearby degrees of freedom. Importantly, the learned local operators in DiSOL are shared across the entire domain and across different geometries.

As a result, the model does not encode global geometric shapes explicitly. Instead, it learns reusable local transformation rules that can be applied consistently regardless of boundary shape or domain topology. This locality is a primary reason why DiSOL exhibits strong geometric generalization: new geometries are handled by reapplying the same local operators at new spatial locations.

\subsection{Multi-scale assembly as implicit domain decomposition}

While local operators capture neighborhood-level interactions, global solution behavior emerges through the hierarchical aggregation of information across scales.

The encoder-decoder structure of DiSOL performs an implicit form of domain decomposition. Coarser feature levels aggregate long-range interactions and global geometric context, while finer levels preserve local detail near boundaries and interfaces. Skip connections ensure that information is assembled coherently across scales.

This multi-scale assembly mechanism is particularly important for handling complex geometries and topological variations, such as domains with internal holes. Even when the domain connectivity changes, information can still propagate through coarse-scale representations and be reassembled at finer resolutions.

From this perspective, DiSOL does not rely on a single global operator. Instead, it constructs the solution through a sequence of local operators assembled across scales, analogous to multi-level solvers in classical numerical methods.

\subsection{Discrete operator learning versus continuous operator approximation}

Many neural operator approaches aim to approximate continuous operators that map input functions to output functions in an infinite-dimensional setting. In contrast, DiSOL is explicitly formulated as a discrete operator acting on fixed-resolution grids.

Rather than learning a continuous mapping $\mathcal{G} : f \mapsto u$, DiSOL learns a resolution-dependent discrete procedure $\mathcal{F_\text{h}}$ that operates on sampled fields. This distinction has important implications.

Because DiSOL is tied to a discrete representation, its behavior is governed by local neighborhood interactions and grid-level assembly rules. This makes the model sensitive to resolution, but simultaneously robust to geometric variability and boundary complexity within the same discretization framework.

This discrete-operator viewpoint explains why DiSOL generalizes well to unseen geometries and topological changes without requiring explicit geometric encoders or mesh-based representations. The learned operator behaves more like a numerical algorithm than a global function approximator.

\subsection{Implications for out-of-distribution generalization}

The out-of-distribution tests considered in this work—topological changes, fragmented boundary conditions, and high-frequency source terms—systematically violate assumptions implicit in the training data.

The strong performance of DiSOL in these settings can be attributed to the separation between local operator learning and global assembly. Local operators remain valid under distribution shifts, while the multi-scale assembly mechanism adapts to new global structures.

This interpretation highlights that DiSOL’s generalization capability arises from architectural inductive biases aligned with classical numerical reasoning, rather than from increased model capacity or task-specific tuning.

\clearpage
\section{Controlled comparison between DiSOL and UNet}

A natural question is whether the performance gains reported in the main text can be attributed to the use of a UNet--like multiscale convolutional backbone rather than to the discrete-operator formulation itself. To address this concern, we provide a controlled, architecture-level comparison between DiSOL and a capacity-matched UNet baseline on the Poisson benchmark. Importantly, this section is \emph{not} intended as a systematic module-wise ablation; instead, it aims to assess whether the discrete-operator structure in DiSOL yields measurable advantages over a conventional image-to-image CNN when backbone capacity and the training protocol are held fixed.

\subsection{Baseline construction and fairness protocol}
The UNet baseline is configured to share the same multiscale convolutional backbone depth and width as DiSOL. Both models take identical input channels and predict the same normalized solution pattern field on the embedded Cartesian grid. All training hyperparameters (optimizer, learning rate schedule, batch size, and number of epochs) and data splits are kept identical. For geometry-dependent domains, losses and evaluation metrics are computed on the active computational region (i.e., restricted to the geometry mask), so that comparisons reflect in-domain solution accuracy rather than trivial outside-domain padding effects. The two models have comparable capacity (UNet: $\sim$0.12M parameters; DiSOL: $\sim$0.13M), where the slight increase in DiSOL arises from the additional discrete-operator components rather than from enlarging the multiscale convolutional backbone. See Supplementary Information \ref{sect:3.2} for the detailed DiSOL architecture and masking protocol.

\subsection{Convergence and generalization comparison}

\begin{figure}
    \centering
    \includegraphics[width=0.8\linewidth]{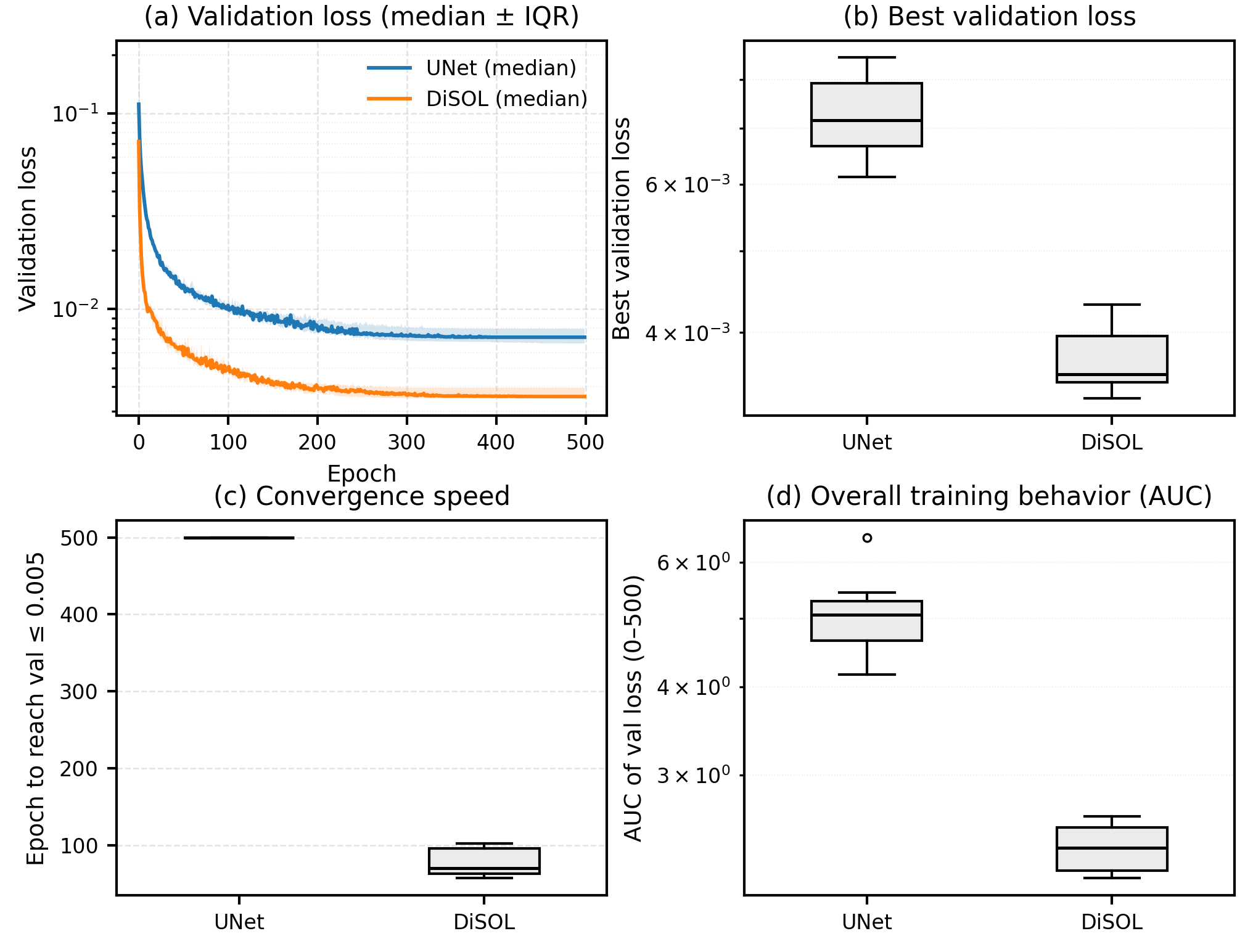}
    \caption{\textbf{Architecture-level comparison between UNet and DiSOL on the Poisson equation problem.} \textbf{a}, Validation loss trajectories summarized by the median and interquartile range (IQR) across 10 independent runs. \textbf{b}, Best validation loss achieved during training. \textbf{c}, Convergence speed measured as the number of epochs required to reach a fixed validation-loss threshold ($\le 5\times 10^{-3}$). \textbf{d}, Area under the validation-loss curve (AUC) over the full training horizon (0--500 epochs), serving as an integrated measure of overall training efficiency and stability (smaller is better). Both models share the same multiscale convolutional backbone depth and width, with comparable parameter counts (UNet: $\sim$0.12M; DiSOL: $\sim$0.13M, due to the additional local-operator and implicit problem-solving components). Across all metrics, DiSOL consistently achieves lower errors, faster and more reliable convergence, and smaller AUC values than the UNet baseline.}
    \label{fig:s_ablation}
\end{figure}

Figure~\ref{fig:s_ablation} summarizes training behavior over 10 independent runs.
DiSOL exhibits consistently faster and more reliable convergence, achieving lower validation losses throughout training and a lower best validation loss (Fig.~\ref{fig:s_ablation}a,b). We further quantify convergence speed using the epoch required to reach a fixed validation-loss threshold (Fig.~\ref{fig:s_ablation}c) and summarize overall training behavior using the area under the validation-loss curve (AUC) over the full training horizon (Fig.~\ref{fig:s_ablation}d), where smaller AUC indicates more efficient and stable optimization. Across all metrics, DiSOL improves both final accuracy and optimization efficiency relative to the UNet baseline.

\begin{figure}
    \centering
    \includegraphics[width=0.6\linewidth]{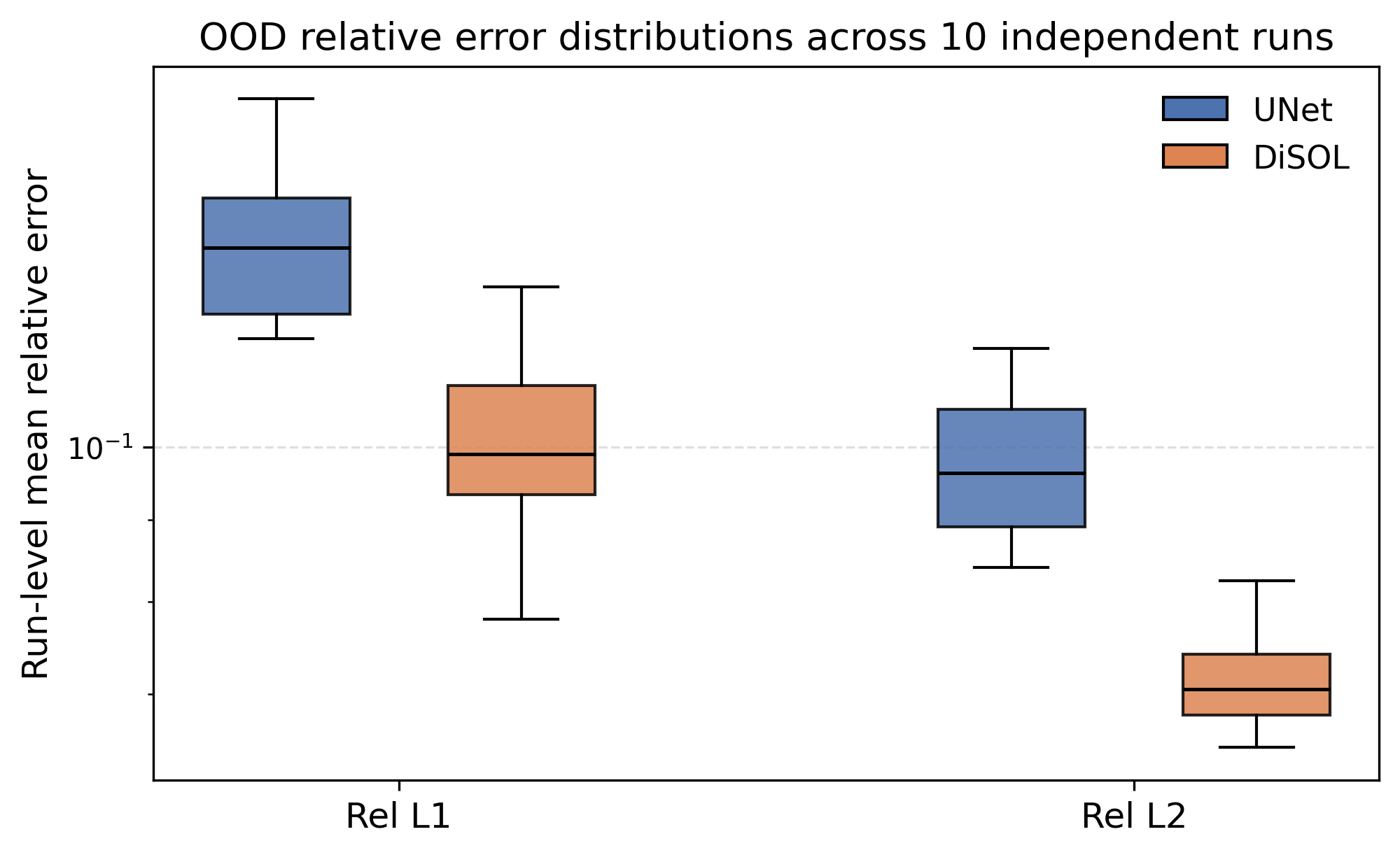}
    \caption{\textbf{OOD relative error distributions across independent training runs.} Boxplots of run-level mean relative errors (Rel L1 and Rel L2) on out-of-distribution (OOD) test geometries, evaluated over 10 independent training runs for UNet and DiSOL. For each run, the error is first averaged over all OOD test samples, and the resulting run-level means are aggregated into boxplots. DiSOL consistently achieves lower median errors and reduced inter-run variability compared to the UNet baseline under both relative error metrics, indicating improved robustness and generalization under geometric extrapolation.}
    \label{fig:s_ablation_OOD}
\end{figure}

We additionally evaluate robustness under geometry extrapolation.
Figure~\ref{fig:s_ablation_OOD} reports OOD test performance aggregated over the same set of 10 independent training runs, using run-level mean relative errors (Rel~L1 and Rel~L2). DiSOL consistently achieves lower median errors and reduced inter-run variability compared with UNet, indicating improved stability and generalization when tested on unseen geometries. Together, these controlled comparisons suggest that the gains of DiSOL cannot be explained solely by the presence of a UNet--style multiscale CNN backbone, but are instead consistent with the discrete-operator formulation and geometry-consistent internal routing.

\clearpage
\bibliographystyle{unsrtnat}
\bibliography{references}  




